\newcommand{\version}{0.9.4}

\newcommand{\prasp}{PrASP\ }

\documentclass[letterpaper]{article}
\usepackage[nottoc]{tocbibind}

\usepackage{amsmath,amssymb,times,xspace,graphicx}

\usepackage{algorithm,algpseudocode}

\usepackage[caption=false]{subfig}

\usepackage{listings}
\usepackage[usenames,dvipsnames]{xcolor}

\newcounter{definitionnumber}

\usepackage{times}
\usepackage{geometry}
\usepackage{hyperref}
\usepackage[usenames,dvipsnames]{xcolor}
\usepackage{graphicx}
\usepackage{textcomp}

\usepackage{afterpage}

\usepackage{graphicx}

\usepackage{fancyhdr}

\usepackage{float}

\usepackage[export]{adjustbox}

\newcommand{\qq}{\symbol{34}}

\newcommand{\col}{\color{Blue}}

\usepackage{amsopn}
\DeclareMathOperator{\hyphen}{-}

\algblock{ParFor}{EndParFor}
\algnewcommand\algorithmicparfor{\textbf{parfor}}
\algnewcommand\algorithmicpardo{\textbf{do}}
\algnewcommand\algorithmicendparfor{\textbf{end\ parfor}}
\algrenewtext{ParFor}[1]{\algorithmicparfor\ #1\ \algorithmicpardo}
\algrenewtext{EndParFor}{\algorithmicendparfor}

\DeclareTextCommandDefault{\nobreakspace}{\leavevmode\nobreak\ } % required for T1

\usepackage{makeidx}
\makeindex
\usepackage[totoc]{idxlayout}

\hypersetup{
  colorlinks,
  citecolor=BrickRed,
  linkcolor=RoyalBlue,
  urlcolor=RoyalBlue
  citebordercolor=RoyalBlue
}

\usepackage{mdwlist}

\newcommand{\dmh}{\rho}

\newcommand{\PrASP}{PrASP }

\newcommand{\kb}{\Lambda}

\newcommand{\as}{\Gamma}

\newcommand*\Let[2]{\State #1 $\gets$ #2}

\title{{\color{black}\vspace{1.1cm}
\begin{center}
\hspace{0cm}{\Large Reference Manual}\textbf{{\col
}}\\
\vspace{1.2cm}
\end{center}
}}

\date{}

\begin{document}

%\maketitle
%
%\bigskip
%\bigskip
%\bigskip
%\begin{center}
%%\col{{\Huge 	\textsf{DRAFT}}}
%\end{center}
%
%\thispagestyle{fancy}
%
%\newpage

\pagestyle{fancy}

\chead{}
\lhead{}
\rhead{}
\renewcommand{\headrule}{}
\cfoot{{\hspace{3.7cm} \thepage} \vspace{1.5cm}  \newline {\normalsize \col\textsf{\prasp \version}}}

\setlength\headheight{0.0pt}
\setlength\voffset{0.0pt}
\addtolength{\textheight}{150pt}

\definecolor{darkBlue}{RGB}{0,0,110}

{\hypersetup{linkcolor=darkBlue}

%\newpage
%\tableofcontents

%\afterpage{\cfoot{\thepage}}

\newpage

\begin{center}
{\Large \textbf{PrASP\footnote{Version 0.9.4} Report}} 
\end{center}
\vspace{0cm}
\begin{center}
Matthias Nickles%$^1$
\end{center}

\begin{center}
%$^1$Digital Enterprise Research Institute (DERI) \& Discipline of Information Technology\\
College of Engineering \& Informatics\\
National University of Ireland, Galway\\
University Road 1, Galway City, Ireland\\
{\small \verb§matthias.nickles@deri.org§}
\end{center}
\vspace{0.4cm}
\noindent \textbf{Abstract.} This technical report describes the usage, syntax, semantics and core algorithms of the probabilistic inductive logic programming framework PrASP. PrASP is a research software which integrates non-monotonic reasoning based on Answer Set Programming (ASP), probabilistic inference and parameter learning. In contrast to traditional approaches to Probabilistic (Inductive) Logic Programming, our framework imposes only little restrictions on probabilistic logic programs. In particular, PrASP allows for ASP as well as First-Order Logic syntax, and for the annotation of formulas with point probabilities as well as interval probabilities. A range of widely configurable inference algorithms can be combined in a pipeline-like fashion, in order to cover a variety of use cases.\\


\section{Introduction}
\label{overview}

\prasp \cite{prasp15,prasp12,prasp14,prasp13,prasp11} (\underline{Pr}obabilistic \underline{A}nswer \underline{S}et \underline{P}rogramming) is a Nilsson-style \cite{nilsson} probabilistic inductive logic programming (PILP) language (e.g., \cite{prism0,icl,progol,spirit,mln,mlnasp,problog,thimm1,plog,cozman1,lee2,Morais}) and an uncertainty reasoning and statistical relational machine learning software, based on \textit{Answer Set Programming} (ASP) \cite{aspasp,asp}. PILP is a form of declarative logical/relational probabilistic programming and ASP is a form of non-monotonic logic programming whose syntax is similar to that of Prolog but whose inference approach is more closely related to SAT, constraint satisfaction problem solving and Satisfiability Modulo Theories (SMT) solving.
Besides probabilistic ASP, PrASP also includes limited support for inference with probabilistic normal logic programs under non-ASP-based semantics (mainly for research and benchmarking purposes).

\subsection{Properties}

%\noindent Main features:
\begin{description}
	\item[High degree of expressiveness] PrASP provides a unified framework with support for Answer Set Programming (ASP) and First-Order Logic (FOL) syntax, interval probabilities (as in probability bounds analysis), conditional probabilities and Annotated Disjunctions \cite{annotatedDisjunctions}. Arbitrary logic formulas can be directly annotated with probabilities or probability intervals. 
	\item[Flexible inference approaches] Several configurable probabilistic inference approaches support a range of different use cases. Inference approaches currently supported by \prasp include linear programming, iterative refinement \cite{spirit}, various sampling approaches,%Gibbs sampling, multivariate Slice sampling (a form of Gibbs sampling with an auxiliary latent uniform random variable), Flip sampling \cite{prasp14}, 
	parallel Simulated Annealing \cite{prasp14}, and MaxWalkSAT \cite{kautz,mln}. 
	\item[Non-monotonic logic programming] Being based on Answer Set Programming, \prasp can be used for probabilistic non-monotonic reasoning. As a logic programming approach, it is able to handle inductive definitions (in contrast to approaches like MLN \cite{mln} which are based on FOL or fragments thereof, such as Description Logic). 
	\item[No mandatory independence assumption.] In contrast to many other PILP frameworks and probabilistic databases, \prasp does, by default, not assume probabilistic independence of uncertain events. However, some of its inference algorithms can make use of event (formula) independence in order to speed up inference. 
	\item[Uncertainty reasoning with annotated as well as unannoted formulas] PrASP approach to nondeterminism is essentially based on logical disjunction. Therefore, it provides seamless interoperability of probabilistically annotated statements and plain disjunctions or ASP choice constructs for the modeling of nondeterminism. 
	\item[Parameter learning.] \prasp can be used to learn the probabilities of given hypotheses from example data. This is a form of Inductive Logic Programming (ILP) which in the area of ASP has so far mainly been approached in the form of non-probabilistic induction of new hypotheses (structure learning) \cite{ASPILP}. 
	\item[Markov Logic] semantics is (experimentally) supported using switch \verb§--mlns§, in the sense of \cite{mlnasp}, i.e. combined with stable model semantics. In contrast to the original Markov Logic Networks, $\mathrm{LP^{MLN}}$ \cite{mlnasp} and \prasp (with its own semantics as well as \verb§--mlns§), being based on logic programming rather than classic FOL, support inductive definitions. (Note that with its native semantics, \prasp is not much related to MLN, apart from the fact that both frameworks support FOL syntax.) 
	%\item[Tools for result analysis] Inference outcomes can be analyzed wrt. criteria such as model set entropy and accuracy. 
	\item[Platform independence.] \prasp is programmed in Scala and thus runs on the widely available Java Virtual Machine (JVM)\footnote{Optionally, it can make use of native code libraries to improve performance on certain platforms.}. Its API can be used with Java, Scala, Clojure, Groovy and other languages which run on the JVM, and with other JVM-based frameworks (such as Spark).	However, for certain procedures (such as Linear Least Squares), \prasp uses platform-dependent external libraries where available (such as CUDA), and Java code as a fallback. 
	\item[Extensible syntax] using macros written in the Scala programming language
	\item[Streaming data] in RDF format can be used as input for inference and learning tasks (experimental feature). 
	\end{description}

 However, it should also be noted that the current version of \prasp is a research prototype which might still have a few rough edges. If you encounter bugs or if you want to suggest improvements of the framework or its documentation, please don't hesitate to contact the developer (see Sect. \ref{license}). \\

Background knowledge \index{Background knowledge} (\prasp \textit{programs}) can be uncertain (by attaching probabilities or probability intervals to formulas). Besides the native Prolog-like input language \textit{AnsProlog}, it is also possible to use formats such as First-Order Logic (FOL, using translation provided internally by \prasp itself or using an external translator such as F2LP \cite{f2lp}), action languages \cite{coala} and planning languages such as PDDL \cite{potassco} as input languages (with the help of external conversion tools). In addition to these syntax forms, so-called \textit{Annotated Disjunctions} are supported as ``syntactic sugar'' (in a similar way as in ProbLog).\\
  At this, \prasp imposes virtually no restrictions on the annotations of formulas with probabilities, although of course not all syntactically valid \prasp programs are consistent or efficient. \\
  
   \prasp includes a number of inference approaches, including linear equation system solving, simulated annealing, iterative refinement (using the approach from SPIRIT \cite{spirit}), model counting, and linear programming (LP). Inference algorithms are described in Section \ref{corealgos}. Currently supported inference tasks are computation of the conditional and non-conditional probabilities of given formulas (\textit{query formulas}), computation of probability bounds of query formulas, and MAP (Maximum A-Posteriori) inference (computing the most probable world). Parameter learning allows to compute the probabilities of given hypotheses inductively from learning examples and background knowledge.  

\subsection{Overview}
\label{overview}

Background knowledge (uncertain as well as certain knowledge provided by some domain expert) is given in form of a \prasp program (a probabilistic logic program). It provides a set of constraints for the set of possible worlds as well as for the valid probability distributions over those possible worlds. In \prasp terminology, possible worlds are also called \textit{models} and correspond to the answer sets\footnote{The idea to associate possible worlds with answer sets stems from \cite{plog}.} of the so-called \textit{spanning program} \cite{prasp12} of the \prasp program \ref{corealgos}. For probabilistic inference, \prasp computes one such probability distribution from the given background knowledge and subsequently uses this probability distribution to compute precise or approximate unconditional or conditional probabilities of the \textit{query formulas} (formulas whose probabilities the user wants \prasp to compute). Given the constraints specified by the \prasp program, each query might have multiple solutions (or more precisely, a range of probabilities), of which \prasp by default computes only a single point probability per query formula\footnote{Unless command-line arguments \texttt{--intervalresults} or \texttt{--ndistrs n} with $n>1$ are specified, see Section \ref{commandline}}.\\

A \prasp program can contain formulas in ASP syntax (\textit{AnsProlog}) as well as formulas in first-order logic (FOL) syntax, which allows for a wide range of applications (e.g, using Event Calculus axioms for probabilistic reasoning about actions and other events). Other formats require conversion to ASP before \prasp is called, using some appropriate translation software. ASP and FOL formulas can coexist within the same \prasp program, however, you should not mix both styles within the same formula.\\

Any formula can be annotated (parameterized) with a probability or an interval of probabilities, in order to specify uncertain beliefs. Such an annotation is called the \textit{weight} of the respective formula.\\
It is furthermore possible to specify conditional probabilities explicitly in background knowledge, although in many case \textit{Annotated Disjunctions} should be considered as a computationally cheaper alternative.\\
Formulas can contain variables (ASP-style as well as FOL variables), however, ``first-order'' inference is restricted to finite discrete domains.\\  
Given the background knowledge, \prasp can infer conditional as well as unconditional probabilities of ground and non-ground query formulas. Weights of given hypotheses can also be \textit{learned} inductively from example data. Such examples can in principle be arbitrary formulas, but typically they are ground atoms.\\

Background knowledge in form of a \prasp program is provided by the user as a file with file name ending  \verb§.prasp§. Such a file consists of a set of \textit{weighted} or unweighted logical formulas, plus optionally meta-statements. Each formula must be concluded by a dot ("\verb§.§") and must not contain line breaks\footnote{For the sake of better readability, we often omit the concluding dot in this document.}.\\
Besides \prasp programs, there are also \textit{query files} (which consist a number of query formulas whose weights should be computed by inference), files with \textit{hypotheses} (formulas whose weight should be learned from examples) and learning \textit{examples}. Examples for these tasks are provided later in this document, in a tutorial-style section (Sect. \ref{basic}).\\

\noindent Here's an example for background knowledge (i.e., a \prasp program) in native annotated AnsProlog syntax. Note that constants and predicates need to start with a lowercase letter whereas variables start with an uppercase letter. 

\begin{verbatim}
[0.45;0.5] coin(1,heads).
[[0.5]] coin(X,heads) :- X != 1.
win :- coin(1,heads), coin(2,heads).
\end{verbatim}

Formulas such as \verb§[0.5] coin(2,heads)§ are annotated with probabilities (\textit{weights}). Weights can be attached to any kinds of formulas (ground or non-ground facts and rules). In contrast to the types of weights found in some other probabilistic languages, the numbers within the square brackets are actual probabilities (e.g., $Pr(coin(2,heads)) = 0.5$) or probability intervals.\\

 Weights come in various syntactic forms, the most simple one being the form \verb#[#$w$\verb#]#, where $w$ is a probability. Probability intervals\index{Probability intervals} \verb§[§$wx$;$wy$\verb§]§ can also be used as weights. The informal meaning of a formula $f$ annotated with weight \verb#[#$w$\verb#]# $f$ is that this formula has the probability $w$ (that is, $f$ represents a random event and $Pr(f) = w$). Weights can be attached to all kinds of formulas in the background knowledge, including rules, facts, disjunctions, and formulas with existential or universally quantified variables over finite domains. For example,\\
\verb|[0.7] p(X) :- q(X).| specifies that the probability of non-ground rule \verb|p(X) :- q(X)| is $0.7$. Formulas which contain variables (here \verb§X§) are called \textit{non-ground}. Weighted non-ground formulas are interpreted as weighted conjunctions of ground formulas, if a single-square bracket annotation is used (there are other forms of annotations too, with different semantics, as explained later).\\

 However, of course not all syntactically valid \prasp programs are equally suitable for time- and memory efficient inference and learning. Therefore a significant part of this document is concerned with inference and optimization methods supported by \prasp, each with different characteristics in terms of speed, accuracy, requirements etc, in order to facilitate a wide range of different use cases for inference and learning (learning makes internally heavy use of inference).\\

Note that \prasp does not necessarily assume probabilistic independence of formulas for which 
no \textit{logical} dependence (e.g., by means of some logical rule which relates two random events) is specified. Again in contrast to many other probabilistic logics, the user can specify which sets of formulas should be treated as independent and which kind of independence should be assumed. E.g., in the program above, \verb§coin(1,heads)§ and \verb§coin(2,heads)§ could be explicitly declared as pairwise independent (by adding a certain meta-statement \ref*{meta} or indirectly using conditional probabilities), or their independence could be left unspecified (e.g., in order to compute interval results of all possible query outcomes), or automatically discovered by \prasp. Independence assumptions 
can have a tremendous influence on inference performance, as they allow for certain simplifications which
are not possible otherwise. On the other hand, inference algorithms which cannot apply such simplifications are slowed down (!) if event independence is being assumed.\\

Various other types of weights exist. E.g., \textit{Computed weights} allow for symbolic weights whose numerical values are computed during \textit{grounding}\footnote{``Grounding'' means that an ASP program with variables is replaced by an equivalent program without variables. Most ASP solvers require grounded input. Grounding is either performed by a separate grounding tool or by a combined grounder/solver. Contemporary ASP grounders typically also perform simplifications during grounding by removing ground formulas of which they know cannot be true. Grounding can be limited by using simplification switches such as \hyperref[cmdline:mod1]{\texttt{--mod1}}.} (either using ASP or using a procedural scripting language) or random weights. For details, please see Section \ref{weights}.\\

It is possible to specify \textit{conditional probabilities} directly in background knowledge, using the following weight annotation syntax: \verb#[#$w$\verb#|#$c$\verb#] #$f$. This declares that $Pr(f|c) = w$, i.e., that the probability of $f$ is \verb|w|, given that we already know that $c$ holds. Both $f$ and $c$ are formulas. While within formulas, symbol \verb#|# stands for logical disjunction, here its first occurrence in annotation \verb#[#$p$\verb#|#$c$\verb#]# separates the probability $p$ from probabilistic condition $c$.
Example: \verb#[0.75|not sunny] rainy | snowy# specifies that the probability that it is either rainy or snowy is 0.75, given that we know it's not sunny. Conditional probabilities in background knowledge come with a performance penalty though, and alternatives available in \prasp, such as certain patterns of annotated rules (Sect. \ref{basic}) or Annotated Disjunctions (Sect. \ref{annotatedDisjunctions}), should be considered.\\
Like in any formulas, variables, logical connectives, etc., logical variables are allowed both within \verb|f| and the condition formula \verb|c|.\\

It is possible to specify weights in conditional as well as unconditional probabilities using so-called double- or triple-square notion (e.g., \verb#[[#$w$\verb#|#$c$\verb#]]# or \verb#[[[#$w$\verb#]]]#), in order to declare uncertain \textit{ground} instances of non-ground formulas in compact form (see Sect. \ref{weights}).  Variables can be used within ordinary weighted formulas (single-square annotations) too, but the semantics is different from double- or triple-square notion (see Sect. \ref{weights}).\\
 
\prasp can discover and warn against probabilistically inconsistent weight annotations (e.g., if the background knowledge contains both \verb|[0.3] p| and \verb|[0.9] not p|), provided the default inference approach is being used (so-called ``\prasp vanilla'' mode), but basically it is the responsibility of the user to provide probabilistically and logically consistent background knowledge.\\

Formulas in ASP syntax can contain most of the syntactic constructs supported by contemporary ASP grounders, such as \textit{aggregates}. FOL syntax is by default as specified for the input language of F2LP \cite{f2lp} (which is akin to the TPTP format), although it is also possible to use the \prasp-internal FOL$\rightarrow$ASP converter (which uses almost the same formula syntax, see switch \verb§--folconv§ for details) or other external converters.\\
In accordance with AnsProlog customs, variable names need to start with an upper-case letter and constants need to start with a lower-case letter. The domains of quantified variables in formulas using FOL-syntax need to be finite (because FOL-formulas are translated into ASP programs).\\

\prasp programs can contain meta-statements understood by the respective external ASP grounding tool, such as as domain declarations, as well as various \prasp-specific meta-statements (see Section \ref{meta}). Meta-statements start with character \verb|#|.\\
 
%In addition, everywhere where the ASP rule operator \verb|:-| is allowed it is allowed to use operator \verb|<--| instead, except for rules where default negation (i.e., \verb|not|) occurs in the body. \verb|h <-- b1, b2, ...| is a shortcut for \verb#h | (-b1 | -b2 | ...) #. 
Like plain ASP, \prasp distinguishes between \textit{classical negation} and \textit{default negation}. Prefix ``\verb|-|'' denotes strong (classical) negation and prefix ``\verb|not|'' stands for default negation, as usual. Note that \prasp works by default with default negation in order to implement $Pr(\neg x) = 1-Pr(x)$ (as an experimental feature which might disappear in future versions, this behaviour can be switched to the use of classical negation).\\

\underline{Important}: \prasp's default settings are rather conservative - by specifying an algorithm and parameters which are specific to the respective inference problem, inference (and indirectly learning) speed can be massively increased (see Section \ref{performance} for simple approaches to achieve this). \\

\noindent The remainder of this technical report is organized as follows:

\begin{itemize}
\item Section \ref{installation} describes how to install PrASP.
\item Section \ref{basic} provides an informal introduction to the use of PrASP for basic inference and learning tasks,
as well as tips for performance tuning.
\item Section \ref{syntax} describes PrASP's syntax.
\item Section \ref{corealgos} describes its semantics, and core inference and learning algorithms.
\item Section \ref{confopts} describes the software's command line options.
\item Section \ref{hints} provides miscellaneous hints for troubleshooting.
\item Section \ref{faq} contains answers to frequently asked questions.
\item Section \ref{license} contains developer contact details and a disclaimer.
\end{itemize}

%\newpage
\section{Installation \& Web Interface}
\label{installation}

\label{simplifiedInstallation}
{\color{Black} 
%\noindent \textit{\textbf{Internal prototype only}}:
%\begin{quote}

Installation of \prasp is easy, since on most machines it only requires unzipping a file archive and copying one or two additional tools into the resulting directory. There needs to be a recent version of Java (Java 8 or newer, 64bit version). If you just want to quickly try out PrASP online, there's also the \textit{PrASP Web Interface}.

\subsection{Web Interface}

The PrASP Web Interface for online inference and learning can be found at\\
\noindent \verb§http://ubuntu1.it.nuigalway.ie:8977/PrASP_WebInterface/static/ABOUT.html §\\
\noindent However, it is limited in several respects compared to the installable version of PrASP and
only suitable for small tasks.

\subsection{Installation steps}

\begin{enumerate}
\item Unzip \verb§prasp.zip§ into a new directory (henceforth called the \prasp directory). Make sure PrASP has writing access to this directory. There is no installation program.
\item Put third-party \textit{external tools} directly into the \prasp directory, if they are not already there. See below for the list of these tools (of which some are required and others are optional but strongly recommended). Make sure the filenames of their executables are as specified below (so that PrASP finds them). However, some of these tools might be already installed and ready to be used, depending on the \prasp file archive you have. See Sect. \ref{externaltools} for details.
\item \prasp can optionally call native code for faster inference on the CPU or a Nvidia GPU. By default, \prasp expects native BLAS/LAPACK libraries. Under Windows and Mac OS X, normally no additional installation is required in this regard, unless you want to install your own BLAS implementation. Under Linux,\\
 \verb§sudo apt-get install libatlas3-base libopenblas-base§ is normally sufficient (tested with Debian/Ubuntu).\\
 You can also disable the use of native BLAS/LAPACK altogether, see Sect. \ref{nativelibs}.\\
 As for  native GPU code, installation of additional libraries is normally not required, but see Sect. \ref{nativelibs} for exceptions (in particular for Mac OS X). See \hyperref[cmdline:linsolveconf]{\texttt{--linsolveconf}} for how to activate native solver code.
\item If you are using Mac OS X or Linux, you should make \verb§prasp.sh§ executable using \verb§chmod +x prasp.sh§
\item Enter \verb|./prasp.sh examples/example1.prasp examples/example1.query| (Linux, MacOSX) or
\verb|prasp.bat examples/example1.prasp examples/example1.query| (Windows) for a first test of your installation\footnote{Example 1 is in Gringo 4 syntax. If you are using a different ASP grounder, you might need to tweak it a bit.}.

\item It is highly recommended to read the rest of this section too (even if you already successfully ran the example), in particular Subsection \ref{externaltools}.

\end{enumerate}

\subsection{External tools}
\label{externaltools}

 As a minimum, you should\footnote{It is possible to perform a - very limited - range of probabilistic inference tasks  without any ASP grounder or solver, e.g., MAP inference using \hyperref[cmdline:maxwalksat]{\texttt{--maxwalksat}}. Also, \prasp contains its own grounder which is able to ground basic non-ground formulas during the preprocessing phase. } have an ASP grounder (recommended: Gringo \cite{potassco}) and an ASP combined grounder/solver (recommended: Clingo \cite{potassco}) installed.  \\
 
 The other external tools listed below (F2LP \cite{f2lp}, CVC4 \cite{cvc4}) are recommended in order to make full use the framework but not strictly required.\\
 
All of the tools listed below are open source and available in binary form for Linux, MacOSX and Windows. Depending on your \prasp distribution, some of them might already be installed.\\
\textbf{To install, simply copy these tools into the top PrASP directory if they aren't already there, and make sure that the names of their executable binary files are exactly as specified below (so that \prasp can find them).} 

\begin{description}
	\item[- Clingo and Gringo]\cite{potassco} are the most appropriate ASP grounder and solver for PrASP. They can be obtained from \verb|http://potassco.sourceforge.net/|. You need to decide if you want to use version 3 or version 4 (or higher) of these tools. Each of these versions has its pros and cons wrt. PrASP, see Sect. \ref{clingo4}. If you are unsure which version to use, choose Clingo 4 and Gringo 4.\\ 
	
	\underline{Installation}: Copy both tools into the top \prasp directory and make sure that the names of the executable files are \textbf{\texttt{gringo}} and \textbf{\texttt{clingo}} under Linux, \textbf{\texttt{clingo\_macosx}} and \textbf{\texttt{gringo\_macosx}} under Mac OS X, and \textbf{\texttt{clingo.exe}} and \textbf{\texttt{gringo.exe}} if you are using Windows.\\
	
	If these files are already in your \prasp folder and you are fine with their versions (check with \hyperref[cmdline:version]{\texttt{--version}}), you don't need to do anything. You can switch between Clingo3/4 and Gringo3/4 using shell scripts \verb§toggle...§\\
	
	Other ASP grounders/solvers might also work if they are compatible with Lparse/Smodels or the ASP language standard. For DLV code whose syntax goes beyond the common ground with Lparse/Gringo, a converter such as \verb§dlvtogringo§ \footnote{\texttt{http://potassco.sourceforge.net/labs.html}} might be helpful (but we haven't tested this yet). See \hyperref[cmdline:grounder]{\texttt{--grounder}} and \hyperref[cmdline:groundersolver]{\texttt{--groundersolver}} for how to specify different ASP grounder/solvers.
	
	\item[- F2LP]\cite{f2lp} (optional) is an external tool for translating program in FOL syntax (under the stable model semantics) into ASP logic programs. It can be obtained from\\ \verb§http://reasoning.eas.asu.edu/f2lp/§. You can alternatively use PrASP's built-in FOL$\rightarrow$ASP converter FOL2ASP, which is active by default (or tell \prasp not to use any converter at all, which increases performance but obviously restricts you to ASP syntax). FOL2ASP provides better support for Gringo/Clingo $\geq$ 4 (as of May 2016), however, F2LP is still somewhat more mature.\\
	Our recommendation is to try with the internal converter first if you are using Gringo/Clingo 4 or higher, and to install F2LP if you are still using Gringo/Clingo 3 or if you experience difficulties with the current (beta) version of the internal converter. You can switch between both at any time using \hyperref[cmdline:folconv]{\texttt{--folconv}}.\\	
		
	\underline{Installation}: Copy the tool into the top \prasp directory. The name of the executable binary file must be \textbf{\texttt{f2lp§}} (Linux), \textbf{\texttt{f2lp.exe}} (Windows) or \textbf{\texttt{f2lp\_macosx}} (Mac OS X) in order to be discovered by \prasp.
	
	\item[- CVC4]\cite{cvc4}  (optional, strongly recommended) can be obtained from \verb|http://cvc4.cs.nyu.edu|. CVC4 is a SMT solver which is invoked by \prasp if the built-in linear system solver couldn't find a solution or if there are intervals in weight annotations and the ``vanilla'' approach to inference is used (however, it is recommended to specify switch \verb§--intervalresults§ in that case, which activates a different solver which deals with intervals much faster than CVC4). CVC4 is not strictly required, but its installation is strongly recommended.\\	
	
	\underline{Installation}: Simply copy the tool (including its external libraries) into the top \prasp directory. Its executable name must be \textbf{\texttt{cvc4}} (Linux), \textbf{\texttt{cvc4\_macosx}} (Mac OS X) or \textbf{\texttt{cvc4.exe}} (Windows). {On a Mac, you might need to create the CVC4 executable} first from \verb§cvc4-1.2_3.mpkg§ (or newer), see below for details. To check if CVC4 is already installed, call \prasp with option \hyperref[cmdline:enforceSMT]{\texttt{--enforceSMT}}.
\end{description}

\noindent \textbf{\underline{Further installation hints}}

\begin{itemize}
	\item On a Mac, the installation of CVC4 might require one further step: First, check if CVC4 is already installed by calling \prasp with option \hyperref[cmdline:enforceSMT]{\texttt{--enforceSMT}}. If you receive an error message indicating that no SMT solver was found, please install CVC4 using \verb|cvc4-1.2_3.mpkg|, copy the resulting executable file from \verb§/opt/local/bin§ into the \prasp folder and rename this file to \verb|cvc4_macosx|
	\item Other ASP grounders/solvers than Gringo/Clingo might be usable as well, to varying degrees, if they are compatible with Lparse/Smodels or the ASPCore-2 language standard. To specify a different grounder/solver use command-line arguments \hyperref[cmdline:grounder]{\texttt{--grounder}} and \hyperref[cmdline:groundersolver]{\texttt{--groundersolver}}. \\
	\prasp also supports the combined ASP / Constraint Satisfaction Problem (CSP) solver clingcon \cite{potassco}. To use clingcon, specify \verb|--groundersolver clingcon| on the command-line.
	\item If you can obtain or build the (non-default) portfolio binary \verb|pcvc4| of CVC4 with multi-threading enabled, this is preferred over the default version of CVC4. If you are using the portfolio binary, invoke \prasp with command-line option \verb|--SMTsolver pcvc4| (if \verb|pcvc4| is the name of the executable binary). For Linux 64bit, pre-build portfolio binaries might be available from\\
	{\small \verb|http://cvc4.cs.nyu.edu/cvc4-builds/portfolio-x86_64-linux-opt/|}
	\item Note that \prasp works generally faster if only ASP syntax is used and command-line option\\
	\hyperref[cmdline:folconv]{\texttt{--folconv none}} is specified.
	\item Modify the \prasp start script \verb|prasp.sh| (Linux, Mac OS X) or \verb|prasp.bat| (Windows) if necessary. In particular, make sure that a \verb|java| is found and that the amount of heap memory assigned to Java is large enough (in case you encounter ``out of memory'' errors).
	%\item \prasp is not very picky regarding Java VM arguments - if some \verb§java§ argument isn't supported with your Java version, just try without it.
	\item The current version of \prasp requires at least a Java JRE 8. If you cannot install Java 8 (=1.8) or higher on your system (or in case of any other installation problem), please contact the \prasp developer for advice (see Sect. \ref{license}).
\end{itemize}

%- If there is only an older binary file version of \verb§http://reasoning.eas.asu.edu/f2lp/§ available for your operating systems, consider compiling a recent version of F2LP from source code - it is pretty easy.

\subsection{About Gringo / Clingo 4 vs. 3}
\label{clingo4}

The ASP examples in this document are mostly still in Gringo/Clingo 3 syntax, however, since v0.8.10, \prasp also supports most of Gringo/Clingo 4 syntax (and the amount of supported syntax will be further increase in future \prasp versions). Gringo/Clingo 4 are already preferred over version 3 for most \prasp use cases.\\

Gringo/Clingo 4 provides significant advantages over older versions in connection with PrASP, in particular wrt. sampling performance. There are a few issues too though... In detail, Clingo/Gringo 4\\

 ...allow for (fast) remote sampling (which is active by default if \prasp discovers Clingo 4 - see switch \texttt{--noremotesampling} for details).\\

...aren't fully compatible with the external FOL$\rightarrow$ASP converter F2LP (at least up to version 1.3). If you experience issues, you might want to specify the internal converter (which is automatically used if you don't install F2LP) or deactivate FOL conversion altogether using command-line argument \hyperref[cmdline:folconv]{\texttt{--folconv}}.\\

...support and require ASP standard syntax\\
 (\verb§https://www.mat.unical.it/aspcomp2013/ASPStandardization§).  Beware that some of the differences compared to Gringo 3/Lparse syntax are quite subtle wrt. their effect (e.g., for aggregates), see Gringo/Clingo 4 user manual. \\
 
 ...do not yet allow for automated discovery of formula independence.\\

 Please refer to the Gringo/Clingo 4 manual for any required syntax modifications if you want to use Gringo/Clingo 4. Note that many of the syntax differences are very simple where a few others are quite subtle, in particular for non-trivial aggregate constructs such as \verb§#count§.\\
 
 Scripts \verb§toggleClingo3to4.sh§ and \verb§toggleClingo4to3.sh§ (Mac OS X, Linux) and\\ \verb§toggleClingo3to4.bat§ and \verb§toggleClingo4to3.bat§ (Windows) switch between Gringo/Clingo 3 and 4, simply by renaming/copying the respective Gringo/Clingo executables files so that \prasp uses them by default. Alternatively, the grounder and grounder/solver can be specified using command-line options \hyperref[cmdline:grounder]{\texttt{--grounder}} and \hyperref[cmdline:groundersolver]{\texttt{--groundersolver}}.

\subsection{Native libraries} 
\label{nativelibs}

\prasp can optionally make use of native CPU and/or GPU code in order to speed up inference. \\
If you want to make use of this, there are the following groups of libraries to consider during installation. After installation, call \prasp with option \hyperref[cmdline:linsolveconf]{\texttt{--linsolveconf}} to configure the use of native solver code. Note that currently only the ``vanilla'' approach to inference makes use of native code, so even with this, speed is rather low compared to approximate inference approaches.

\begin{description}
	\item[BLAS/LAPACK libraries (CPU):] \textbf{Mac OS X} and \textbf{Windows} users normally don't need to install additional libraries in this regard, as these libraries are either already included with the operating system (CPU LAPACK/BLAS code on MacOSX) or with \prasp itself. However, with regard to LAPACK/BLAS under Linux and Windows, it might make sense to install variants of these which provide higher performance compared to the default ones. 
	
	To install native BLAS/LAPACK libraries under \textbf{Linux} (tested with \textbf{Debian/Ubuntu}), do, e.g., the following in order to install OpenBLAS (but there are alternative BLAS/LAPACK implementations too):
	
\begin{verbatim}
	sudo apt-get install libatlas3-base libopenblas-base
	sudo update-alternatives --config libblas.so
	sudo update-alternatives --config libblas.so.3
	sudo update-alternatives --config liblapack.so
	sudo update-alternatives --config liblapack.so.3
\end{verbatim}

See \verb§https://github.com/fommil/netlib-java/§ for details. You might need to edit \verb§prasp.sh§ or \verb§prasp.bat§ afterwards.\\
You can configure which BLAS/LAPACK installation is used using the \verb§-Dcom.github.fommil.netlib§... arguments in \verb§prasp.sh§ or \verb§prasp.bat§ (Windows).\\
Again, see \verb§https://github.com/fommil/netlib-java/§ for details.

To \textbf{deactivate} the use of native BLAS/LAPACK libraries, change the \verb§-Dcom.github.fommil.netlib.BLAS§, \verb§-Dcom.github.fommil.netlib.LAPACK§, \verb§-Dcom.github.fommil.netlib.ARPACK§ parameter values in the \verb§java§ call in \verb§prasp.sh§ or \verb§prasp.bat§ to
\begin{verbatim}
-Dcom.github.fommil.netlib.BLAS=com.github.fommil.netlib.F2jBLAS
-Dcom.github.fommil.netlib.LAPACK=com.github.fommil.netlib.F2jLAPACK
-Dcom.github.fommil.netlib.ARPACK=com.github.fommil.netlib.F2jARPACK
\end{verbatim}

\item[Native CUDA libraries (Nvidia GPUs):]

Under \textbf{Linux} and \textbf{Windows}, this should work without the need to install additional libraries, since the necessary redistributable runtime CUDA libraries are part of \prasp. 
However, you might have to update your Nvidia graphics card driver to enable CUDA 7. Installing the (free) CUDA 7 Toolkit (\verb§https://developer.nvidia.com/cuda-toolkit-archive§) is one possibility to ensure that your system have the correct driver installed. Also, relevant runtime libraries, if still needed, can be obtained this way (but you won't need the compiler and other developer tools included in the toolkit).\\

Under \textbf{Mac OS X}, it is recommended to install the free CUDA 7.0 Toolkit\\ (\verb§https://developer.nvidia.com/cuda-toolkit-archive§), this should automatically provide suitable native CUDA libraries. If you experience "Image not found" errors on El Capitain (due to a well known issue with this platform), see \verb§prasp.sh§ for a possible workaround.

\end{description}

\section{Working with \prasp}
\label{basic}

\subsection{Basic probabilistic inference} 

\subsubsection{Invoking \prasp for query answering}

\noindent For probabilistic inference, \prasp is typically called as follows:\\

\noindent\verb|./prasp.sh|\footnote{With Windows, use \texttt{prasp.bat} instead of \texttt{prasp.sh}.}\verb| <file1.prasp> <file2.query>|\\

\verb|<file1.prasp>| is a \prasp program file which represents the available certain and/or uncertain \textit{background knowledge}, that is, the probabilistic and non-probabilistic rules and facts from which the query answers can be inferred. The syntax of background knowledge is described in Sections \ref{overview} and \ref{syntax}. \\

\verb|<file2.query>| is the \textit{query file}, consisting of a number of \textit{query formulas} of the form \verb|[?] f.| or \verb#[?|c] f.# or \verb#[[?]] f.# or \verb#[[?|c]] f.# or \verb#[[[?|c]]] f.#\\

With these arguments, \prasp is instructed to compute the probabilities of the query formulas, i.e., to find concrete values for the question marks in the formula annotations.\\

The file name extensions need to be \verb|.prasp| for the background knowledge file and \verb|.query| for the query file.\\
\noindent The command line above is an abbreviation of \verb|-b <file1.prasp> -q <file2.query>|.
Using the latter form, the file name extensions don't need to be  \verb|.prasp| / \verb|.query|.\\

For programs with a large number of answer sets and/or a large number of query formulas in the query file, it is recommended to provide additional command-line arguments in order to speed up inference. Please see Section \ref{performance} for details.\\

\noindent Results are written to \verb|stdout|.\\ 

A query file (\verb|<file2.query>| above) can contain multiple query formulas.
There are several syntax forms for query formulas: use \verb|[?]| $f$ to obtain an approximation of the unconditional probability of $f$. Use \verb#[?|#\kern-.05em $c$\kern-.02em \verb|]| $f$ to get an approximation of the conditional probability $Pr(f|c)$ of $f$ given $c$. Both $f$ and $c$ can be any formulas (in ASP or FOL syntax). $c$ is typically the observed evidence. Later in this document further forms of query formulas (using double- and triple-square bracketed weights) will be introduced. \\

\noindent Resulting probabilities are represented in the form \verb§[computedWeight] queryFormula§. Conditional probabilities  $Pr(f|c)=p$ are displayed in the form \verb#[p|#\kern-.05em $c$\kern-.02em \verb|]| $f$.\\

\noindent In case \prasp couldn't compute the probability of a query formula for some reason, \verb§[?] queryFormula§ is returned.\\

Conditional probabilities $Pr(f|c)$ can also be used in background knowledge using syntax  \verb#[p|#\kern-.05em $c$\kern-.02em \verb|]| $f$ (like all formulas, they need to be concluded by a dot). See further below in this section for examples and computationally more efficient alternatives.\\

Numerical weights are not allowed within query files. But optionally, unannotated formulas and \verb§#domain§ meta-statements can be put in the query file in order to specify domain predicates for the bindings of variables, which is useful if the respective domains are only needed in query formulas (otherwise, domains and domain predicates can simply be specified in the background knowledge file, like any other facts). \\

In case of large programs, inference and learning becomes tractable only by activating program optimizations or approximation algorithms such as \hyperref[cmdline:itrefinement]{\texttt{--itrefinement}}, \textit{simplification of conjunctions of independent formulas} (command-line option \hyperref[cmdline:mod0]{\texttt{--mod0}}), \textit{modularization} (command-line option \hyperref[cmdline:mod1]{\texttt{--mod1}}) or \hyperref[cmdline:simanneal]{\texttt{--simanneal}}. Shortcuts for combinations of certain optimization-related switches are \verb§-o1§, \verb§-o2§, \verb§-o3§, etc. (see \ref{commandline}). However, optimization measures might lead to less accuracy. Details about how to improve inference or learning speed are described in Sect. \ref{performance} and in Sect. \ref{commandline}.

\subsubsection{A first example: Coin flipping}
\label{coins}
\noindent Here's a first simple example for probabilistic inference. Let's assume we want to model a simple coin tossing game with two independent coins (\verb|coin1| and \verb|coin2|). If both coins show ``heads'', the player wins. The first coin is biased (heads appears with probability 0.6). We will later see how to model this scenario more elegantly using variables, but for the moment, we stick to a simple formalization and store the following \prasp program in file \verb|kb1.prasp| (the background knowledge):

\begin{verbatim}
#indep
[0.6] coin1(heads).
[0.5] coin2(heads).
#endIndep
1{coin1(heads),coin1(tails)}1. 
1{coin2(heads),coin2(tails)}1. 
win :- coin1(heads), coin2(heads).
\end{verbatim}

\noindent We are now interested in the probability of winning the game, and in various other probabilities inferable from the background knowledge. We put the following queries into file \verb|test1.query|:

\begin{verbatim}
[?] coin1(tails).
[?] coin2(heads).
[?] coin1(heads) | coin1(tails).
[?] not (coin1(heads) | coin1(tails)).
[?] coin1(heads) & coin2(heads).
[?] win.
[?|coin1(heads) & coin2(heads)] win.
\end{verbatim}

\noindent Calling \prasp using 

\noindent \verb|./prasp.sh|\footnote{With Windows, invoke \texttt{prasp.bat} instead of \texttt{prasp.sh}. From now on, we omit denoting the start script.} \verb|kb1.prasp test1.query |\\

\noindent gives the following results:

\begin{verbatim}
% query1.query

[0.4] coin1(tails).
[0.5] coin2(heads).
[1] coin1(heads) | coin1(tails).
[0] not (coin1(heads) | coin1(tails)).
[0.3] coin1(heads) & coin2(heads).
[0.3] win.
[1|coin1(heads) & coin2(heads)] win.
\end{verbatim}

The first line tells us that the probability that coin number 1 comes up with ``tails'' is 0.4. This follows from the
fact that the probability of ``heads'' is 0.6 for this coin and that formula \verb|1{coin1(heads),coin1(tails)}1| in the background knowledge
enforces that in each possible world either \verb|coin1(heads)| or \verb|coin1(tails)| is true, but never both together. \\
\verb#[1] coin1(heads) | coin1(tails)# means that the probability that either ``heads'' or ``tails'' appears is 1.0, and  
\verb|[0.3] coin1(heads) & coin2(heads)| denotes probability 0.3 for the event that both coins show ``heads'' after being tossed, which is also the unconditional probability of \verb|win|, i.e., of winning the game. Observe that formula \verb|coin1(heads) & coin2(heads)| is in FOL/F2LP \cite{f2lp} syntax.\\

\noindent The last result line (\verb#[1.00|coin1(heads) & coin2(heads)] win#) reads:\\
``The probability of \verb|win| given that we know that \verb|coin1(heads) & coin2(heads)| holds is 1.00'' (that is, $Pr(win|coin1(heads) \wedge coin2(heads)) = 1$).\\

The example also show that ASP and FOL syntax can be freely mixed in \prasp programs and queries (albeit not within the same formula). However, the semantics
for the non-probabilistic part of our logic is always the stable model semantics used in Answer Set Programming - first-order formulas are internally transformed into equivalent ASP constructs (more precisely, into disjunctive logic programs).

%%Note that \verb|1{coin1(heads),coin1(tail)}1| does not ensure that the probability 0.4 of 
%%\verb|coin1(tail)| can be inferred here, even though it seemingly guarantees that tail 
%%comes up with frequency 1 - 0.6 = 0.4 for coin 1. But since we use command-line option \verb|--solve|, the probabilities
%%of possible worlds are not computed from the frequencies of answer sets but by
%%solving a system of linear equations. However, \prasp can infer the probability of \verb|coin1(tail)|
%%because it adds \verb|[1-p] not (f)| for each annotated formula \verb|[p] f|.\\
%If \verb|--solve| is not specified, frequencies of possible worlds (i.e., answer sets) are used to infer probabilities,
%so theoretically, \verb|1{coin1(heads),coin1(tail)}1| would be sufficient to fix $Pr($\verb|coin1(tail)|$) = 0.5$.
%Yet omitting \verb|--solve| works only in special cases. (...)\\

\subsubsection{Probabilistic inference in the absence of weighted beliefs}

Probabilistic inference does not necessarily rely on the presence of weighted formulas in the background knowledge. Consider the following variant of the famous ``Tweety''-example for non-monotonic logic programming \cite{potasscoUserGuide}:

\begin{verbatim}
fly(X) :- bird(X), not neg_fly(X).
neg_fly(X) :- bird(X), not fly(X).
neg_fly(X) :- penguin(X).

bird(tweety).
chicken(tweety).
bird(tux).
penguin(tux).
\end{verbatim}

\noindent We store this as a \verb§.prasp§ file and put the following into the \verb§.query§ file:

\begin{verbatim}
[?] fly(tux).
[?] fly(tweety).
\end{verbatim}

\noindent The inference result is

\begin{verbatim}
[0] fly(tux).
[0.5] fly(tweety).
\end{verbatim}

We get these results because our background knowledge encodes that Tux is a penguin and penguins definitely don't fly, whereas Tweety is not a penguin and thus Tweety may fly or not. Due to entropy maximization (which will be addressed later), the uncertainty about whether Tweety can fly or not is quantified as 0.5 (which is exactly ``in the middle'' between 0 (cannot fly) and 1 (can fly)).\\

\noindent If we call \prasp with switch \hyperref[cmdline:intervalresults]{\texttt{--intervalresults}}, we obtain 

\begin{verbatim}
[0;0] fly(tux).
[0;1] fly(tweety).
\end{verbatim}

\noindent This also reflects our uncertainty about Tweety's flying capability. However, in contrast to the previous result, it shows the full range of probabilities instead of a point probability.

\subsubsection{Weighted rules and conditional probabilities}

\noindent How about weighted rules? Let's get back to the coin flipping example (\ref{coins}) and add the following to the background knowledge (i.e., file  \verb|kb1.prasp|):

\begin{verbatim}
[0.8] happy :- win.
\end{verbatim}

\noindent With the following additional queries:

\begin{verbatim}
[?] happy.
[?|win] happy.
\end{verbatim}

\noindent ...we get, maybe surprisingly, the following results:

\begin{verbatim}
...
[0.1] happy.
[0.333333333|win] happy.
\end{verbatim}

Why that? The reason is that \verb|[0.8] happy :- win| does not denote the conditional probability 0.8 of being \verb|happy| given \verb|win|, but instead the probability 0.8 of the event which is specified as the logical implication $win \rightarrow happy$, i.e. $\neg win \vee\ happy$.\\
This exemplifies that in general, logical rules do not correspond to conditional probabilities. Examples for proper use of weighted rules are shown further below in this document. Anyways, in order to specify the conditional probability $Pr(happy|win) = 0.8$, we could replace 
\verb|[0.8] happy :- win| in the background knowledge with the following two lines:

\begin{verbatim}
[0.8|win] happy.
:- happy, not win.
\end{verbatim}

\noindent (The last line expresses that there shall be no happiness without winning.) Now the query results are as expected:

\begin{verbatim}
...
[0.24] happy.
[0.8|win] happy.
\end{verbatim}

\noindent Note that conditions (the formulas between "\verb§|§" and "\verb§]§") are not restricted to facts, they can be any formulas.\\

\noindent \underline{Remark}: the probabilities of all conditioned as well as unconditioned weighted formulas (both in the background knowledge and in queries) are implicitly conditioned on the background knowledge, or more precisely, on the so-called spanning program - what that exactly means will be described later, but basically it just says that each weighted formula has the conjunction of all non-weighted formulas in the background knowledge as its implicit condition (or as its implicit additional condition, in case the weight of the formula is already in conditional form \verb§[p|c]§).\\

Despite their convenience, specifiation of conditional probabilities in background knowledge are computationally quite expensive. A computationally cheaper approach supported by \prasp (and certain other PLP frameworks, such as ProbLog) is to use so-called \textit{Annotated Disjunctions} \cite{annotatedDisjunctions}, see Sect. \ref{annotatedDisjunctions} for details. But for now, we just emulate the effect of Annotated Disjunctions manually, by replacing \begin{verbatim}
[0.8|win] happy. 
:- happy, not win.
\end{verbatim} 
\noindent with 
\begin{verbatim}
[0.8] h.
happy :- win, h.
\end{verbatim} 
The last two lines correspond to annotated disjunction \verb§0.8::happy :- win§ in ProbLog syntax - the annotation \verb§0.8::§ refers to the head of the rule (happy), not to the entire rule as in \prasp. We will later introduce a more convenient syntax for annotated disjunction.\\
Note that their semantics is not the same as that of the statements above with the conditional probability.\\ 

Observe that \prasp does by default not assume any specific prior probability distribution over uncertain atoms or clauses whose probabilities are not specified by the user (or deducible using rules, etc).\\
For formulas which appear nowhere in a rule head, inference computes, in accordance with stable world semantics, probability 0, i.e., it makes a closed-world assumption which is typical for logic programming. \\

\noindent Consider the following example background knowledge file, which consist only of a single line:

\begin{verbatim}
[0.6|friend(brad, janet)] influences(brad, janet).
\end{verbatim}

If we issue query \verb#[[?|friend(brad, janet)]] influences(brad, janet)# to test whether the inference result reflects the given background knowledge, the result is\\
\verb#[?|friend(brad,janet)] influences(brad,janet)#, which indicates that \prasp couldn't find a solution. The reason is that \verb#friend(brad, janet)# appears nowhere else, so its probability is computed as 0. To solve this issue, we could either add \verb#[.] friend(brad, janet)# to the background knowledge (see Section \ref{overview}) or use command-line option \hyperref[cmdline:fullspan]{\texttt{--fullspan}}. With either of these, the query results in the expected\\ \verb#[0.6|friend(brad,janet)] influences(brad,janet).#\\  

Let's take a look at another example \cite{wiki1} for inference using conditional probabilities which is traditionally modeled as a Bayesian network. The following \prasp program models the probability of a pedestrian being hit by a car if (s)he crosses the road at a crossing without paying attention to the traffic lights. The traffic lights are for the cross traffic (i.e., the cars), so it should be more likely to be hit in case the light is green rather than red or yellow.

\begin{verbatim}
1{red,yellow,green}1.

[0.99|red] not hit.
[0.01|red] hit.

[0.9|yellow] not hit.
[0.1|yellow] hit.

[0.2|green] not hit.
[0.8|green] hit.

[0.2] red.
[0.1] yellow.
[0.7] green.
\end{verbatim}

The first line expresses that the states red, yellow and green of the traffic lights are mutually exclusive. This line is important because otherwise \prasp would assume that they were probabilistically independent if seen as events (unless command-line flag \hyperref[cmdline:noautoindeps]{\texttt{--noautoindeps}} is used). \prasp doesn't make this assumption for all facts, but it does make it for those facts which do not depend on anything in a logical sense (see Sect. \ref{meta}).\\

Formula \verb#[0.01|red] hit# represents the conditional probability of being hit by a car given that the traffic light is red. (The negative \verb#[0.99|red] not hit# could be omitted, but we include it in order to stay close to the original example \cite{wiki1}.)\\

\verb|[0.2] red| etc. specify the unconditional (unconditioned) probabilities of the three states of the traffic light.\\

\prasp is now able to compute the marginal probabilities of being hit and of not being hit, using queries \verb|[?] hit| and \verb|[?] not hit|, respectively. The results are 0.572 and 0.428, respectively. Note that there is no need to explicitly provide the joint probability distribution of being hit/not hit and the three states of the traffic lights. E.g., $Pr(red \wedge \neg hit) = 0.198$ could be inferred by \prasp given the above knowledge alone.\\

We could easily converter the above background knowledge and queries into ASP syntax by replacing \verb|not| with \verb|:-| everywhere. We could then omit a FOL$\rightarrow$ASP converter software or use command-line option \hyperref[cmdline:folconv]{\texttt{--folconv none}} to speed up inference.\\

An important use case of conditional probabilities is the inference of $Pr(A|B)$ with knowledge of $Pr(B|A)$ and Bayes' theorem. E.g., given the example background knowledge above, we might be interested in $Pr(yellow|hit)$, that is, the probability of lights showing yellow when being hit by a car. \prasp computes this using query \verb#[?|hit] yellow#, which results in approximately 0.0175. We could get this result also by applying Bayes' theorem {\large {\Large $\frac{Pr(hit|yellow)Pr(yellow)}{Pr(hit)}$}}. \\

\noindent Conditional probabilities can also appear in queries (e.g.,\\
\verb#[?|coin1(heads) & coin2(heads)] win# in coin flipping), with the $c$ in $Pr(f|c)$ typically describing the observed evidence. \\

It should be stressed again that placing conditional probabilities in background knowledge is not the same as using uncertain rules. Various types of uncertain rule-like constructs exist, each with a meaning different from the others. E.g., the semantics of\\

\verb§[0.01|red] hit.§\\

\noindent is different from the semantics of, e.g.,\\

\verb§[0.01] hit :- red.§\\

\noindent ...which is in turn semantically different from \\

\verb§[0.01] w.§\\
\indent \verb§hit :- red, w.§\\

or from \textit{Annotated Disjunctions} (Sect. \ref{annotatedDisjunctions}). Further forms occur in connection with non-ground formulas and double- and triple-square brackets in annotations (more about these later in this document).\\

Remark: Conditional probabilities in background knowledge don't harmonize well with the use of command-line switch \hyperref[cmdline:mod1]{\texttt{--mod1}} (see Sect. \ref{performance}). This is because for principled reasons, a conditional probability cannot be used to generate a spanning formula (Sections \ref{semantics},\ref{spanGen}) which is sufficient for a full dependency analysis (\prasp shows a warning message in this case).\\
For conditional probabilities in query files (where they are typically more useful) there is no such issue.

\subsubsection{Another example: The Monty Hall problem}

As the final example for working with conditional probabilities in background knowledge, we use \prasp to model the (in-)famous \textit{Monty Hall Problem} using conditional probabilities, following \cite{MontyHall}. This example also introduces variables and independence declarations (which will be covered in more detail in later sections). We provide two approaches. One possibility to model Monty Hall using \prasp is as follows:

\begin{verbatim}
door(1..3).

#domain door(I).
#domain door(J).
#domain door(K).

#indep
[[0.33333333]] c(I).
#endIndep

[[0]] condPr(h(I,J), c(K)) :- I == J.

[[0]] condPr(h(I,J),c(K)) :- J == K.

[[0.5]] condPr(h(I,J), c(K)) :- I == K, J != K.

[[1]] condPr(h(I,J), c(K)) :- I != K, J != K, J != I.
\end{verbatim}

In this scenario (named after game show host Monty Hall), which is famous for misleading many people to give a wrong answer, a game show contestant faces the following question: ``Suppose you're on a game show, and you're given the choice of three doors: Behind one door is a car; behind the others, goats. You pick a door, say No. 1, and the host, who knows what's behind the doors, opens another door, say No. 3, which has a goat. He then says to you, 'Do you want to pick door No. 2?' Is it to your advantage to switch your choice?'' (cited after \cite{MontyHallWiki}). The correct yet unintuitive answer is ``Yes, I should switch to the other door'', as it doubles the chance of winning the car from 1/3 to 2/3. For a detailed explanation please refer to \cite{MontyHallWiki}.\\

In the code above, \verb§h(x,y)§ stands for the event ``Contestant selected door x first. Game show host selected door y''. \verb§c(x)§ denotes that the car is behind door x.\\

\verb§condPr(f, c)§ represents $Pr(f|c)$ and is used to represent conditional probabilities as heads of rules (which is different from the semantics of \verb§[p|c]§ or \verb§[[p|c]]§ annotations). Variables and the double-square syntax for weights will be discussed in detail in a later section - in short, each formula annotated with such a weight is expanded into a set of ground formulas each with the respective weight and variables (like \verb§I§, \verb§J§ and \verb§K§ above) replaced with their instances from their respective domains (which are in our example the door numbers 1,2,3) in all possible combinations. The boolean conditions in the rule bodies (e.g., \verb§I == K, J != K§) are resolved during that grounding and expansion process and vanish afterwards.\\

Remark: \prasp (respectively the ASP grounder) doesn't always require that variable domains are declared globally, we could in the example above likewise add \verb§door(I)§ etc directly to the respective rules instead. \\

\noindent E.g., \verb§[[0.5]] condPr(h(I,J), c(K)) :- I == K, J != K§ is expanded to\footnote{Double-square brackets do not generally expand to facts, they can likewise expand to a set of ground rules (if the grounder is not able to simplify away the rule bodies).} 

\begin{verbatim}
[0.5|c(1)] h(1,2).
[0.5|c(1)] h(1,3).
[0.5|c(2)] h(2,1).
[0.5|c(2)] h(2,3).
[0.5|c(3)] h(3,1).
[0.5|c(3)] h(3,2).
\end{verbatim}

The conditional probabilities model the different choices the game show host can make:\\
He never opens the door selected by the contestant:\\
\verb§[[0]] condPr(h(I,J), c(K)) :- I == J§ (which denotes $Pr(h_{i,j}|c_k)=0 \mathrm{\ where\ } i=j$).\\
He never opens the door with the car:\\
\verb§[[0]] condPr(h(I,J),c(K)) :- J == K§.\\
If the contestant initially selected the door with the car, the host opens any of the two goat-doors with equal probability:\\
\verb§[[0.5]] condPr(h(I,J), c(K)) :- I == K, J != K§.\\
If the contestant initially chose one of the goat-doors, the host always opens the remaining goat door:\\
\verb§[[1]] condPr(h(I,J), c(K)) :- I != K, J != K, J != I§.\\

If we assume without loss of generality that the contestant initially selected door 1 and the host opens door 3, 
we can ask \prasp for the probability of winning the car by switching to door 2. The result of query \verb§[?|h(1,3)] c(2)§ is approximately 0.66. On the other hand, if the contestant sticks to their original choice of door 1, the winning probability is just \verb§[0.33|h(1,3)] c(1)§.\\

Note that we haven't specified all existing independence-relations among events (the initial door selection is independent of the number of the door behind which the car stands). Thus \prasp would be ``free'' to make any assumption about how the contestant makes their initial choice. To obtain a more determined result, we call \prasp with switch \hyperref[cmdline:maxentropy]{\texttt{--maxentropy}}.\\

Remember that a formula of form \verb§[w|c] f § \textit{alone} in background knowledge doesn't generate models in which $c$ hold. This is correct behavior, but, as we have seen already, it can lead to unexpected results if not observed.\\

\noindent Here's an alternative formalization of the Monty Hall problem:

\begin{verbatim}
#indep
[0.333] c1.
[0.333] c2.
[0.333] c3.
#endIndep

[0.333|x1] c1.
[0.333|x1] c2.
[0.333|x1] c3.

[0.5|2{c1,x1}2] h3.
[1|2{c2,x1}2] h3.
[0|2{c3,x1}2] h3.
\end{verbatim}

\verb§ci§ stands for ``Car is behind door i'', \verb§x1§ stands for ``contestant selected door 1'' and \verb§h3§ stands for ``Monty opened door 3''. The three formulas \verb§[0.333|x1] ci§ express that \verb§x1§ is independent from the \verb§ci§ (we could alternatively provide \verb§#indep§-meta-statements).\\

However, if we run inference with query \verb§[?|2{h3,x1}2] c2§, \prasp fails to find a solution. The reason is that we only stated that \textit{if} \verb§x1§ occurs, the probability of \verb§c1§, etc, is 0.333, but we said nowhere that \verb§x1§ can actually occur in any possible worlds! \\

To solve this problem, we simply add \verb§[.] x1§ to the program (an abbreviation of \verb§x1 | not x1§). (Adding \verb§[0.333] x1§ would in principle fulfill the same purpose, however, if the chosen solving method is easily disturbed by numerical inaccuracies, \verb§[.]§ provides more flexibility as it doesn't commit us to a certain probability.)\\

With this, we retrieve the expected solution \verb§[0.66|2{h3,x1}2] c2§ (i.e., the contestant wins with probability 0.66 if (s)he switches from door 1 to door 2).\\

The examples above show that the ability to put conditional probabilities in background knowledge provides a powerful modeling means. However, placing conditional probabilities in background knowledge slows down inference (and learning). So-called \textit{Annotated Disjunctions}, which we investigate in the next section, provide an alternative.

\subsection{Annotated Disjunctions}
\label{annotatedDisjunctions}

\prasp supports so-called \textit{Annotated Disjunctions} (ADs) \cite{annotatedDisjunctions}. They provide an intuitive and computationally relatively cheap approach (compared to constructs which are computationally more expensive in \prasp, such as conditional probabilities) to uncertain disjunctive clauses.\\

E.g., \verb§[0.7] heads(coin); [0.3] tails(coin) ::- tossed(coin), biased(coin)§ models a biased coin using an AD. Either ``heads'' or ``tails'' comes up with the respective given probability (but never both), if we tossed the coin and it is biased. \\

\noindent Observe that we use \verb§::-§ in ADs instead of \verb§:-§.\\

\prasp translates ADs into plain non-probabilistic rules and probabilistic facts according to the scheme presented in \cite{annotDisjProblog}. Observe that the weight notation within ADs uses single square brackets even though the denotation (which \prasp shows you with command-line switch \hyperref[cmdline:showexpansion]{\texttt{--showexpansion}}) is actually closer to the semantics of weights with double-square notion.\\

\noindent Restriction: weights within AD clauses must be ground (numerical literals) and non-conditional.\\

\noindent For more details about ADs, please refer to \cite{annotatedDisjunctions}. \\

\noindent Another example (from \cite{ad2}):

\begin{verbatim}
person(david).
person(jennifer).

#domain person(X).

[0.3] strongSneezing(X); [0.5] moderateSneezing(X) ::- flu(X).  

[0.2] strongSneezing(X); [0.6] moderateSneezing(X) ::- hayFever(X). 

flu(david).
hayFever(david).
flu(jennifer).
\end{verbatim}

\noindent Querying this background knowledge, we get, e.g.,
\verb§[0.8] moderateSneezing(david)§.\\ 

Remark: we can dramatically reduce computation time here by providing command-line switch \hyperref[cmdline:noautoindeps]{\texttt{--noautoindeps}}, however, the result is slightly less accurate then (0.799 instead of 0.8). The reason for this effect is that the above ADs generate a number of probabilistic facts which are mutually independent from each other. \\

Even better, we could provide \hyperref[cmdline:mod1]{\texttt{--mod1}} instead - this simplifies the program (subject to the respective query) while preserving accuracy. E.g., if we are only interested in David's sneezes in our query, \hyperref[cmdline:mod1]{\texttt{--mod1}} removes all statements from the grounded background knowledge which refer to Jennifer.

\subsection{Probability intervals\index{Probability intervals} as weights} 
\label{intervalweights}

Formulas in background knowledge can be annotated with intervals rather than point (precise) probabilities, allowing for imprecise probabilities (for background and alternative approaches, see, e.g., \cite{weichsel,cozman2}).\\

\noindent E.g., \verb§[0.5;0.7] coin(heads).§\\
specified that the probability of \verb§coin(heads)§ is between $0.5$ and $0.7$. Observe that \prasp uses a semicolon for intervals but commas in lists of point probabilities.\\

We can also instruct \prasp to \textit{obtain} interval \textit{results} for queries (facilitated using switch \hyperref[cmdline:intervalresults]{\texttt{--intervalresults}}, see Sect. \ref{intervalresults} and \ref{commandline}). Both features can be used together or independently from each other.\\

\noindent {\color{Black} More tbw. }

\subsection{Event independence}
\label{independence}

In contrast to many other probabilistic logics, \prasp does - by default - not assume independence of probabilistic events. Also, in contrast to Bayesian Networks, by default, it doesn't assume any conditional independence deduced from a graph (or rule) structure. Instead, it assumes that independence/dependence between events which are not explicitly related by rules or other constraints in the background knowledge is \textit{unspecified}. E.g., if (in default mode) the background knowledge merely specifies for each of two coins the individual probability 0.5 of the coin coming up with heads but doesn't explicitly specify that these two coins are independent from each other, the range of possible query results for query $coin1(heads) \wedge coin2(heads)$ reflects any possible dependence or independence between the two coins (e.g., probability 0 (``heads'' are mutually exclusive), 0.25 (independence) and probability 0.5 (both coins are magically connected and behave the same)). The user can either explicitly specify that both coins shall be independent (see Section \ref{meta}) or advice \prasp to automatically assume independence if there is no explicit dependence specified in background knowledge (see \ref{commandline}). In both cases, the result for $coin1(heads) \wedge coin2(heads)$ is then 0.25.\\

\noindent Event independence is closely related to the following topic:

\subsection{Point, sample, maximum entropy and interval solutions}
\label{intervalresults}

Using default settings, \prasp computes a single point probability per query formula per query file. However, if the uncertain background theory is consistent but underspecified (degrees of freedom remain - the typical case), an infinite number of further solutions exist. In that case, \prasp returns one of these solutions per query and ignores the others. Using switch \hyperref[cmdline:maxentropy]{\texttt{--maxentropy}}, you can instruct \prasp generate a probability distribution with higher than default or maximum entropy (depending on the inference method) to compute the query solutions. This follows the so-called \textit{Principle of Maximum Entropy} \cite{james}, which states that among all allowed distributions, the one with the largest entropy best represents the current state of knowledge (here: our knowledge stated in the \verb§.prasp§ file). (The so-called \textit{Principle of Indifference}, observed e.g. by \cite{plog}, is a special case of this where the uninformative prior is the uniform distribution.)\\
%Principle of indifference

\noindent Note: \hyperref[cmdline:maxentropy]{\texttt{--maxentropy}} is experimental, its algorithms are subject to change in later versions of \prasp.\\

%Important: a high entropy of some probability distribution over of sampled set of possible worlds alone does not guarantee a ``good'' solution (low information bias), if this set is not sampled uniformly from the total set of possible worlds.\\

With the default inference approach, \prasp attempts to find a high entropy distribution using iterative \textit{information projection} from the uniform distribution. Optionally, the desired precision of the search process can be specified, using a positive number $0 <$ \verb§prec§ $< 1$ after \hyperref[cmdline:maxentropy]{\texttt{--maxentropy}}, e.g., \hyperref[cmdline:maxentropy]{\texttt{--maxentropy 1E-8}}. Lower \verb§prec§ means higher precision.  \verb§prec§ is the threshold of the value of $\Arrowvert-\bigtriangleup(\mathit{CD} \Arrowvert U)\Arrowvert_2$ for which the maximum entropy distribution search stops. $\mathit{CD} \Arrowvert U$ denotes the Kullback\verb§-§Leibler divergence between the current candidate distribution and the uniform distribution.\\

With simulated annealing (if also \hyperref[cmdline:nosolve]{\texttt{--nosolve}} is given), entropy maximization is not fully supported, however, if \hyperref[cmdline:maxentropy]{\texttt{--maxentropy}} is provided and the \verb§initPhase§ argument of \hyperref[cmdline:simanneal]{\texttt{--simanneal}} is 9, an initial probability distribution is used for preconditioning the simulated annealing process which fully accounts for weighted formula independence (limited only by \hyperref[cmdline:limitindepcombs]{\texttt{--limitindepcombs}}). The full consideration of event independence is a requirement for maximum entropy. (The optional parameters of \hyperref[cmdline:maxentropy]{\texttt{--maxentropy}} are ignored here.) The same effect could in theory be achieved by omitting \hyperref[cmdline:noindepconstrs]{\texttt{--noindepconstrs}} with \hyperref[cmdline:simanneal]{\texttt{--simanneal}}, however, this would be computationally a lot more costly. Furthermore, \hyperref[cmdline:simanneal]{\texttt{--simanneal}} with argument \verb§initPhase§ being 1 assumes all weighted formulas to be mutually independent (regardless of actual or declared independence). \\
In the quite unusual case that \hyperref[cmdline:simanneal ]{\texttt{--simanneal }}is specified but \textit{not} \hyperref[cmdline:nosolve]{\texttt{--nosolve}}, \hyperref[cmdline:maxentropy]{\texttt{--maxentropy}} has the same effect as in the case of default inference (using linear system solving), but the set of models (possible worlds) is the set sampled using simulated annealing, not the total set of models (limited by \hyperref[cmdline:models]{\texttt{--models}}) as without \hyperref[cmdline:simanneal]{\texttt{--simanneal}}. \\

With default inference and without switch \hyperref[cmdline:maxentropy]{\texttt{--maxentropy}}, \prasp computes a small number of random candidate distributions and chooses the one(s) with the largest entropies among these. This computationally comparatively lean approach does of course not guarantee that a maximum entropy distribution is used but provides ``better than worst'' distributions (where ``worst'' means a distribution with lowest entropy).\\

Use switch \hyperref[cmdline:ignoreentropy]{\texttt{--ignoreentropy}} to make \prasp compute distribution(s) without considering their entropies. This can  be useful: 1) to obtain a somewhat faster result or 2) to compute a number of randomly sampled probability distributions in connection with \hyperref[cmdline:ndistrs n]{\texttt{--ndistrs n}}, in order to estimate the range of possible solutions (as an alternative to \hyperref[cmdline:intervalresults]{\texttt{--intervalresults}}).\\

Generally, \hyperref[cmdline:ndistrs n]{\texttt{--ndistrs n}} asks \prasp to compute a specific number of sample solutions, but there is no way other than \hyperref[cmdline:ignoreentropy]{\texttt{--ignoreentropy}} to influence the distribution of the sampled distributions. \hyperref[cmdline:ndistrs]{\texttt{--ndistrs}} works only with the default inference method.\\  

Try also \hyperref[cmdline:itrefinement]{\texttt{--itrefinement}}, or \hyperref[cmdline:nosolve]{\texttt{--nosolve}} alone (without \hyperref[cmdline:simanneal]{\texttt{--simanneal}}) using the weighted flip-sampling algorithm (where applicable) with \hyperref[cmdline:initsample]{\texttt{--initsample}}, in order to achieve a high entropy.\\

It is also possible to let \prasp compute the full range of possible query solutions: with switch \hyperref[cmdline:intervalresults]{\texttt{--intervalresults}}, you ask \prasp for an interval of probabilities for each query formula. The returned intervals are exhaustive, i.e., they cover the entire sets of solutions per query (unless \hyperref[cmdline:simanneal]{\texttt{--simanneal}} is specified. In that case the resulting interval denotes a subset of the full interval). Alternatively,  \hyperref[cmdline:ndistrs n]{\texttt{--ndistrs n}} can be used (with or without \hyperref[cmdline:ignoreentropy]{\texttt{--ignoreentropy}} or \hyperref[cmdline:maxentropy]{\texttt{--maxentropy}}), but this does not guarantee any particular extend of the returned solution ranges.\\

\noindent Example:

\begin{verbatim}
[0.5] coin_out(1,heads).
[0.5] coin_out(2,heads).

1{coin_out(1,heads), coin_out(1,tails)}1. 
1{coin_out(2,heads), coin_out(2,tails)}1. 

win :- coin_out(1,heads), coin_out(2,heads).
\end{verbatim}

If we call \prasp with default settings (i.e., no command-line switches), it assumes that the relation between the two coins is not specified - they might be either independent or their outcomes might be correlated in some hidden way. Consequentially, there are infinitely many valid results for \verb§Pr(win)§, namely all real numbers between 0 (the two coins are inversely correlated and never come up with ``heads'' simultaneously) and 0.5 (either both coins come up with ``heads'' or both come up with ``tails''). From this range, \prasp picks 0.25. This is the point solution obtained from the probability distribution over possible worlds which has the highest entropy, however, without switch \hyperref[cmdline:maxentropy]{\texttt{--maxentropy}}, \prasp does not try very hard to maximize the entropy and might as well return a point solution which carries some random information bias. \\
Note that ``entropy'' always refers to the entropy of the underlying possible world probability distribution used to solve all queries in a particular query file; there is no concept of individual query formula ``entropy'' in \prasp.\\

To see the full range of valid answers, call \prasp with switch \hyperref[cmdline:intervalresults]{\texttt{--intervalresults}}. \prasp now returns \verb§[0;0.5] win§, i.e., $Pr(win) \in [0..0.5]$.\\

\noindent If we surround  
\begin{verbatim}
[0.5] coin_out(1,heads).
[0.5] coin_out(2,heads).
\end{verbatim}
with \verb§#indep§ ... \verb§#endIndep§ (see \ref*{indep}), the interval result for win becomes\\
\verb§[0.25;0.25] win§, since now we know $Pr(\mathit{coin\_out(1,heads)} \wedge \mathit{coin\_out(2,heads)}) = $\\
$Pr(\mathit{coin\_out(1,heads)})Pr(\mathit{coin\_out(2,heads)}) = 0.25$.\\

\hyperref[cmdline:intervalresults]{\texttt{--intervalresults}} can be used with all \prasp programs and queries, and in combination with most other switches and with intervals as given weights (Sect. \ref{intervalweights}). However, inference results which occur internally during parameter learning (Section \ref*{paramlearn}) are always reduced to point probabilities (more precisely: the mean values of the intervals).\\

\noindent \underline{Important}: An interval result only makes sense considering the respective query formula  in isolation. Values within the interval are not necessarily consistent with all values in intervals returned for other query formulas in the same query file. In detail: each interval is computed using a \textit{different} probability distribution over possible worlds (namely those distributions which minimize (respectively maximize) the probability of the respective query formula). This means that a point probability picked from a certain interval is \textit{not} necessarily consistent with point probabilities picked from other intervals. In other words, it is not in general possible to derive a consistent \prasp program from the set of query results computed with \hyperref[cmdline:intervalresults]{\texttt{--intervalresults}}.\\

\hyperref[cmdline:intervalresults]{\texttt{--intervalresults}} also works with formulas and queries with conditional probabilities. This has a subtle consequence: if there are any conditional queries (and \hyperref[cmdline:simanneal]{\texttt{--simanneal}} is not provided also), \prasp automatically adds constraints which specify that the probabilities of all conditions in the queries need to be larger than zero (\prasp computes conditional queries using $\frac{Pr(f \wedge c)}{Pr(c)}$ and without these additional constraints, the linear optimizer which is used to find appropriate probability distributions might find distributions which result in $\frac{x}{0}$ even though a conditional probability exists). \\

\hyperref[cmdline:intervalresults]{\texttt{--intervalresults}} together with \hyperref[cmdline:simanneal]{\texttt{--simanneal}} can lead to intervals which are non-exhaustive (i.e., sub-intervals of the widest intervals). This is caused by the fact that \hyperref[cmdline:simanneal]{\texttt{--simanneal}} searches for lower and upper bounds of query probabilities using a simple (meta-)sampling algorithm which assumes that interval boundaries are uniformly distributed. Future versions of \prasp might behave better in this regard.

\subsection{Working with non-ground and first-order formulas} 
\label{nonground}

\prasp supports ASP-style as well as FOL-style variables and quantifiers. A formula (program) which doesn't contain any variables is called a \textit{ground} formula (program). The set of ground formulas which are gained by instantiating the variables within a non-ground formula with all possible combinations of values from the respective variable domains are called the \textit{instances} of this non-ground formula.\\

As a first simple example for background knowledge with non-ground formulas, consider the following \prasp program:

\begin{verbatim}
p(1).
p(2).
p(3).
#domain p(X).

#pIndep
[0.5] v(1). 
[0.5] v(2).
[0.5] v(3).
#endPIndep

[0.1] v(X).
\end{verbatim}

The weight 0.1 in the last line refers to formula \verb|v(X)| \textit{as a whole}, i.e., this line is \underline{not} syntactic sugar for the three formulas \verb|[0.1] v(1). [0.1] v(2). [0.1] v(3).| but instead stands for $Pr(v(1) \wedge  v(2) \wedge v(3)) = 0.1$ (note that this does not preclude that the probability of\verb§ v(1)§ could be 0.1, provided the lines \verb|[0.5] v(1).| would be omitted). \\

The alternative semantics where a single given weight is individually assigned (distributed) to each ground instance of some annotated non-ground formula is also supported by \prasp (using \verb§[[...]]§ for weights), it will be introduced later in this section.\\

\noindent The \verb§#pIndep§ section specifies that \verb§v(1)§, \verb§v(2)§, \verb§v(3)§ are pairwise independent.\\

The \verb§#domain§ meta-statement is optional here. Alternatively, we could declare the variable domain locally, by writing \verb§[0.1] v(X) :- p(X)§ instead of \verb§[0.1] v(X)§. However, in some other cases, \verb§#domain§ declarations prevent that grounding of formulas fails due to so-called ``unsafe variables''. \verb§#domain§ doesn't work with Gringo/Clingo4, in which case you need to bind any variable using atoms (with domain predicates) in the rule body, see Gringo/Clingo manual.\\

\noindent Using the above background knowledge, the following query file:

\begin{verbatim}
[?] v(X).

#domain p(Z).

[?] ![Z]: v(Z).
[?] ?[Z]: v(Z).

[?] v(1).
[?] v(2).
[?] v(3).

[?] v(1) & v(2).
[?] v(1) & v(2) & v(3).

[?] :- v(X).
\end{verbatim}

\noindent ...give us these results: 

\begin{verbatim}
[0.1] v(X).
[0.1] ![Z]: v(Z).
[0.85] ?[Z]: v(Z).
[0.5] v(1).
[0.5] v(2).
[0.5] v(3).
[0.25] v(1) & v(2).
[0.1] v(1) & v(2) & v(3).
[0.15] :- v(X).
\end{verbatim}

The result of query \verb|[?] ![Z]: v(Z)| is $Pr(\forall z.v(z)) = 0.1$, which is also the result of the semantically equivalent queries \verb|[?] v(1) & v(2) & v(3)| and \verb|[?] v(X)|. This unconditional probability was directly given as weight in the background knowledge. In contrast to \verb|X|, variable \verb|Z| is a variable in the sense of first-order logic and bound by universal quantifier \verb|![Z]:| (at this, remember that \prasp requires the domains of all variables to be finite). \\

Of course, weighted formulas with existential or universal quantifiers can also be used in background knowledge.\\

The result of \verb|?[Z]: v(Z)| is $Pr(\exists z.v(z))$ (where \verb|?[Z]:| represents an existential quantifier) and could likewise be calculated manually using the inclusion-exclusion principle as $Pr(v(1)\vee v(2) \vee v(3)) = Pr(v(1)) + Pr(v(2)) + Pr(v(3)) - Pr(v(1) \wedge v(2)) - Pr(v(1) \wedge v(3)) - Pr(v(2) \wedge v(3)) + Pr(v(1) \wedge v(2) \wedge v(3)) = 0.85$.\\

It is important to note that in this example, the result of \verb|[?] :- v(X)| is 0.15 and not 0.9, even though the result of \verb|[?] v(X)| is 0.1. The reason is that in ASP, rule \verb|:- v(X)| corresponds to $not\ v(1) \wedge not\ v(2) \wedge not\ v(3)$. 
To obtain 0.9, i.e., $Pr(not\ (v(1) \wedge v(2) \wedge v(3)))$, we would need to use, for example, query
\verb#[?] not (v(1) & v(2) & v(3))# or \verb#[?] not ![X]: v(X)#.\\

Note that it is not necessary to (re-)declare variable \verb|X|, since \prasp evaluates query formulas in the context of the background knowledge file, where the domain of \verb|X| is already defined, as well as predicate \verb|p/1|. However, we need to declare the new variable \verb|Z| in the query file (using \verb|#domain p(Z)|). \\

There is one exception from the rule that variables need to be declared only in the background knowledge: if you put a query with double-square brackets \verb|[[...]]| in the query file (see below), you need to put also all declarations required to ground the query formula into the query file in case the formula contains global ASP-style variables, even if they are already found in the background knowledge file (this even allows to declare different domains for global variables with the same name, although this would be hardly sensible). \\

\noindent Let's see now what happens if we remove the following lines

\begin{verbatim}
#pIndep
[0.5] v(1).
[0.5] v(2).
[0.5] v(3).
#endPIndep
\end{verbatim}

from background knowledge: the effect is that all inferred probabilities except for the last one become $0.1$, and the probability of \verb|:- v(X)| becomes 0.9. The reason is that \prasp uses by default precisely the answer sets of the background knowledge as possible worlds (and not the set of all models resulting from all combinations of literals in the program)\footnote{This behavior can be changed, see command-line option \texttt{--fullspan} in Section \ref{commandline}.}. The only answer sets are now \verb|{p(1),p(2),p(3)}| and
\verb|{v(1),v(2),v(3),p(3),p(2),p(1)}|. In other words, the three propositions \verb|v(i)| become dependent of each other and appear either all together or not at all. Adding a formula such as \verb|1{v(Z):p(Z)}1.| (or reinserting the removed three weighted formulas) would create separate possible world for each of the \verb|v(i)| again. Another way to create different possible worlds for each ground instance is to use the \verb|[[...]]| or \verb|[[[...]]]| syntax, as explained below, or experimental command-line switch \texttt{--fullspan} (but using the latter is not recommended and leads to a very large number of answer sets). 

\subsection{Assigning weight annotations to ground instances of non-ground formulas} 

Sometimes the aforementioned treatment of annotated formulas with variables is not desirable, namely if we want a weight to denote the probability of \textit{each} of the ground instances of a given non-ground formula (i.e., a formula which contains ASP-style variables). This can be achieved by using \verb|[[weight]]| instead of \verb|[weight]| as annotation. Consider the following example which generalizes the coin-example from Section \ref{basic} for any number of coins, using ASP-style variables:

\begin{verbatim}
coin(1..3).

#indep
[0.6] coin_out(1,heads).
[[0.5]] coin_out(N,heads) :- coin(N), N != 1.
#endIndep

1{coin_out(N,heads), coin_out(N,tails)}1 :- coin(N).

n_win :- coin_out(N,tails), coin(N).
win :- not n_win. 
\end{verbatim}

The first line specifies that there are three coins. Therefore, in the rest of the program, variable \verb|N| can take 
values 1, 2, or 3. \verb|[0.6] coin_out(1,heads)| denotes that the first coin is biased: the probability of landing on its head is 0.6. Line \verb|[[0.5]] coin_out(N,heads) :- coin(N), N != 1| corresponds\footnote{More precisely, this line is expanded to the two formulas \texttt{[0.5] coin\_out(2,heads) :- coin(2), 2 != 1} and \texttt{[0.5] coin\_out(3,heads) :- coin(3), 3 != 1}, i.e., the original line is actually not syntactic sugar for uncertain facts but for uncertain rules. As a non-rule alternative, consider using so-called conditionals, as in \texttt{[[0.5]] coin\_out(N,heads) : coin(N).}} to the following two weighted ground formulas, replacing variable \verb|N| with its instances:

\begin{verbatim}
[0.5] coin_out(2,heads).
[0.5] coin_out(3,heads).
\end{verbatim}

This specifies that all other coins (i.e., coins number 2 and 3) behave unbiased: their heads-probability is 0.5.\\

Query \verb|[?] win| results in \verb|[0.15] win|, which reflects that, assuming the three coins are uncorrelated: $Pr(win) = Pr(coin1=heads \wedge coin2=heads \wedge coin3=heads) = Pr(coin1=heads) * Pr(coin2=heads) * Pr(coin3=heads) = 0.6 * 0.5 * 0.5 = 0.15$.\\

\noindent Another example: replacing in the background knowledge of the first example in Sect. \ref{nonground} the following 

\begin{verbatim}
#pIndep
[0.5] v(1).
[0.5] v(2).
[0.5] v(3).
#endPIndep

[0.1] v(X).
\end{verbatim}

\noindent with: 

\begin{verbatim}
#pIndep
[[0.5]] v(X).
#endPIndep
\end{verbatim}

\noindent gives us the following new inference results:

\begin{verbatim}
[0.125] v(X).
[0.125] ![Z]: v(Z).
[0.875] ?[Z]: v(Z).
[0.5] v(1).
[0.5] v(2).
[0.5] v(3).
[0.25] v(1) & v(2).
[0.125] v(1) & v(2) & v(3).
[0.125] :- v(X).
\end{verbatim}

\noindent ...because in the modified background knowledge,

\begin{verbatim}
[[0.5]] v(X).
\end{verbatim}

\noindent is an abbreviation (``syntactic sugar'') for

\begin{verbatim}
[0.5] v(1).
[0.5] v(2).
[0.5] v(3).
\end{verbatim}

...and because we have removed the assignment of probability 0.1 to \verb|v(X)| (which corresponded to $Pr(v(1) \wedge v(2) \wedge v(3)) = 0.1$).\\

\verb|[[|...\verb|]]| also works with formulas with multiple globally and/or locally defined variables. E.g., the following \prasp program:

\begin{verbatim}
q(a).
q(b).
q(c).
#domain q(X).

number(1..5).

% definition of predicates w and u omitted

[[0.5]] v(X) :- 1{w(X),u(N) : number(N)}3.
\end{verbatim}

\noindent ...expands to:

\begin{verbatim}
...
[0.5] v(a):-1#count{w(a),u(5),u(4),u(3),u(2),u(1)}3.
[0.5] v(b):-1#count{w(b),u(5),u(4),u(3),u(2),u(1)}3.
[0.5] v(c):-1#count{w(c),u(5),u(4),u(3),u(2),u(1)}3.
\end{verbatim}

\noindent (using Gringo 3 as grounder).\\

Variable domains can be declared globally but also locally. But note that locally declared variables within rule heads, such as in \verb#[[w]] v(X) : d(X)#, lead to an expansion to, e.g., \verb#[w] v(1)|v(2)|v(3)| ...# (assuming that \verb|d(1)|, \verb|d(2)|, \verb|d(3)|, ... hold). Use  \verb#[[w]] v(X) :- d(X).# instead if you would like to generate the following expansion:
\begin{verbatim}
[0.6] v(1).
[0.6] v(2).
[0.6] v(3).
...
\end{verbatim}

\verb|[[|...\verb|]]| can be used for the annotation of query formulas also. A query of the form
\verb|[[?]] f| (line end dot as usual omitted) expands to a set of queries \verb|[?] f|$_i$, where the \verb|f|$_i$ are the ground instances of formula \verb|f|.\\

\noindent Double-square brackets can also be used for the specification of conditional probabilities in background knowledge  (using syntax form \verb#[[p|cond]] f#) as well as for inference of conditional probabilities, using syntax form \verb#[[?|cond]] f.# Here, the expansion creates a set of weighted conditional formulas \verb#[p|ci]] fi# where each pairing \verb§ci§, \verb§fi§ is a ground instance of \verb§cond§ or \verb§f§, respectively, and where a certain \verb§ci§ is paired with a certain \verb§fi§ \textit{iff} the grounder emitted both in the same position in the list of ground instance (for further details please see Sect. \ref{weights}).\\

\noindent Example:

\begin{verbatim}
q(a).
q(b).
q(c).
#domain q(X).

[[0.2|w(X)]] v(X).
\end{verbatim}

\noindent ...expands to:

\begin{verbatim}
[0.2|w(a)] v(a).
[0.2|w(b)] v(b).
[0.2|w(c)] v(c).
\end{verbatim}

However, as already pointed out, in contrast to queries annotated with \verb|[|...\verb|]|, the use of \verb|[[|...\verb|]]| (or \verb|[[[|...\verb|]]]|, see below) within a query file requires that any global variable domain declaration needed for the grounding of such query formulas is found in the query file \textit{itself} (at a position in the file before that of the respective query formula). E.g., in the following \textit{query} file:

\begin{verbatim}
q(a).
q(b).
#domain q(X).

[[?|w(X)]] v(X).
\end{verbatim}

\noindent ...the query formula expands to:

\begin{verbatim}
[?|w(a)] v(a).
[?|w(a)] v(b).
\end{verbatim}

\noindent ...even in case you have (purposefully or accidentally) specified a different domain for variable \verb|X| in the background knowledge!\\

\textit{Computed weights} (symbolic weights whose numerical values are computed by the grounder, cf. Sect. \ref{weights}) can only be used in background knowledge (.prasp-files).\\

Using \textit{triple}-square brackets allows to generate \textit{all} combinations of the ground instances of \verb§cond§ and \verb§f§. Note that this can quickly lead to a combinatorial explosion or possible worlds and is rarely useful.\\

\noindent Example:

\begin{verbatim}
q(a).
q(b).
q(c).
#domain q(X).

[[[0.2|w(X)]]] v(X).
\end{verbatim}

\noindent ...expands to:

\begin{verbatim}
[0.2|w(a)] v(a).
[0.2|w(b)] v(a).
[0.2|w(c)] v(a).
[0.2|w(a)] v(b).
[0.2|w(b)] v(b).
[0.2|w(c)] v(b).
[0.2|w(a)] v(c).
[0.2|w(b)] v(c).
[0.2|w(c)] v(c).
\end{verbatim}

Here's another, more realistic example which models a social network (an instance of the well-known ``smokers network'' example originally modeled using Markov Logic):

\begin{verbatim}
person(1).
person(2).
person(3).
person(4).		

#domain person(X).
#domain person(Y).

#pIndep
[[0.3]] stress(X).
[[0.2]] influences(X,Y).
#endPIndep

#gIndep
[0.4] ha.
[0.4] hb.
[0.4] hc.
[0.4] hd.
#endGIndep

smokes(X) :- stress(X).
smokes(X) :- friend(X,Y), influences(Y,X), smokes(Y).

asthma(1) :- smokes(1), ha.
asthma(2) :- smokes(2), hb.
asthma(3) :- smokes(3), hc.
asthma(4) :- smokes(4), hd.		

friend(1,2).
friend(2,1).
friend(2,4).
friend(3,2).
friend(4,2).
\end{verbatim} 		

Observe that for the performance-related reasons stated above, we have encoded 
the probabilistic rules that smoking causes asthma with probability $0.4$ in
ProbLog-style instead of using conditional probabilities (remark: since \prasp version 0.8, we could instead directly use annotated disjunction syntax for this, see Sect. \ref{annotatedDisjunctions}).\\

\noindent As query file, we use 

\begin{verbatim}
#def evidence = 2#count{smokes(2), not influences(4,2)}2

[?|evidence] smokes(1).
[?|evidence] smokes(2).
[?|evidence] smokes(3).
[?|evidence] smokes(4).

[?|evidence] asthma(1).
[?|evidence] asthma(2).
[?|evidence] asthma(3).
[?|evidence] asthma(4).
\end{verbatim} 	

\noindent We call \prasp using command-line arguments\\
\verb§ --folconv none --simanneal 0.0001 1e-300 0.9 --nosolve --sirndconf --noindepconstrs§\\

This means we don't need a FOL converter (since all formulas are plain ASP), we use simulated
annealing as approximate inference approach (because of the large number of possible worlds) and
omit the generation of explicit independence enforcement constraints (these are typically not suitable
with simulated annealing, see \ref{performance}). Observe that all query formulas are so-called
simple formulas \ref{simpleFormulas}, which additionally speeds up inference. The major factor
which determines inference speed is however the fact that all weighted formulas are mutually independent.
Inference with this example works therefore quite fast.\\

Note that the meta-statements \verb§#pIndep§ and \verb§#gIndep§ are not actually used with
the command-line options above - simulated annealing always makes the default assumption that
all uncertain events are independent (this is in absence of other knowledge a reasonable default
assumption even if events are actually not independent). However, if we would change the 
\hyperref[cmdline:simanneal]{\texttt{--simanneal}} arguments to \\
\hyperref[cmdline:simanneal 0.0001 1e-300 0.9]{\texttt{--simanneal 0.0001 1e-300 0.9}} , the algorithm would perform a sampling step 
at each iteration which actually uses the meta-information provided by \verb§#gIndep§ etc.
This would improve performance (only) in case we would add some non-independent weighted formulas
to the background knowledge.

\subsection{Probabilistic inference with mere \texttt{--nosolve}} 
\label{merecounting}
\label{choiceConstructs}

\subsubsection {Using unannotated knowledge}

We have already seen that probabilistic inference can also be performed with ordinary logic programs, that is, without annotated formulas (see, e.g., the Tweety example in Section \ref{basic}). Under certain circumstances, \prasp allows to perform probabilistic inference over unannotated knowledge also by means of pure frequency determination (boiling down to pure model counting), that is, by omitting the regular inference step. This is achieved by specifying command-line argument \verb#--nosolve# (this switch is not strictly necessary in case formula weights can be ignored, but speeds up computation if there are no given weight annotations in background knowledge). In contrast to annotating formulas explicitly with probabilities, we now identify probabilities with the relative frequencies of those answer sets in which the respective formulas are true. The constrain solving phase which normally uses the formula weights to compute possible world probabilities is omitted. However, this approach can not always be taken, as we explain below.\\

As the possibly most simple example, half of the answer sets of program \verb#a | b# support \verb#a# and the other half support \verb#b#, so that this program indirectly represents a probability of 0.5 for \verb#a# and 0.5 for \verb#b# (under the i.i.d. assumption regarding the truth values of \verb#a# and \verb#b#). As a slightly more complex example, consider the following background knowledge file:  

\begin{verbatim}
1 { h1, h2, h3 } 1.
p :- h1.
q :- not h1.
\end{verbatim}

Asking for the probabilities of \verb#p# and \verb#q# results in:

\begin{verbatim}
[0.3333333333333333] p.
[0.6666666666666666] q.
\end{verbatim}

Here, the rule \verb|p :- h1| effectively imposes a weight of $\oldstylenums{1} / \oldstylenums{3}$ on the uncertain fact \verb|p| (but note that this is of course no general scheme for ``weight emulation''). \\

It is also possible to use switch \verb#--nosolve# with \textit{weighted} background knowledge. In that case each weighted formula \verb#[w] f.# is treated as \verb#f | not f.# and the specific weight is ignored, unless either \hyperref[cmdline:initsample]{\texttt{--initsample}} or \verb#--weights2cc# is specified on the command-line. 

\subsubsection {Using weighted knowledge}
\label{merecountingwithweights}

In some cases, \prasp can perform inference with merely \hyperref[cmdline:initsample]{\texttt{--initsample}} and \hyperref[cmdline:nosolve]{\texttt{--nosolve}} even in the presence of weighted formulas in background knowledge. This form of inference can be highly efficient. Consider

\begin{verbatim}
coin(1..10).

[0.6] coin_out(1,heads).
[[0.5]] coin_out(N,heads) :- coin(N), N != 1.

1{coin_out(N,heads), coin_out(N,tails)}1 :- coin(N).

n_win :- coin_out(N,tails), coin(N).
win :- not n_win. 
\end{verbatim}

With command-line arguments \hyperref[cmdline:initsample 4 --nosolve --folconv none]{\texttt{--initsample 4 --nosolve --folconv none}}, \prasp assumes that the coins are mutually independent (regardless of independence declarations or automatically discovered independence) and samples from all possible non-uncertain instances of the \textit{spanning program} (Sections \ref{semantics},\ref{spanGen}). The result is a list of possible worlds (with repetitions) where the occurrence frequencies of possible worlds in the list reflect the weights of the annotated formulas which hold in the respective possible world. This is already sufficient for performing approximate inference. What makes this approach efficient is that neither linear programming nor simulated annealing or iterative refinement are required to solve the queries. Nevertheless, it is possible to combine it with one of these approaches, e.g., by adding \hyperref[cmdline:simanneal]{\texttt{--simanneal}} to the command-line. In that case, the samples computed using \hyperref[cmdline:initsample]{\texttt{--initsample}} serve as initial samples fed into the simulated annealing algorithm.\\

This approach, if used stand-alone (without subsequent universal inference step such as simulated annealing or iterative refinement), is comparatively fast but gives wrong results if events are actually not mutually independent. It also doesn't work if there are any conditional probabilities in background knowledge. However, conditional probabilities can still be used in queries.\\

Precision and computation time depend on number and quality of samples and can be influenced using \hyperref[cmdline:models]{\texttt{--models}}, \hyperref[cmdline:xorconf]{\texttt{--xorconf}}, \hyperref[cmdline:flipsampconf]{\texttt{--flipsampconf}} and \hyperref[cmdline:sirndconf]{\texttt{--sirndconf}}. \\

In principle, \hyperref[cmdline:nosolve]{\texttt{--nosolve}} (without \hyperref[cmdline:simanneal]{\texttt{--simanneal}}) can also be combined with any other form of initial sampling. {\color{Black} (More tbw.) }\\

Another experimental inference approach is letting \prasp convert probabilistic programs into ordinary answer set programs: this way, we use \textit{weighted} formulas in background knowledge without the need to solve an (possible computationally expensive) system of linear equality constraints in order
to obtain possible world probabilities. Instead, weighted formulas are converted into certain unweighted formulas 
which are conditioned on certain ASP choice constructs, such that unweighted model counting over possible worlds reflects 
the formula weights.

However, this feature is experimental and currently intended only for those cases where each weighted formula is probabilistically independent from
any other formula in the background knowledge (in other words, each weighted formula needs to represent an independent experiment). This can be verified using command-line option \hyperref[cmdline:showindeps]{\texttt{--showindeps}}.

\noindent As an example, consider the above coins game (with ten coins) again. All
coins are mutually independent, so we can safely use \hyperref[cmdline:weights2cc]{\texttt{--weights2cc}}\hyperref[cmdline:nosolve]{\texttt{ --nosolve}} with this example and get correct results (albeit after a certain delay):

\begin{verbatim}
[0.001171875] win.
[0.998828125] not win.
[0.6] coin_out(1,heads).
[0.5] coin_out(2,heads).
\end{verbatim}

Generally, \prasp does not impose any specific prior probability distribution on unweighted ground literals (in contrast to most other probabilistic logics). Instead, it chooses amongst multiple possible world probability distributions the one with maximum entropy and additionally imposes independence constraints on logically independent ground atoms. However, probabilities of literals and more complex formulas generated solely from such unweighted choice formulas or disjunctions like \verb#a | b# are determined entirely by the set of answer sets computed by the answer set solver - i.e. in this case 0.5 for \verb|a| and 0.5 for \verb|b|, since in the elements of one partition of the set of answer sets \verb|a| holds (but not \verb|b|) and in the other, equally large partition of answer sets \verb|b| holds (but not \verb|a|).

\subsection{Working with inconsistent background knowledge} 
\label{inconsistency}

Inconsistency means that one or both of the following is true: 1) the spanning program (Sections \ref{semantics},\ref{spanGen}) doesn't have models, i.e., there are no possible worlds (logical inconsistency), 2) given the weights in background knowledge, no probability distribution over the possible worlds exists. While inconsistency in the sense of 2) is formally an error condition, \prasp might still be able to obtain a result in that case - without any guarantees regarding the degree of correctness of this result. With the default settings, you get a warning "Specified probabilities appear to be inconsistent. Results might be inaccurate or wrong" in case PrASP cannot satisfy all constraints induced by the probabilities you provided in the background knowledge. Nevertheless, you typically receive query results as the linear equation solver just times out after searching for an approximately correct solution after a while and returns the best approximation it could find until then. The method of Least Squares (which you can enforce using \hyperref[cmdline:nnls]{\texttt{--nnls}}) tries to make the sum of the squares of the differences between right-hand and left-hand sides of the equation system obtained from your constraints as small as possible, which ideally gives you a result which is closer to being a solution than all other results, so to say - however, there is no guarantee for this. How other approximation approaches built into \prasp, such as simulated annealing or iterative refinement, cope with inconsistencies depends on their respective algorithm.\\

\prasp can discover logical and probabilistically inconsistencies in background knowledge (withing limitations of accuracy, rounding errors, etc), but this needs to be activated using switch \hyperref[cmdline:checkconsistency]{\texttt{--checkconsistency}}.\\

You can use \hyperref[cmdline:check]{\texttt{--check}} to check to which degree the solver reproduces the given probabilties (but this still doesn't give you information about results for other formulas).\\

\prasp currently cannot compute results in the presence of inconsistencies of type 1) (unsatisfiable spanning program), except when doing MAP inference using \hyperref[cmdline:maxwalksat]{\texttt{--maxwalksat}} (where no spanning program is required). But see \cite{mlnasp} in this regard.

\subsection{Markov Logic Programming (sort of) and inductive definitions}
\label{markov}

PrASP, being a logic programming framework, is able to handle inductive definitions, in contrast to, e.g., Markov Logic Networks (MLN)\footnote{This restriction of MLN is due to a restriction of FOL \cite{denecker}. While \prasp supports unrestricted FOL syntax (provided domains are finite), it is form of logic programming framework whose inferencer can be seen as a generator of a finite number of models, not a FOL theorem prover. }. To illustrate this, consider the following example taken almost literally from \cite{mlnasp} (which describes the language $\mathrm{LP^{MLN}}$ which combines MLN-style weight semantics with stable model semantics which can also deal with such definitions) and minimally adapted to satisfy \prasp syntax rules:

\begin{verbatim}
friend(a,b).
friend(b,c).

#dontExternalize
[[0.7]] influences(X, Y) :- friend(X, Y).
#endDontExternalize

influences(X, Y) :- influences(X, Z), influences(Z, Y).
\end{verbatim}

(Remark: We surround the uncertain formula with \verb§#dontExternalize..#endDontExternalize§ because we want the grounder to use the ground instances of that formula to bind the variables in\\ \verb§influences(X, Y) :- influences(X, Z), influences(Z, Y)§. If we would omit\\ \verb§#dontExternalize..#endDontExternalize§, \prasp would make \verb§influences§ \verb§#external§\footnote{For the meaning of \texttt{\#external}, please see Gringo/Clingo manual.} and we would need to state the domain of variables \verb§X§, \verb§Y§, \verb§Z§ otherwise (using a domain predicate, such as \verb§person(a;b;c)§, see further below), and this would lead to an unnecessarily large set of ground instances.)\\

\noindent Using the following queries:

\begin{verbatim}
[?] influences(a,b).
[?] influences(b,c).
[?] influences(a,c).
\end{verbatim}
 
\noindent it looks like \prasp correctly determines that the probabilities of the first two query formulas are equal whereas the probability of the third one is less. As with the example in \cite{mlnasp}, four possible worlds (stable models) are generated (we just present them in a different order, to match the order in \cite{mlnasp}):
\begin{verbatim}
pw1: influences(a,b),friend(a,b),influences(b,c),friend(b,c)
pw2: influences(a,b),friend(a,b),friend(b,c)
pw3: friend(a,b),influences(b,c),friend(b,c)
pw4: friend(a,b),friend(b,c)
\end{verbatim}

However, things are not as simple as they may seem... The program above is inconsistent under \prasp semantics. To see why, it is sufficient to see that 
\verb§[0.7] influences(a,c) :- friend(a,c)§
is among the annotated ground instances of \verb§[[0.7]] influences(X, Y) :- friend(X, Y)§. However, \verb§not friend(a,c)§ holds in each of the four possible worlds (since \verb§friend(a,c)§ is absent in all worlds), which is inconsistent with the assignment of probability 0.7 to this ground rule. \\

To emulate the MLNASP example more closely, we need to switch to an emulation of MLN semantics in the style of \cite{mlnasp} (which in contrast to actual Markov Logic Networks supports inductive definitions) using command line switch \hyperref[cmdline:mlns]{\texttt{--mlns}}. Observe that weights specified in background knowledge are not probabilities under this option, whereas the weights resulting from queries are still probabilities.\\

We also need to change the weight of \verb§influences(X, Y) :- friend(X, Y)§ to 0.9999..., since we need to make \verb§influences(X, Y) :- friend(X, Y)§ a soft rule with a weight as closely to 1 as possible (1 would make the rule a hard rule in \prasp). Furthermore, we now add a domain predicate person in order to bind the variables in \verb§influences(X, Y) :- influences(X, Z), influences(Z, Y)§ and we annotate this formula with \verb§[[1]]§ so that it is actually expanded into its 27 ground instances. Neither adding \verb§person§ nor the \verb§[[1]]§ annotation are actually required with \prasp (and they negatively impair performance), we just do these to resemble the original $\mathrm{LP^{MLN}}$  example a closely as possible.\\

\noindent Our full ``Markov Logic Programming'' program is as follows (in Gringo 3 syntax):

\begin{verbatim}
friend(a,b).
friend(b,c).

person(a;b;c).
#domain person(X).
#domain person(Y).
#domain person(Z).

[[0.99999]] influences(X, Y) :- friend(X, Y).

[[1]] influences(X, Y) :- influences(X, Z), influences(Z, Y). 
\end{verbatim}

\noindent Using Gringo 4 syntax:

\begin{verbatim}
friend(a,b).
friend(b,c).

person(a;b;c).

[[0.99999]] influences(X, Y) :- friend(X, Y), person(X), person(Y).

[[1]] influences(X,Y):-influences(X, Z),influences(Z, Y),person(X),person(Y),person(Z).
\end{verbatim}

\noindent With this program and switch \verb§--mlns§, we obtain about the same results as those presented in \cite{mlnasp} for the original $\mathrm{LP^{MLN}}$ program and the query file above.\\

As a side note, \prasp also supports MLN-style MAP inference using the MaxWalkSAT algorithm (see \hyperref[cmdline:maxwalksat]{\texttt{--maxwalksat}}). In contrast to the approach described above, this form of inference doesn't compute answer sets.

\subsection{Weight learning} 
\label{paramlearn}

Weight learning (also called parameter estimation) allows to determine approximate weights of a given hypothesis such that the likelihood of the given data (the so-called learning examples) is maximized.  \\

For learning, \prasp requires a file with formulas which constitute the hypothesis and a file with examples. Also, a file with background knowledge needs to be provided again. So the command-line has the following form for learning:\\

\noindent \verb|<file1.prasp> <file2.hypoth> <file3.examples>|\\
(If these file name extensions are not observed, you need to use command-line\\ \verb|--bgk <file1> --examples <file2> --learn <file3>| instead.) \\

Since learning relies on a search process using inference, inference-related command-line options such as those described in Section \ref{performance} and \ref{commandline} also have an effect on the learning task.\\

File  \verb|<file1.prasp>| contains background knowledge $b$ (in form of some \prasp program, as before).\\
\verb|<file2.hypoth>| contains a hypothesis in form of a set of formulas $h_1,...,h_n$, each prefixed with \verb|[?]|\\
\verb|<file3.examples>| contains a set of weighted or unweighted formulas $e_1,...,e_m$.\\

Weight learning means that \prasp finds for each formula in \verb|<file3.hypoth>| a weight such that these formulas support best the examples in \verb|<file3.examples>|, i.e., such that the probability of the examples (the data) is maximized  - a so-called \textit{maximum likelihood estimation} task. More precisely, we make the i.i.d. assumption and the learning algorithm searches for formula weights such that $\prod_i Pr(e_i|\theta)$ is maximized (if command-line option \hyperref[cmdline:maxconjexamples]{\texttt{--maxconjexamples}} is not used), or that $Pr(\bigwedge_{e_i}|\theta)$ is maximized (if \hyperref[cmdline:maxconjexamples]{\texttt{--maxconjexamples}} is given). The examples do not need to be complete, that is, it is not required that they unambiguously represent complete models.\\
In the previous formula, $\theta$ denotes a program consisting of all weighted hypothesis formulas and any background knowledge. The maximization algorithm used in \prasp \version\ by default is the Barzilai-Borwein algorithm \cite{Barzilai}, a variant of the gradient ascent method.\\

Let's consider an example. The following is our background knowledge (stored in file \verb|smokersNetwork.prasp|):

\begin{verbatim}
person(1).
person(2).
person(3).
person(4).

friend(1,2).
friend(2,1).
friend(2,4).
friend(3,2).
friend(4,2).

#domain person(X).
#domain person(Y).

smokes(X) :- stress(X).
smokes(X) :- friend(X,Y), influences(Y,X), smokes(Y), X != Y.

[0.8] stress(4).
\end{verbatim}

This models a social network consisting of four persons who may influence each other wrt. their smoking habits. The last formula expresses that person 4 is probably stressed.\\

\noindent As learning example, we store the following in file \verb|smokersNetwork.examples|:

\begin{verbatim}
smokes(2).
\end{verbatim}

We are now looking for a hypothesis (and its weights) which explains the above example. A reasonable assumption would be that person 2 might smoke because (s)he is influenced by person 4, about whom we already now that (s)he likely smokes (because (s)he is probably stressed). Observe that person 2 is a friend of person 4. Another explanation for the example would be that person 2 is stressed himself/herself. We put these assumptions to the test by creating a file  \verb|smokersNetwork.hypoth| with the following content:

\begin{verbatim}
[?] influences(4,2).
[?] stress(2).
\end{verbatim}

\noindent ...and invoking \prasp with command-line arguments\\

\noindent \verb|smokersNetwork.prasp smokersNetwork.examples smokersNetwork.hypoth| \verb|--folconv none| \hyperref[cmdline:extiidanalysis]{\texttt{--extiidanalysis}}\\

\noindent \prasp \version\ returns the following result:

\begin{verbatim}
[0.6] influences(4,2).
[0.9] stress(2).
\end{verbatim} 

This means that there is an over average probability that person 2 is influenced by person 4 and a very high probability that person 2 is stressed. Note that the examples are neither knowledge nor belief - there is no guarantee that any weighted hypothesis exists which makes them all certain or maximizes their individual likelihoods equally well. However, if there are multiple formulas in the examples file and we provide command-line option \hyperref[cmdline:maxconjexamples]{\texttt{--maxconjexamples}}, \prasp maximizes the likelihood of $e_1 \wedge ... \wedge e_n$ (where the $e_i$ are the examples), which is a stronger maximization target. \\

\noindent Remarks:
\begin{itemize}
	
	\item{Both the examples file and the hypothesis file can each contain multiple formulas. Weights of formulas in the hypothesis file are not learned in turn, but the entire file is treated as a single hypothesis which consists of multiple parts (the individual formulas whose weights are learned). }
	
	\item{Formulas in the examples file do not need to be atomic or ground, they could also be, e.g., rules (although that would be rather unusual). However, using \hyperref[cmdline:maxconjexamples]{\texttt{--maxconjexamples}} together with multiple non-ground literals currently requires that a FOL$\rightarrow$ASP converter is installed (see command-line option \hyperref[cmdline:folconv]{\texttt{--folconv}}).}
	
	\item Learning results are generally approximations. Also, the weight learning algorithm comprises a random search component, so learning results might not always be exactly reproducible.
	
	\item{Weighted example formulas are not yet supported.}
	
\end{itemize}

\subsection{Uncertainty reasoning and learning with streaming data}

\prasp can act as a reasoning and learning server fed by stream data from a local or remote host on the Internet. Clients send a stream to \prasp (the reasoning server) and emit the results returned from \prasp (see Fig. \ref{fig:streamReasoning}). Stream processing is an experimental feature of \prasp.\\

\begin{figure}
	\centering
	\includegraphics[width=1.1\linewidth, scale = 1, trim=5mm 5mm 0 10mm]{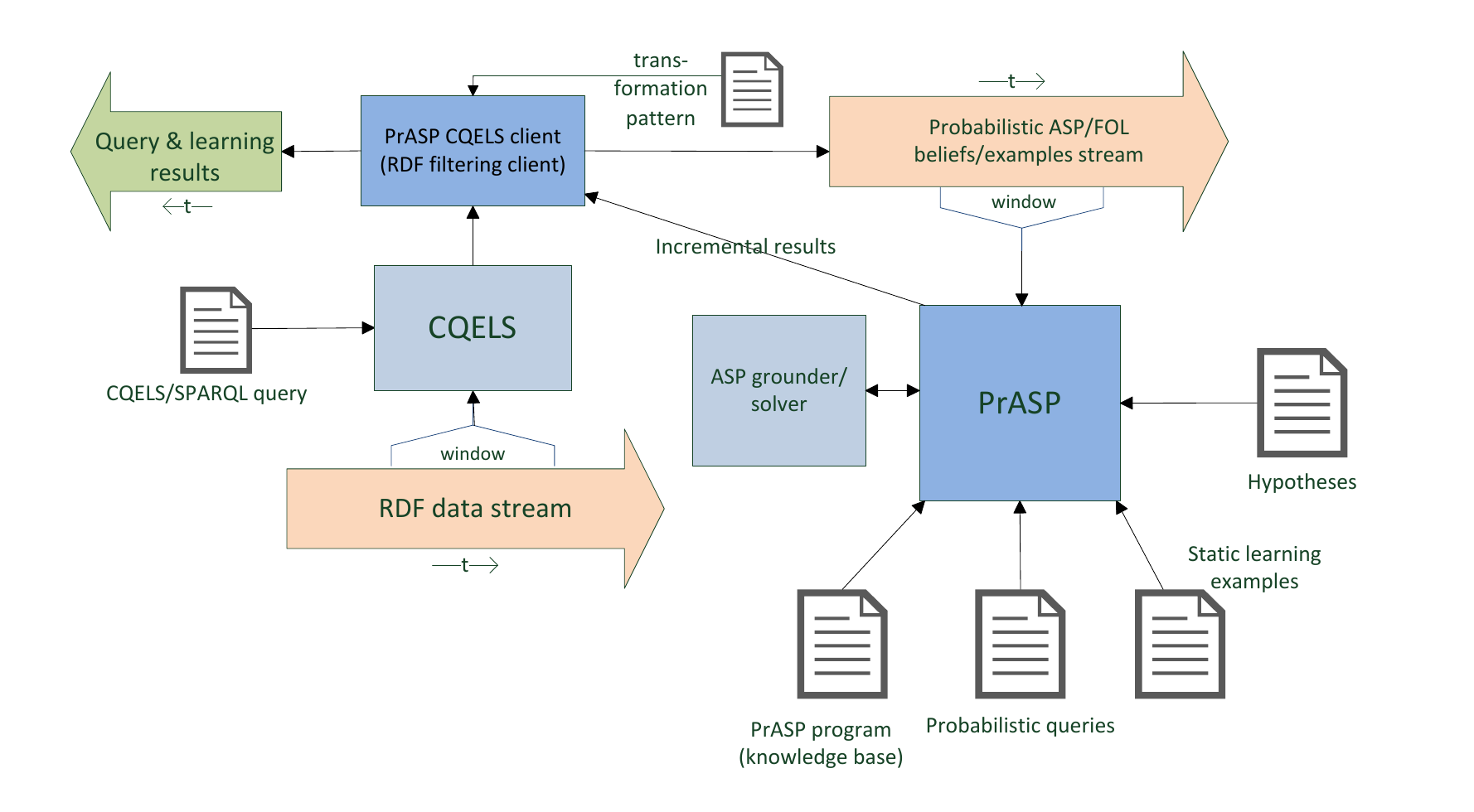}
	\caption{\prasp stream reasoning overview} 
	\label{fig:streamReasoning}
\end{figure}

Currently there are two clients: 1) a console client, where the stream data is entered manually in the console in ASP/\prasp syntax (mainly for testing and debugging purposes) and 2) an RDF/SPARQL streaming client (\textit{PrASP CQELS client}) which employs the Linked data stream processing engine CQELS \cite{cqels}. This way, \prasp can be used as a tools for uncertainty reasoning about RDF streams (that is, about results of CQELS queries), whereas the console client can be seen as a simple REPL for PrASP (although the \prasp Web Interface (Sect. \ref{installation}) might be easier to use for trying out PrASP).\\

\noindent The console client is located in directory \verb§PrASP_StreamConsole§.\\
The stream client (PrASP CQELS client) is located in directory \verb§PrASP_RDF_StreamClient§.\\

\noindent \textbf{1)} To use the \underline{\textbf{PrASP Console Client}} (folder \verb|prasp_StreamConsole|), first start \prasp as usual, but specify additional argument \verb|--stream 7777| (or some other port number which must match the port number specified in the PrASP Console Client script). \prasp expects all the usual files at startup - if you want it to start without initial background knowledge or initial learning examples, you need to provide an empty \verb|.prasp| (or an empty \verb|.examples| file, respectively).\\

As soon as \prasp shows "Listening for stream data ...", start the console client using \verb|\prasp_cc.sh| (Linux, MacOSX) or \verb|\prasp_cc.bat| (Windows)\\

Now you can send new beliefs or examples to \prasp by typing after \verb|"> "|. Beliefs are incrementally added to the initial background knowledge (i.e., given by the \verb|.prasp|-file specified when \prasp was launched).
Sending examples only make sense if you started \prasp in learning mode, i.e., using 
a \verb|.hypoth| and an possibly empty \verb|.examples| file. Sending examples and beliefs can also be mixed.\\

E.g., send a fact (belief) by entering \verb|B coin2(heads).|  (always conclude formulas with a ``dot''.)\\
Send an uncertain belief using, e.g., \verb|B [0.7] coin2(heads).| 
(You can also send certain and uncertain rules, negative formulas, conditional probabilities, etc. as beliefs.)
Send an example using, e.g., \verb|E coin2(heads).| \\

After each step, \prasp replies with updated results (query results or re-weighted hypotheses). Enter an empty line to see the current query results.\\

Optionally, you can specify the lifetime of a formula using a number directly behind "\verb§B§" or "\verb§E§".
E.g. \verb|B10 [0.7] coin(heads).| means that formula coin(heads) remains in the KB only during a time window consisting of the next 10 steps and is then removed. A ``step'' is a single interaction with \prasp, but for the calculation of formula lifetime only those steps count where a belief or example was added.\\
If omitted, life expectancy is infinite. If you always use the same number, you have a window.\\

Formulas can also be retracted, using \verb|RB f| (retract a belief) or \verb|RE f| (retracts an example). All formulas need to be concluded by a dot, as usual. Note that it is only possible to retracts formula which have been added before using the client, not formulas
in the KB file (the \verb|.prasp|-file) or in the examples file provided at \prasp startup.\\

Commands \verb|SB| and \verb|SE| show the list of currently valid external beliefs or examples, respectively.\\

\noindent \textbf{2)}\underline{ \textbf{PrASP CQELS client}}, for supplying \prasp with a filtered stream (results of a SPARQL/CQELS \cite{cqels} query) which itself originates from an RDF data stream:\\

\noindent This client is located in folder \verb|prasp_RDF_StreamClient|.\\

First, edit \verb|prasp_rdfsc.sh| or \verb|prasp_rdfsc.bat| in order to specify data files, home directory (where CQELS should put cache files etc), data URIs, and the file with the extended CQELS-like query (e.g., \verb§query1.psquery§).

Then, start \prasp as usual, but specify additional argument \verb|--stream 7777| (or some other port number, as defined in \verb|\prasp_rdfsc|).\\
Example (using the ``floorplan'' example files in folder \verb|\prasp_RDF_StreamClient|):\\

\verb§prasp floorplan.examples floorplan.prasp floorplan.hypoth --folconv none §\\
\verb§    --mod1 --noautoindeps --noindepconstrs --cacheinfsetup --stream 7777 §\\

Instead of this command-line, you can specify any other switches and tasks, e.g., inference instead of learning. The only difference compared to a normal \prasp invocation is argument \hyperref[cmdline:stream]{\texttt{--stream 7777}} (and optionally \hyperref[cmdline:cacheinfsetup]{\texttt{--cacheinfsetup}}).\\

As soon as \prasp shows "Listening for stream data ...", start the client using \verb|.\prasp_rdfsc.sh| or \verb|prasp_rdfsc.bat|\\

\noindent Note that \verb§floorplan.examples§ is empty (but it still needs to be provided).\\

\noindent In addition to the usual \prasp files, you also need an extended CQELS/SPARQL-like query file (file ending \verb§.psquery§):\\

This file needs to be specified using \verb§set QUERY=...§ within \verb§prasp_rdfsc.sh§/\verb§.bat§. It provides the SPARQL-like query pattern for the CQELS-based client and should not be confused with the \verb§.query§ file necessary to specify inference queries (which you need additionally if you want to use the streaming data for inference instead of learning).\\

\noindent The transformation pattern \textit{SPARQL-Result}$\rightarrow$\textit{(Pr)ASP facts} is specified directly 
\underline{inside} the CQELS SPARQL-like query, after keyword \texttt{\prasp}:\\

\noindent\texttt{...\\
	SELECT ...\\
	\prasp \textit{pattern} [OMIT \textit{URIs}]\\
	...}\\

\noindent At this, \textit{pattern} is the same as for the console client, e.g.,\\
\verb|E20 detectedAt(?person, ?loc, TIMESTAMP).|  (don't forget the dot).
Any \textit{?vars} in pattern need also to occur after \verb|SELECT|. \verb|TIMESTAMP| is an automated counter 1, 2,...
\texttt{[OMIT \textit{URIs}]} specifies that the listed URIs (separated by commas) shall be omitted in the ASP formulas.
\texttt{[OMIT \textit{URIs}]} and use of \verb|TIMESTAMP| are optional.\\

\noindent An example query file is provides as \verb§query1.psquery§\\

Reasoning about streams might profit significantly from \prasp command-line switch \hyperref[cmdline:cacheinfsetup]{\texttt{--cacheinfsetup}} in case the stream contains repeated subsequences.\\

\noindent For details about the pure CQELS query syntax (without \prasp extensions), please consult \cite{cqels}.\\

\noindent {\color{Black} (More tbw.) }

\subsection{Performance tuning}
\label{performance}

\prasp's default settings are rather conservative. This section shows how to increase inference (and  indirectly learning) speed using problem-specific command-line options. But foremost, the background knowledge influences inference performance - while \prasp's input language is quite expressive, not all
valid probabilistic logic programs and query types are likewise suitable for fast or even just tractable inference. 

\subsubsection{A rule of thumb}

The number of options available in \PrASP can appear a bit overwhelming at first. Here's a rule of thumb which proved to be a good heuristic in many of our own experiments:

\begin{enumerate}
	\item  Wherever possible, use Clingo/Gringo 4 (or higher) instead of 3. You can use the \texttt{toggle...} scripts to switch between the two versions. Restricting yourself to ASP or even Datalog-like syntax (see \ref{cmdline:folconv} and \ref{simpleFormulas}) also helps of course since reasoning with full FOL is significantly more time consuming.
	\item Firstly, invoke \PrASP with the default (``vanilla'') settings. Try different solvers using \hyperref[cmdline:linsolveconf]{\texttt{--linsolveconf}}.
	\item If this turns out to be too slow, try next with \hyperref[cmdline:itrefinement]{\texttt{--itrefinement --models 0}}. Given a full set of models (enforced with \hyperref[cmdline:models 0]{\texttt{--models 0}}), this will still give you an accurate result (modulo round-off errors).
	\item As further measure, let \PrASP try simplifying the problem using \hyperref[cmdline:mod1]{\texttt{--mod1}} (together with \hyperref[cmdline:folconv]{\texttt{--folconv none}}, if possible). 
	\item In some cases, neither default settings nor \hyperref[cmdline:itrefinement]{\texttt{--itrefinement}} are useful due to the complexity of the problem. Use \hyperref[cmdline:simanneal]{\texttt{--simanneal}} (in combination with \hyperref[cmdline:nosolve]{\texttt{--nosolve}}) to approach such problems. 
	\item There is a pretty large number of other command-line switches and arguments which have an influence on inference speed. They are detailed further below in this section and in Section \ref{commandline}. 
	\item Since version 0.9, PrASP also supports MaxWalkSat as inference algorithm (see \hyperref[cmdline:maxwalksat]{\texttt{--maxwalksat}}), which turns PrASP into a sort-of-Markov-Logic MAP inference tool (however, this restricts the sort of formulas you are able to use in your background knowledge and queries).
\end{enumerate}   

\subsubsection{Motivation}

\noindent To obtain a deeper understanding of performance issues and their causes, let's take a look at the parameterized coins example again:

\begin{verbatim}
coin(1..20).

#indep
[0.6] coin_out(1,heads).
[[0.5]] coin_out(N,heads) :- coin(N), N != 1.
#endIndep

1{coin_out(N,heads), coin_out(N,tails)}1 :- coin(N).

n_win :- coin_out(N,tails), coin(N).
win :- not n_win. 
\end{verbatim}

Even with only 20 coins, inference using default ``vanilla'' settings (linear system solving) is slow: all coin tosses are independent from each other, which causes a combinatorial explosion of possible worlds - we get 1048576 possible worlds and accordingly
the same number of rows in the equation system we would need to solve. However, the real bottleneck here is not so much the linear system solving step (PrASP's native LAPACK solver can compute the solution for systems with $10^6$ variables/possible worlds in a few seconds on contemporary single CPUs) but the time required to generate possible worlds.\\

Generally, the more weight-annotated formulas and other choices (disjunctions) a spanning program contains, the larger the number of possible worlds (stable models, answer sets), and the more time is required for inference and learning.\\

\noindent However, there are a number of approaches built into \prasp which cope much better with larger systems than the ``vanilla'' setup:

\subsubsection{Use of Clingo 4 and remote sampling}

This implementation of some of the model generation and sampling methods is typically 
much faster than older Gringo 3/Lparse based approach.\\
However, it requires you to use Gringo and Clingo version 4 or higher, which has a number of consequences for \prasp program syntax. See \hyperref[cmdline:noremotesampling]{\texttt{--noremotesampling}} for details. This switch actually \textit{de}activates Clingo 4-based remote (i.e., fast) model generation, but also provides some background information about it. To activate remote sampling, you just need to install Gringo 4 \& Clingo 4 and to make sure that \prasp finds them in the \prasp directory (see Section \ref{installation}).\\

The only required change to the coins-example program above would be to replace the comma with a semicolon within \verb§1{coin_out(N,heads), coin_out(N,tails)}1 :- coin(N)§, in order to meet Gringo 4 syntax requirements.\\

However, with the coins example above, inference would be still slow with this measure alone. For this particular example, e.g., simulated annealing (see below) in combination with \hyperref[cmdline:nosolve]{\texttt{--nosolve}} (to omit the linear system solving step) and \hyperref[cmdline:noindepconstrs]{\texttt{--noindepconstrs}} \hyperref[cmdline:noautoindeps]{\texttt{--noautoindeps}} (to omit the generation of independence constraints) would be an appropriate and fast approximation approach. 

\subsubsection{Iterative refinement as inference algorithm}
\label{itrefinespeedtipps}

Another approximate inference method for medium-size or large systems is i\textit{terative refinement} (switch \hyperref[cmdline:itrefinement]{\texttt{--itrefinement}}). Please see Sect. \ref{itrefinement} for details.

\subsubsection{Simulated annealing as inference algorithm}
\label{simannealspeedtipps}

With \hyperref[cmdline:simanneal]{\texttt{--simanneal}}, \prasp uses a variant of \textit{simulated annealing} for approximative inference. For problems with many degrees of freedom (large number of uncertain formulas in background knowledge), this approach (see Sect. \ref{simanneal}) is typically faster compared to the default approach for medium-size or large systems. \hyperref[cmdline:simanneal]{\texttt{--simanneal}} also copes better than \hyperref[cmdline:itrefinement]{\texttt{--itrefinement}} with small numbers of models. It is often the only way to approach complex systems for which no accurate solution can be found.\\

Simulated annealing works often much faster for such large systems compared to the default inference approach, however, it does not necessarily discover the best solution (the one with maximum entropy). Also, consideration of formula independence is a tricky issue: in some cases, independence speeds up the algorithm, in other situations it slows it down (details below).\\

Random sample models are computed until the probability distribution over all sampled models matches the probabilities (formula weights) specified in the background knowledge. With \hyperref[cmdline:simanneal]{\texttt{--simanneal}}, the equation solving phase is not required (and therefore \hyperref[cmdline:simanneal]{\texttt{--simanneal}} is typically used in connection with \hyperref[cmdline:nosolve]{\texttt{--nosolve}}). However, in some cases, it might make sense to omit \hyperref[cmdline:nosolve]{\texttt{--nosolve}} - in that case \hyperref[cmdline:simanneal]{\texttt{--simanneal}} works as a possible world sample generator and the probability distribution over the sampled possible worlds is computed using equation solving, which can lead to increased accuracy (at the price of being much slower than \hyperref[cmdline:nosolve]{\texttt{--nosolve}} unless the system is very small).\\

\hyperref[cmdline:simanneal]{\texttt{--simanneal}} can optionally be configured using various parameters which have a tremendous influence on computation time and accuracy of results - please see \ref{commandline} for details.\\
Also various command-line switches have a strong influence on how simulated annealing behaves, such as \hyperref[cmdline:unisample]{\texttt{--unisample}}, \hyperref[cmdline:ascheckmode]{\texttt{--ascheckmode}}, \hyperref[cmdline:mod1]{\texttt{--mod1}} and \hyperref[cmdline:noindepconstrs]{\texttt{--noindepconstrs}}. Typical combinations
are achieved with \verb§-o2§ or \verb§-o2asp§.\\

\noindent \textbf{Tips}:

\begin{itemize}
	
	\item \hyperref[cmdline:simanneal]{\texttt{--simanneal}} performs a complex multi-goal optimization task. It works best if there are no or only few conditional probabilities specified in background knowledge (as these generate an  large set of candidate samples) and if known mutual or pairwise independence constraints are not considered (\hyperref[cmdline:noindepconstrs]{\texttt{--noindepconstrs}}). \hyperref[cmdline:simanneal]{\texttt{--simanneal}} can still make use of independence assumptions if \hyperref[cmdline:noindepconstrs]{\texttt{--noindepconstrs}} is provided, \prasp just doesn't generate additional constraints then besides the weighted formulas in background knowledge.\\
	Also, all involved formulas should ideally be in ASP syntax and \hyperref[cmdline:folconv]{\texttt{--folconv none}} should be set. \hyperref[cmdline:simanneal]{\texttt{--simanneal}} works with other settings too, but significantly slower.
	
	\item Ideally, all weighted formulas should be mutually independent (as in ProbLog \cite{problog}). In that case, the algorithm can typically compute the result in a single step. This does not contradict the previous advice (to specify \hyperref[cmdline:noindepconstrs]{\texttt{--noindepconstrs}}),
	since \hyperref[cmdline:noindepconstrs]{\texttt{--noindepconstrs}} only prevents the generation of independence \textit{enforcing} constraints but not the \textit{assumption} of independence.
	
	\item Annotated formulas in background knowledge, query formulas, learning examples and hypotheses should ideally be so-called \textit{simple formulas}, see \ref{simpleFormulas}. Again, this is not mandatory, but it speeds up \prasp significantly. If this is not possible, consider specifying a different model checking approach using \hyperref[cmdline:ascheckmode]{\texttt{--ascheckmode}} in case the number of models of the spanning program is very large (see below).
	
	\item \hyperref[cmdline:simanneal]{\texttt{--simanneal}} can be combined with other optimization approaches (e.g., \hyperref[cmdline:mod1]{\texttt{--mod1}}).
	
	\item Although \hyperref[cmdline:simanneal]{\texttt{--simanneal}} can be used with \hyperref[cmdline:intervalresults]{\texttt{--intervalresults}}, this combination has its issues (see \ref{intervalresults}).
	
	\item The ``energy'' parameter of \hyperref[cmdline:simanneal]{\texttt{--simanneal}} specifies the acceptable euclidean distance (which \prasp aims to minimize) between the vectors of actual and specified probabilities of uncertain formulas in background knowledge, given the current list of sampled models of the spanning program. A larger ``energy'' argument means less accurate (but quicker) results. Further details can be found in Sect. \ref{simanneal}.
	
\end{itemize}

\subsubsection{Using alternative approaches to checking of formula validity}
\label{altcheck}

At various points, \prasp needs to check whether a formula $f$ holds in a certain possible world $m$. The obvious approach to this is to check if the given possible world is an element of the enumeration of all possible worlds of the given formula in conjunction with the spanning program. There are two ways to make this faster: 1) use simple formulas (see \ref{simpleFormulas}) for which \prasp can check if they hold in a given model without the need to call the ASP solver and/or 2) enforce a different model checking approach using command-line switch \hyperref[cmdline:ascheckmode]{\texttt{--ascheckmode}} (but by default, \prasp typically uses a fast approach here already). \\

There is also the possibility to restrict the number of models requested from the ASP solver using \hyperref[cmdline:models n]{\texttt{--models n}}, however, even though these can be sampled near-uniformly, there is no guarantee that results are correct using these samples.

\subsubsection{Combining and filtering background knowledge formulas}

\noindent\textbf{\texttt{--mod0}}\\
\label{mod0}

Using  \hyperref[cmdline:mod0]{\texttt{--mod0}}, \prasp combines unweighted ground ASP rules (except for headless rules) which have the same head and whose bodies consist of independent sets of ground literals \textit{only}, and replaces them with weighted formulas. See Sect. \ref{meta} for how to declare such independence. To obtain a speed gain, a significant number of rules suitable for simplification has to exist in the \prasp program. If this is not the case, \hyperref[cmdline:mod0]{\texttt{--mod0}} slows down inference. Weighted formulas with single-bracket weight annotations in background knowledge need to be ground for \hyperref[cmdline:mod0]{\texttt{--mod0}} to work.\\

\hyperref[cmdline:mod0]{\texttt{--mod0}} can be combined with \hyperref[cmdline:mod1]{\texttt{--mod1}} (which is explained below). This allows \prasp to remove independent formulas from the background knowledge in case these have only been used within rule bodies which have been simplified using \hyperref[cmdline:mod0]{\texttt{--mod0}}.\\

Also see \hyperref[cmdline:limitindepcombs]{\texttt{--limitindepcombs}} in Section \ref{commandline} and the general explanation of how \prasp deals
with probabilistic independence in \ref{indep} (under meta-statement \texttt{\#indep}).\\

Provide switch \hyperref[cmdline:verbose]{\texttt{--verbose}} for information about which rules couldn't be simplified by \prasp, and why.\\

Observe that \prasp generally only simplifies formulas in background knowledge, not any formulas in query files (at least not beyond any query simplifications performed by the ASP grounder).\\

\verb§--mod0§ doesn't yet support full Gringo 4 syntax.\\

\noindent\textbf{\texttt{--mod1}}\\
\label{mod1}

With \hyperref[cmdline:mod1]{\texttt{--mod1}} (which can be used alone or together with  \hyperref[cmdline:mod0]{\texttt{--mod0}}), \prasp analyses the background knowledge and removes all parts on which the result of the queries cannot depend. The speed gain depends on the respective \verb§.prasp§ program and can be either tremendous (if the query depends only on a small module of the background knowledge), or moderate, or even negative (if the cost for the dependency analysis prevails the gain from the modification). For the coin flipping example above, the speed gain is still about 50\%, even though query \verb§w§in depends on all 13 coins (the reason for this is that \hyperref[cmdline:mod1]{\texttt{--mod1}} removes facts \verb§coin(1)§, \verb§coin(2)§, etc at some stage of the process).\\

\hyperref[cmdline:mod1]{\texttt{--mod1}} has shortcomings too: 1) weighted formulas with single-bracket weight annotations in background knowledge need to be ground (this is typically not a severe restriction).\\
2) All query formulas are converted into sets of grounded formulas. Multiple ground instances are shown in \verb§#{...}# §-brackets in query results.\\ 3) \hyperref[cmdline:mod1]{\texttt{--mod1}} together with conditional probabilities in background knowledge might lead to errors or inaccuracies (\prasp therefore shows a warning if \hyperref[cmdline:mod1]{\texttt{--mod1}} is used and such formulas are found in the \verb§.prasp§ file)\footnote{If the spanning formula generated for the conditional probability conveys all information necessary for the dependency analysis performed by \texttt{--mod1}, there is no problem. But \prasp cannot guarantee this and therefore warns against combining \texttt{--mod1} with conditional formulas in background knowledge.}. For conditional probabilities within query files, there is no such issue.\\
4) Finally, analyzing and optimizing a program using \hyperref[cmdline:mod1]{\texttt{--mod1}} takes time itself which needs to be traded off against the speed gain obtained with the automatically optimized program. \\

Observe that \hyperref[cmdline:mod1]{\texttt{--mod1}} prunes a program and can thus turn a logically inconsistent program into a consistent one (which is normally \textit{not} desired) by removing parts on which no query depends.\\

Since \hyperref[cmdline:mod1]{\texttt{--mod1}} and other (internal) optimization approaches use the information found in query-files in order to determine which parts of the background knowledge are relevant, it can make sense to put different query formulas into different query-files: for each \hyperref[cmdline:query]{\texttt{--query}}-occurrence in the command-line, and for each file with file name ending \verb§.query§ which is not bound by any other argument, the background knowledge is\textit{ processed anew} (so that different query-dependent optimizations can be applied). In contrast, for each set of query files bound by a certain \hyperref[cmdline:query]{\texttt{--query}} in the command-line, \prasp applies the same optimizations to the background knowledge and the actual inference task.

\subsubsection{Using ``simple formulas''}
\label{simpleFormulas}

\prasp deals with formulas which are plain ASP rules or conjunctions which consist of ground literals only considerably faster 
than with other kinds of formulas, because the models of such formulas can be computed relatively easily. 
It is recommended to use such simple formulas whenever possible, as they speed up inference and learning without
sacrificing much expressiveness (many probabilistic logics can deal only with even simpler formula classes). Other types of formulas can still be used to, provided they are translated into ``simple'' format during grounding phase (e.g., by annotating them with double-square weights, although this only guarantees grounding). \\

Benefits of using simple formulas apply to all kinds of input (including background knowledge and queries), 
even if only some formulas are in this format, there is already a speed gain.

\prasp recognizes formulas of the syntactic forms \verb§a1 & a2 & ... & an§ or n\verb§#count{a1,a2,...,an}§n or n\verb§{a1,a2,...,an}§n as simple formulas if all \verb§ai§ are ground literals, and also combinations thereof. Also, ASP rules of the forms \\
\verb§a1 :- a2, ... ,an§ and \verb§:- a2, ... ,an§ are ``simple formulas'' if all \verb§ai§ are ground literals. From a purely syntactical point of view, ``simple rules'' are thus similar to ground Datalog rules.

%\subsubsection{Adding enforcement/prevention constraints for event combinations}
%\label{enforcementConstraints}
%
%{\color{red}tbw.}

\subsubsection{Disabling full FOL syntax}

Another means to increase speed which is applicable alone or in combination with most other command-line options is to disable the FOL$\rightarrow$ASP converter by specifying\\
\hyperref[cmdline:folconv]{\texttt{--folconv none}}. Of course, all formulas need to be in ASP syntax then. For the coins example, we can achieve this simply by replacing  \verb|[?] not win| with  \verb|[?] :- win| in the queries file.

\subsubsection{Disabling independence enforcement constraints ot automatic detection of independence}

For known (declared or automatically discovered) mutually or pairwise independent sets of events (propositions), \prasp generates certain constraints. While these constraints can speed up \prasp in some cases (e.g., \ref{mod0}), in other cases they can slow down inference and learning. To disable them entirely, use switch \hyperref[cmdline:noindepconstrs]{\texttt{--noindepconstrs}}. To disable them for event combinations of certain lengths only, use \hyperref[cmdline:limitindepcombs]{\texttt{--limitindepcombs}} (\ref{commandline}).\\
Note that \hyperref[cmdline:noindepconstrs]{\texttt{--noindepconstrs}} has no influence on
whether or not formulas are treated as independent in other contexts - e.g., \hyperref[cmdline:mod0]{\texttt{--mod0}} can still make use
of independence declarations even if \hyperref[cmdline:noindepconstrs]{\texttt{--noindepconstrs}} is provided.\\

\hyperref[cmdline:noautoindeps]{\texttt{--noautoindeps}} omits the automated search for a (globally) mutually independent set of atoms. In case you have specified all independent formulas in the background knowledge file (see Section \ref{meta}), \hyperref[cmdline:noautoindeps]{\texttt{--noautoindeps}} does not only lead to faster inference but it also avoids the potential problem that the automated identification erroneously misidentifies some atom.\\
Even without speed considerations, it is a good idea to declare independence explicitly using the means described in \ref{meta} and to deactivate automatic discovery, since the latter relies on grounder-dependent heuristics which does not always work as expected. 

%A stronger potential means for performance tuning is option  \verb|--dnf|. With this (and omitting  \verb|--nosolve| and  \verb|--weights2cc|), inference remains accurate and becomes viable with 13 coins. But although \verb|--dnf| is a powerful tool for the reduction of computation time in some cases (provided formulas in background knowledge are relatively short), for the 13 coins example, it is alone not sufficient. Note also that \verb|--dnf| doesn't observe stable model semantics (models are the clauses of the disjunctive normal form instead of answer sets). For the coins example, this doesn't make a difference, but for other logic programs or queries,  \verb|--dnf| might lead to different inference results compared to the default settings. Whether \verb|--dnf| actually improves performance cannot be generally told for all uses cases, so time-critical inference and learning tasks should be tried with and without \verb|--dnf|. \\

\subsubsection{Answering queries using mere initial sampling and \texttt{--nosolve}}

See Sect. \ref{merecounting}

\subsubsection{Using \prasp for plain ASP reasoning}

PrASP subsumes plain ASP reasoning about non-probabilistic knowledge. Besides generating answer sets (\hyperref[cmdline:pwsamples]{\texttt{--pwsamples}}), it adds a query facility to ASP. E.g., using the following program:

\begin{verbatim}
flies(X) :- bird(X), not ab(X).
bird(tweety).
bird(sam).

bird(X) :- penguin(X).
ab(X) :- penguin(X).
penguin(sam).
\end{verbatim}

...queries such as \verb§[?] flies(tweety)§ or \verb§[?] 2{flies(sam),bird(sam)}2§ can be used to compute truth values (represented by 1 and 0) for these formulas, provided the answer sets are in agreement about them (otherwise you would of course get probabilities other than 0 and 1, e.g., when you replace \verb§penguin(sam)§ with \verb§0{penguin(sam)}1§).\\

However, these results are still computed with the help of probabilistic inference, which has two not-so-nice consequences: 1) there is a certain time overhead involved compared to a direct call of the ASP grounder/solver, and 2) due to round-off errors, you might sometimes get in fact, e.g., 0.999999999 instead of 1. 

\subsubsection{Further means to speed up inference and learning}

- Grounding of non-ground formulas can take up a significant amount of time, even with a fast grounder such as Gringo. \prasp the following options and meta-statements for influencing the grounding process:

\begin{itemize}
	\item \hyperref[cmdline:groundingconf]{\texttt{--groundingconf}}
	\item \verb§#dontExternalize..#endDontExternalize§ (see Sect. \ref{meta})
	\item \verb§#external§ (via Lparse/Gringo)
\end{itemize}

- If \hyperref[cmdline:mod1]{\texttt{--mod1}} is specified (or an optimization switch which comprises \hyperref[cmdline:mod1]{\texttt{--mod1}}), those parts of the background knowledge are removed on which no query formulas depend. It follows that the number of query formulas per query task should be kept to a minimum then. On the other hand, the probability distribution computed without \hyperref[cmdline:mod]{\texttt{--mod1}} is reused for multiple different query formulas, i.e., in this case it is better to solve all query formulas at once.\\

- Significance speed improvements in \textit{repeated similar} inference and learning tasks can be achieved using command-line switch \hyperref[cmdline:cacheinfsetup]{\texttt{--cacheinfsetup}} provided the task comprises some repetition (e.g., repeated floating windows in stream reasoning). It needs to be used with care though, since the cache controller makes the assumption that all inferencer setups are identical.\\

- If the user already knows a suitable estimation of the required number of initial sampled models, this number should be provided using \hyperref[cmdline:models]{\texttt{--models}} instead of letting \prasp figure out a number heuristically (or, even worse, use all models of the spanning program). Estimating the number of models of the spanning program is a costly task and not very precise.\\

- Switch \hyperref[cmdline:ignoreentropy]{\texttt{--ignoreentropy --ndistrs 1}} make \prasp compute probability distributions somewhat faster if the default inference approach is being used (which is generally rather slow unless for very small systems) (see \ref{intervalresults}).\\

- Switch \hyperref[cmdline:checkconsistency]{\texttt{--checkconsistency}} slows down inference significantly.\\ 

%- Another switch which can expedite \prasp inference is \verb|--noindepconstrs|. If omitted, \prasp does not enforces the probabilistic independence of those atoms which are logically independent according to the ASP solver. With this switch, inference (and thus also learning) results will typically be delivered faster, but in some cases results might be unintuitive, because truth values of independent formulas might not be uniformly distributed anymore.
%In particular, \verb|--noindepconstrs| should not be used for learning tasks. \\

- Learning becomes a little faster with switch \hyperref[cmdline:nonorm]{\texttt{--nonorm}}, which deactivates a result normalization step. With  \hyperref[cmdline:nonorm]{\texttt{--nonorm}}, hypotheses weights are shown as computed by the parameter estimation algorithm, even in case these weights are probabilistically inconsistent in the context of the background knowledge. Otherwise, a normalization step aims for leveling out such inconsistencies. \\

- Most of the considerations before also apply to learning (Section \ref{paramlearn}), since learning is based on inference. \\

In the particular case of our coins-example (but not in general!), we can obtain an approximation of the correct inference results using  \verb|--models n|, where
\verb|n| is the approximate number of possible worlds \prasp should compute. For 13 coins,  we obtain results which are fairly accurate for \verb|n| $\geq$ 4000. However, \hyperref[cmdline:models]{\texttt{--models}} is in general not a valid optimization approach since it is virtually impossible to predict whether the sampled models are sufficient input
for the equation solver so that the resulting probability distribution over possible worlds is accurate enough. A better approach to sampling (embedded in an optimization task) is via \textit{simulated annealing} (command-line
option \hyperref[cmdline:simannealing]{\texttt{--simannealing}}).\\

- Finally, using a different grounder/solver can make a difference. E.g., using Windows,\\
\noindent \verb§ --groundersolver gringoThenClaspMT.bat§\\
\verb§"-n%NZNUM --verbose=1 --project"§\\
\verb§"--project --rand-freq=0.9 --sign-def=3 --seed %SEED"§\\
specifies that a multithreaded version of Clasp 2/3 should be used (see script \verb§gringoThenClaspMT.bat§ for details).\\

\section{Syntax}
\label{syntax}

\subsection{ASP and FOL syntax}

The syntax of ASP formulas without their weights corresponds mostly to the input syntax of the grounder. The default grounder is Gringo \cite{potassco} - please see the Gringo/Clingo manual \cite{potasscoUserGuide} for a detailed description of the input language. Observe that Gringo/Clingo 3 and Gringo/Clingo 4 differ significantly wrt. input syntax.\\

If the optional FOL$\rightarrow$ASP converter F2LP is installed or the internal FOL$\rightarrow$ASP converter is specified (see switch \hyperref[cmdline:folconv]{\texttt{--folconv}}), formulas can alternatively be in FOL syntax as specified for F2LP (see \cite{f2lp}). Within the same file, formulas of both types (ASP and FOL/F2LP) can occur, however, ASP and FOL syntax should \underline{not} be mixed within the same formula (although there is some support for ASP constructs such as aggregates within FOL formulas if you use F2LP).\\
 
Syntax restrictions are listed in Sect. \ref{hints}. In particular, a single formula currently cannot extend over more than one line of text.\\

Irrespectively of the specific grounder or FOL converter being used, \prasp supports quoted \texttt{"strings"} and \textquotesingle\verb§strings§\textquotesingle\ in term positions. Because ASP grounders typically don't support strings directly, \prasp translates them into certain constants (names consisting of alphanumeric characters or underscores) and thus strings can be used with those built-in predicates provided by the ASP grounder which work with constants (see grounder manual). The translation from strings to constants preserves string equality and lexicographic order of strings. E.g., \verb§"abc" < "xyz"§ holds, using Gringo 3 as grounder.\\

Within strings, escape sequences can occur (e.g., \verb§"abc\ndef"§). In addition to the usual characters that need to be escaped (like newline),
it is also required to escape character '\verb§%§'. E.g., you need to write \verb§"abc\%xyz"§ instead of \verb§"abc%xyz"§. \\

The full syntax of string literals and the list of supported escape sequences (except the handling of '\verb§%§') can be found in \\
\verb§ http://www.ecma-international.org/ecma-262/5.1/#sec-7.8.4§\\

The other kinds of literals which are supported in term positions are integer numbers and alphanumeric constants.\\

\noindent \textit{More tbw.}

\subsection{Comments}
\label{comments}

\noindent Single-line \textit{comments} are prefixed by \verb|%|. Multi-line comments need to be enclosed in \verb§%*...*%§. Example:
\begin{verbatim}
[0.05] a & b.  % a single-line comment
%* This is
a multiline 
comment. *%
\end{verbatim}

\subsection{Parsing order}
\label{parsing}

The sequence of background knowledge parsing and preprocessing steps performed by \prasp is depicted in Figure \ref{fig:parsingOrder}. The sequence for processing query and other types of files is similar, but of course without generation of a \textit{spanning program} (Sections \ref{semantics},\ref{spanGen}).

\begin{figure}
	\centering
	\includegraphics[scale = 0.8,trim=15mm 70mm 85mm 65mm]{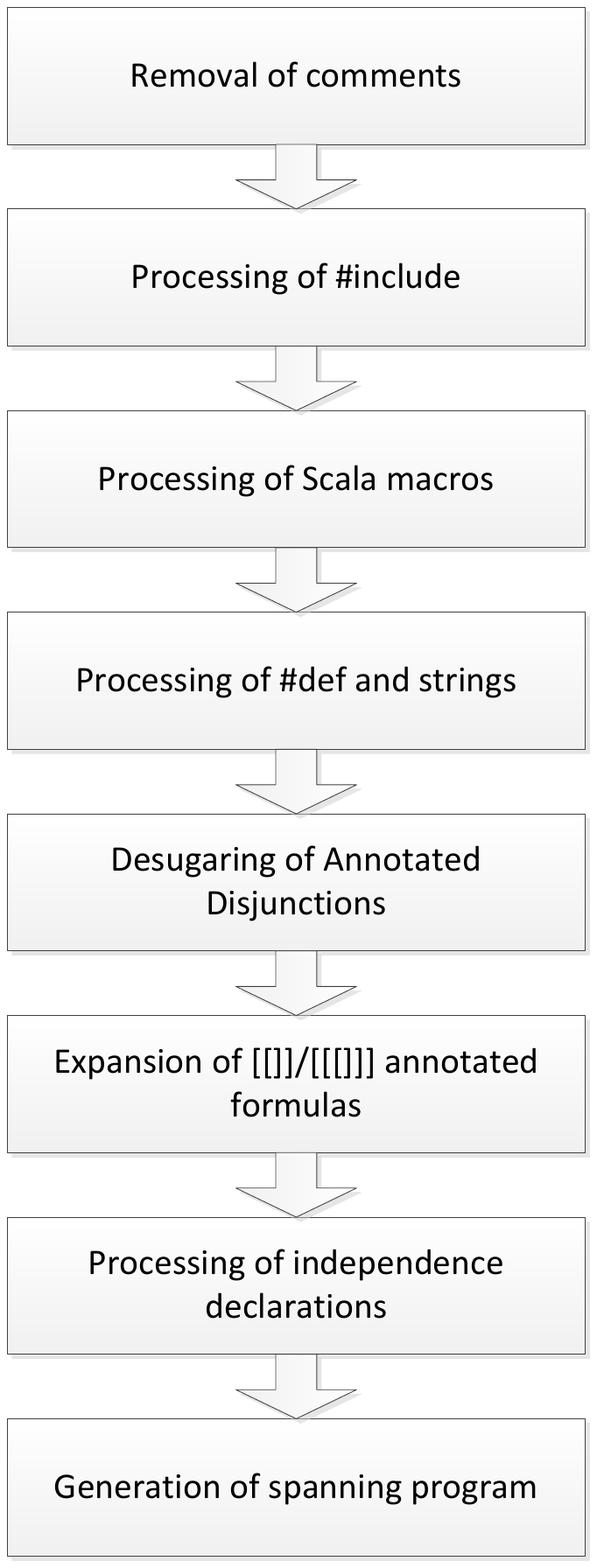}
	\caption{\prasp background knowledge parsing sequence} 
	\label{fig:parsingOrder}
\end{figure}

\subsection{Formula annotations and weights}
\label{weights}

In principle, any ASP or FOL formula can be annotated with a probability. Such weight annotations in background knowledge (\verb#.prasp# files) can take the syntactic forms listed below. In query and hypothesis files, annotation syntax is similar, with the difference that numerical weights need to be replaced with \verb|?| and that \textit{computed weights} (see below) are not allowed. In \verb|.example| files, no annotations are allowed.\\

\noindent \underline{Terminology}: an expression of the form \texttt{[\textit{W}]} or \texttt{[[\textit{W}]]} or \texttt{[[[\textit{W}]]]} in front of a plain ASP or FOL formula $f$ is called \textit{annotation} (of $f$).\\
If \texttt{\textit{W}} is a single number between 0 and 1 (inclusively) or consists of two such numbers separated by a semicolon (the second one larger than or equal to the first), \texttt{\textit{W}} is called the \textit{weight} of the formula. \texttt{\textit{W}} then directly represents either a point probability (in case \texttt{\textit{W}} is a single number) or an interval of probabilities $wx \leq Pr(f) \leq wy$ (if \texttt{\textit{W}} has the form \texttt{\textit{wx;wy}}). A weight of the form \texttt{wx;wy} is called \textit{probability interval}.\\
If \texttt{\textit{W}} is an expression of the form \verb§w|c§ where \verb§c§ is a formula and \verb§w§ is either a single number or an interval as defined above, the annotation is called a \textit{conditional} probability annotation. In that case, the weight specifies $Pr(f|c)$.\\

The annotated formula $f$ is called a \textit{weighted} formula. If the weight is neither 1 or \verb§1;1§, the formula is also called \textit{uncertain}.\\

\noindent If \texttt{\textit{W}} is \texttt{?} or \verb§?|c§, the formula is called \textit{query formula} (or \textit{query}). \\

\noindent In addition to these, some other types of annotations exist. Full list of formula annotations:

\begin{description}

\item[\texttt{[w] f}] where $w$ is a number between 0 and 1 which represents the marginal probability of formula $f$. $f$ can be ground or non-ground (containing variables). $f$ can be either in ASP- or in FOL/F2LP-syntax. \\

\noindent A few examples for well-formed \prasp formulas with non-conditional point probabilities as weights:
\begin{verbatim}
 [0.3] a.
 
 [0.7] :- a.
 
 [0.7] not a.
 
 [0.1562] a & (b | c) & d -> e | f & not g.
 
 [0.5] heads(coin(3)).

 [0.9] heads(X) :- coin(X), not lose.
 
 heads(X) :- coin(X), not lose.  % same semantics as previous formula
 
 [.3] heads(X) | tails(X) :- coin(X), X < 4.  % But note that non-ground  
 % formulas are usually annotated with other forms of weights, see further 
 % below in this section.
  
 [0.95] ![A,B]: (foo(A) | bar(B)) -> fooBar(A,B).
  
 [0.01] ?[X]: foo(X) & bar(X).
 
 [0.615] 2#count{coin_out(N,heads):coin(N)}5.
 
 [0.6] 1#sum{coin_out(N,heads):coin(N)}10 | v(X) :- not q(X,Y).

\end{verbatim}

In these examples, each probability refers to the respective formula \textit{in toto} (that is, a single probability weights the entire expansion of this formula into a set of ground formulas), not to each ground instance individually. If, for example, the domain of variable \verb§X§ is $\{1,2,3\}$, \verb§[0.5] p(X)§ declares that the probability of $p(1) \wedge p(2) \wedge p(3)$ is 0.5, which is generally not the same as $Pr(p(1))=0.5 \wedge Pr(p(2))=0.5 \wedge Pr(p(3))=0.5$. If you want to let the weight refer to individual groundings of \verb§p(X)§, use one of the multi-bracket annotations instead, as explained below. Sect. \ref{nonground} contains more details about how \prasp deals with non-ground formulas.

\item[\texttt{[[w]] f}] where $w$ is a number between 0 and 1 which represents the marginal probability of each \textit{ground} instance $f_i$ of non-ground formula $f$\footnote{If $f$ is ground, \texttt{[[w]] f} has the same meaning as \texttt{[w] f}.}. A unique index of each $f_i$ and the total number $|\{f_i\}|$ of ground instances is available in $f$ as \verb§#INDEX§ and \verb§#NUMBER§, respectively. These pseudo-variables are mainly useful in regards to \textit{computed weights} of the form \texttt{[[\textit{pred}]] f} (see further below) where a more detailed explanation is provided.\\ 

\underline{Note (1)}: the order of the generated ground instances is determined by the grounding algorithm and some post-processing (multi-token lexicographic sorting of ground formulas). It can be influenced using \hyperref[cmdline:groundingconf]{\texttt{--groundingconf}}.\\

\underline{Note (2)}: the ground instances of \texttt{[[w]] f} are generated either locally by PrASP itself (\textit{internal grounding}) or by the external ASP grounder which might apply simplifications. Again this can be influenced using \hyperref[cmdline:groundingconf]{\texttt{--groundingconf}}. In the latter case, the result of the \texttt{[[w]] f} expansion comprises only formulas and formula parts which are not simplified away by the grounder, given the so-called \textit{grounding precontext} (see Section \ref{grounding}).\\

The grounding precontext is added to formula \verb§f§ when it is grounded by an external grounder. It consists of the entire file except annotated formulas and PrASP meta-statements. Furthermore, with Gringo or Lparse as grounder, it contains automatically generated \verb§#external§ statements for all predicates found within facts or heads of rules of annotated non-query formulas, in order to protect these from being simplified away by the grounder (since their truth values are naturally not known yet at this point). If you don't want \verb§#external§ statements to be generated, surround the respective annotated formulas with \verb§#dontExternalize...#endDontExternalize§\footnote{It is also possible to put \texttt{\#external} statements manually into \prasp programs, e.g., to selectively protect certain predicates. In combination with \texttt{\#dontExternalize} this gives you full control over the externalization of predicates. Yet another option is to use switch \texttt{--keep-facts} with Gringo 4. See \hyperref[cmdline:groundingconf]{\texttt{--groundingconf}} for how to pass options to the grounder}.\\
Also, if an external grounder is being used, \prasp pools and grounds all formulas annotated with \verb§[[...]]§ using a single external grounder call, provided the annotated formulas don't require special treatment (e.g., in case of symbolic weights) and they are not grounded internally. \\
 
 For non-background knowledge files (such as \verb§.query§-files), the grounding precontext also includes all non-annotated formulas and non-PrASP meta-statements (such as \verb§#domain§ statements or Lua- or Python-script) in the background knowledge file (this behavior can be deactivated using\\ \hyperref[cmdline:groundingconf]{\texttt{--groundingconf}}).\\
 
\underline{Note (3)} For formulas where grounding is performed internally by \prasp, no grounding precontext is added to \verb§f§, however, the grounding precontext is used to determine the domains of ASP variables. In this case \prasp still calls the external grounder at one point, in order to determine the set of all ground atoms in the context. \hyperref[cmdline:groundingconf]{\texttt{--groundingconf}} allows you to change this default behavior so that only ground atoms are considered for variable domains which are explicitly provided (such as \verb§people(mary;john;anna)§), avoiding the external grounder call.\\

\underline{Note (4)}: All variables in \texttt{[[w]] f} should be bound by domain predicates in the grounding precontext (predicates which can be fully evaluated during grounding). This is not a strict rule, but adhering to it avoids hard to detect grounding errors due to incorrectly bound variables.\\

\underline{Note (5)}: In the specific context of \verb§[[..]]§ and \verb§[[[...]]]§ also formulas with intervals (\verb§..§) which expand to two or more formulas by grounding count as ``non-ground'', even if they don't contain variables.\\ E.g., \verb§[[0.5]] twoCoins(1..3,1..2)§ expands to\\
 \verb§[0.5] twoCoins(1,1)§, \verb§[0.5] twoCoins(1,2)§, \verb§[0.5] twoCoins(2,1)§, etc.\\

The above notes apply to background knowledge as well as to queries and other files in which \verb§[[...]]§ can occur, and analogously also to formulas annotated with \verb§[[[...]]]§ (see below).\\

Remark: unannotated or single-square annotated non-ground formulas such as\\
\verb§[0.8] asthma(P) :- smokes(P), person(P)§ eventually need to be grounded too, but this happens at a different stage of inference, namely when we check if formulas have certain models. In contrast to \verb§[[...]]§, their respective weight does not refer to each individual ground instance but to the entire formula (conjunction of ground instances in case of an ASP formula).\\

Here is a complete example for the correct use of \verb§[[...]]§:

\begin{verbatim}
person(1..3).
[0.2] smokes(1).
[0.3] smokes(2).
smokes(3).
[[0.8]] asthma(P) :- smokes(P), person(P).
\end{verbatim}

In this example, variable \verb§P§ is correctly bound by predicate \verb§person/1§.

On the other hand, the following code is \underline{erroneous} under the assumption that we intend to compute the probability of having asthma from smoking:

\begin{verbatim}
person(1..3).
[0.2] smokes(1).
[0.3] smokes(2).
smokes(3).
[[0.8]] asthma(P) :- smokes(P).
\end{verbatim}

While syntactically correct, the above PrASP program would give unexpected results since \verb§P§ is not bound by any domain predicate. \verb§P§ is bound by predicate \verb§smokes§, however, \verb§smokes§ is not a domain predicate (since it cannot be fully evaluated during grounding).

  Simplification of ground instances can be influenced using switch \hyperref[cmdline:groundingconf]{\texttt{--groundingconf}}, however, this option should be used with care.\\ 

It is a good idea to take a look at the resulting expansion into weighted ground instances of the original formulas using \hyperref[cmdline:showexpansion]{\texttt{--showexpansion}} if unsure about which ground formulas are actually generated. \\

\noindent Further examples:\\

\noindent Example (1):
\begin{verbatim}
coin(1).
coin(2).
coin(3).

[[0.5]] heads(X) :- coin(X).
\end{verbatim}
...expands to
\begin{verbatim}
coin(1).
coin(2).
coin(3).

[0.5] heads(3) :- coin(3).
[0.5] heads(2) :- coin(2).
[0.5] heads(1) :- coin(1).
\end{verbatim}

...which is automatically simplified\footnote{\prasp fully supports probabilistic rules, but in the example above, the ASP grounder removes the \texttt{:- coin(\_)} parts as redundancies (they are only required to determine the domain of variable \texttt{X} and not needed anymore after grounding since they are equivalent to \texttt{true} in the three example rules). } to
\begin{verbatim}
[0.5] heads(3).
[0.5] heads(2).
[0.5] heads(1).
\end{verbatim}

Example (2):

\begin{verbatim}
dice(1..2).
face(1..6).

[[0.7]] foo(D,F) & bar(#INDEX) & total(#NUMBER) <-  dice(D) & face(F).
\end{verbatim}

...expands to (using Gringo 4\footnote{Note that we would get a different order of ground instances using Gringo 3. }):
\begin{verbatim}
dice(1..2).
face(1..6).

[0.7] 3{foo(1,1);bar(1);total(12)}3.
[0.7] 3{foo(1,2);bar(2);total(12)}3.
[0.7] 3{foo(1,3);bar(3);total(12)}3.
[0.7] 3{foo(1,4);bar(4);total(12)}3.
[0.7] 3{foo(1,5);bar(5);total(12)}3.
[0.7] 3{foo(1,6);bar(6);total(12)}3.
[0.7] 3{foo(2,1);bar(7);total(12)}3.
[0.7] 3{foo(2,2);bar(8);total(12)}3.
[0.7] 3{foo(2,3);bar(9);total(12)}3.
[0.7] 3{foo(2,4);bar(10);total(12)}3.
[0.7] 3{foo(2,5);bar(11);total(12)}3.
[0.7] 3{foo(2,6);bar(12);total(12)}3.
\end{verbatim}

Example (3) - see Sect. \ref{markov} for details:

\begin{verbatim}
friend(a,b).
friend(b,c).

#dontExternalize
[[0.7]] influences(X, Y) :- friend(X, Y).
#endDontExternalize

influences(X, Y) :- influences(X, Z), influences(Z, Y).
\end{verbatim}

To expand a non-probabilistic non-ground formula to its ground instances, annotate it with weight \verb§[[1]]§. But notice that this is not required to handle non-ground formulas. Example:

\begin{verbatim}
person(a;b;c).

#domain person(X).
#domain person(Y).

[[1]] friend(X,Y).
\end{verbatim}

expands to

\begin{verbatim}
[1] friend(a,a).
[1] friend(a,b).
[1] friend(a,c).
[1] friend(b,a).
[1] friend(b,b).
[1] friend(b,c).
[1] friend(c,a).
[1] friend(c,b).
[1] friend(c,c).
\end{verbatim}

With the program above you would obtain the same query results as with the following \prasp program:

\begin{verbatim}
person(a;b;c).

#domain person(X).
#domain person(Y).

friend(X,Y).
\end{verbatim}

The only difference here is that the latter program is not grounded by \prasp itself. \prasp rather checks for each stable model (there is only one) whether the non-ground formula holds in that model (however, to check this, the external grounder needs to ground \verb§friend(X,Y)§). In contrast, with the second program, \prasp first grounds \verb§friend(X,Y)§ and later checks for each ground instance whether it holds in the only stable model (which is the same as the stable model of the second program).

\item[\texttt{[w|c] f}] where $w$ is a number between 0 and 1 which represents conditional probability $Pr(f|c)$. $f$ and $c$ are formulas (ground or non-ground). \\ 

\noindent Example (1):
\begin{verbatim}
coin(1).
coin(2).
coin(3).

[0.8|heads(X) :- coin(X)] win :- coin(X).
\end{verbatim}

Example (2), which shows that conditions can also be rules (although in practice they are typically (ground) facts):
\begin{verbatim}
[0.3|D <= 2 :- dice(D)] faceObserved(F) :- face(F).
\end{verbatim}

\item[\texttt{[[w]] condProb(f,c) :- b }] where $w$ is a number between 0 and 1 which represents conditional probability $Pr(f_i|c_i)$ for any pair $(c_i, f_i)$ of corresponding ground instances $f_i$, $c_i$ of \textit{atomic} formulas $f$ and $c$ such that $b$ holds \textit{and} $b$ can be evaluated during grounding. \verb§condProb§ is a reserved keyword in \prasp which is treated by the ASP grounder as a predicate although it is actually just a means to combine$f$ and $c$ in one formula.\\
The purpose of this syntax is to generate the list of all groundings of $f|c$ where the variables in $f$ and $c$ are restricted using some expression $b$ which affects both $f$ and $c$. Groundings of conditional probabilities can also be generated using \verb§[[w|c]] f§ or \verb§[[[w|c]]] f§ (see below), however, there the two formulas $f$ and $c$ are expanded separately from each other.\\  

The statement about the grounding-context for \texttt{[[w]] f} above also applies here (and other \verb§[[...]]§/\verb§[[[...]]]§ style annotated formulas).\\

Both $f$ and $c$ must be atomic formulas here. Observe the double square brackets around the weight.\\

To see the result of the expansion, use command-line switch \hyperref[cmdline:showexpansion]{\texttt{--showexpansion}}.\\

\noindent Example (Gringo3/Lparse version):

\begin{verbatim}
x(1..3).
#domain x(I).
#domain x(J).
#domain x(K).
[[0.3]] c(K).
[[0.5]] condPr(h(I,J),c(K)) :- J == K.
\end{verbatim}

The annotated formulas above expand to 

\begin{verbatim}
[0.3] c(1).
[0.3] c(2).
[0.3] c(3).

[0.5|c(3)] h(1,3).
[0.5|c(2)] h(1,2).
[0.5|c(1)] h(2,1).
[0.5|c(3)] h(3,3).
[0.5|c(2)] h(2,2).
[0.5|c(2)] h(3,2).
[0.5|c(1)] h(1,1).
[0.5|c(1)] h(3,1).
[0.5|c(3)] h(2,3).
\end{verbatim}

Alternatively, we can encode this example as follows. This
version works with both Gringo3 and Gringo4 (since it doesn't use \verb§#domain§):

\begin{verbatim}
x(1..3).
[[0.3]] c(K).
[[0.5]] condPr(h(I,J),c(K)) :- J == K, x(I), x(J), x(K).
\end{verbatim}

\item[\texttt{[[w|c]] f}] where $w$ is a number between 0 and 1 which represents conditional probability $Pr(f_i|c_i)$ for any pair $(c_i, f_i)$ of corresponding ground instances $f_i$, $c_i$ of formulas $f$ and $c$. $f_i$ and $c_i$ correspond if they have the same position in the sequence of ground instances returned by the ASP grounder for $f$ or $c$, respectively. The number of ground instances of $c$ and $f$ must be identical.\\
Exception: if $f_1$ is the only ground instance of $f$, the pairings are $(c_1, f_1), (c_2, f_1), ..., (c_n, f_1)$.\\ 

Both $f$ and $c$ can be any formulas, including rules.\\

Observe that if $f$ is a rule, the rule body doesn't affect $c$. So, in order to restrict the groundings using some boolean expression $b$ (in form of a rule body which can be resolved during grounding), this body needs to be present in $f$ as well as $c$. See Example (3) below. Where this is not possible, use the \verb§condProb(f,c) :- b§ syntax instead (see above).\\   

The statement about the grounding-context for \texttt{[[w]] f} above also applies here (and other \verb§[[...]]§/\verb§[[[...]]]§ style annotated formulas).\\

To see the result of the expansion, use command-line switch \hyperref[cmdline:showexpansion]{\texttt{--showexpansion}}.\\

\noindent Example 1 (Gringo3/Lparse-version):
\begin{verbatim}
person(1..3).

#domain person(P).

[[0.3|smokes(P)]] asthma(P).

[0.1] smokes(1).
smokes(2).
[0.85] smokes(3).

\end{verbatim}

\noindent Example 1 (Gringo4/ASP language standard-version):
\begin{verbatim}
person(1..3).

[[0.3|smokes(P) :- person(P)]] asthma(P) :- person(P).

[0.1] smokes(1).
smokes(2).
[0.85] smokes(3).
\end{verbatim}

Example 1 expands to

\begin{verbatim}
[0.3|smokes(1)] asthma(1).
[0.3|smokes(2)] asthma(2).
[0.3|smokes(3)] asthma(3).
...
\end{verbatim}

\noindent Example (2):
\begin{verbatim}
person(1).
person(2).
person(3).

[[0.3|smokes(P) :- person(P)]] some_asthma.
\end{verbatim}
...expands to

\begin{verbatim}
[0.3|smokes(1)] some_asthma.
[0.3|smokes(2)] some_asthma.
[0.3|smokes(3)] some_asthma.
\end{verbatim}

\noindent Example (3):
\begin{verbatim}
person(1).
person(2).
person(3).

[[0.3|smokes(P) :- person(P), P != 2]] asthma(P) :- person(P), P != 2.
\end{verbatim}
...expands to

\begin{verbatim}
person(1).
person(2).
person(3).

[0.3|smokes(1)] asthma(1).
[0.3|smokes(3)] asthma(3).
\end{verbatim}

\item[\texttt{[[[w|c]]] f}] where $w$ is a number between 0 and 1 which represents conditional probability $Pr(f_i|c_i)$ of \textit{any} combination of ground instances $f_i$, $c_i$ of formulas $f$ and $c$. \\ 

Observe that if $f$ is a rule, the rule body of $f$ doesn't affect $c$. So, in order to restrict the groundings using some boolean expression (in form of a rule body which can be resolved during grounding), this body needs to be present in $f$ as well as $c$. See Example (2) below. Where this is not possible, consider using the \verb§condProb(f,c) :- b§ syntax instead (where possible, see above).\\   

The statement about the grounding-context for \texttt{[[w]] f} above also applies here (and other \verb§[[...]]§/\verb§[[[...]]]§ style annotated formulas).\\

To see the result of the expansion, use command-line switch \hyperref[cmdline:showexpansion]{\texttt{--showexpansion}}.\\

\noindent Example:
\begin{verbatim}
person(1).
person(2).
person(3).
#domain person(P).

[[[0.3|smokes(P)]]] asthma(P).
\end{verbatim}
...expands to

\begin{verbatim}
[0.3|smokes(3)] asthma(3).
[0.3|smokes(2)] asthma(3).
[0.3|smokes(1)] asthma(3).
[0.3|smokes(3)] asthma(2).
[0.3|smokes(2)] asthma(2).
[0.3|smokes(1)] asthma(2).
[0.3|smokes(3)] asthma(1).
[0.3|smokes(2)] asthma(1).
[0.3|smokes(1)] asthma(1).
\end{verbatim}

\noindent Example (2):
\begin{verbatim}
person(1).
person(2).
person(3).

[[[0.3|smokes(P) :- person(P), P != 2]]] asthma(P) :- person(P), P != 2.
\end{verbatim}
...expands to

\begin{verbatim}
person(1).
person(2).
person(3).

[0.3|smokes(1)] asthma(1).
[0.3|smokes(3)] asthma(1).
[0.3|smokes(1)] asthma(3).
[0.3|smokes(3)] asthma(3).
\end{verbatim}

(...which is probably not very useful, but exemplifies the difference to the effect of using double square brackets or \verb§condProb§.)

\item[\texttt{[\textit{pred}] f}]  where \textit{pred} is the predicate name of an unary atom which occurs in formula $f$. The meaning is as that of the corresponding form with a numerical literal as weight, however, here the numerical weight is computed (taken from the argument of \textit{pred} computed during the grounding phase). In others words, this syntax allows for \textit{computed weights}.\\
Since most ASP grounders don't support fractions, the weight is the argument of \textit{pred} divided by 100000.\\

The part of formula $f$ starting from the first occurrence of \textit{pred(...)} until the end of the formula is afterwards replaced with \verb#true.# Therefore, \textit{pred(...)} should either be the last literal in the body of a rule $f$, or otherwise any other literal following \textit{pred(...)} in the body should be used only for determining the weight, as it will be removed.\\

Requirements: all ground instances of $f$ need to agree on the value of the argument of \textit{pred}. At least one atom of the form \textit{pred(...)} occurs in the ground instances of $f$ and has a numerical argument which is between 0 and 1 after division by 100000.\\

Example (1) (where \textit{pred} = \verb#wght#):

\begin{verbatim}
person(1).
person(2).
person(3).
#domain person(P).

[wght] asthma(P) :- wght(50000/2).
\end{verbatim}

...expands to

\begin{verbatim}
#domain person(P).

[0.25] asthma(P) :- true.
\end{verbatim}

\noindent Note that variable \verb#P# is not expanded (since we used single square brackets around the weight).\\

Another example (2) lets the weight depend on some other predicate defined elsewhere:

\begin{verbatim}
person(1).
person(2).
person(3).
#domain person(P).

#volat

arg(1..1000).

inverseP(I, N) :- I = 100000 / N, arg(N).

#endVolat

[weight] asthma(P) :- weight(W), inverseP(W, 100).
\end{verbatim}

...is expanded to

\begin{verbatim}
#domain person(P).

[0.01] asthma(P) :- true.
\end{verbatim}

Since \verb#inverseP# is only needed to compute the weight, we put its definition into a \verb|#volat| segment (see Sect. \ref{meta}), which causes it to be removed after expanding the weighted formula. 

\item[\texttt{[[\textit{pred}]] f}] 

Analogous to \texttt{[\textit{pred}] f} but such that the weight refers to each ground instance of \verb§f§. As with a numerical weight in double square brackets, $f$ is expanded into its ground instances, but the weight is taken from the argument of the first \textit{pred} atom in these instances.\\
Because sometimes it is helpful to have the number of ground instances of $f$ and an index (unique numerical index of a certain ground instance) available, \prasp exposes these values as \verb|#NUMBER| and \verb|#INDEX| (see example (2) below), albeit these pseudo-variables can be used with numerical weights in double or triple square annotations too.\\

The statement about the grounding-context for \texttt{[[w]] f} above also applies here (and other \verb§[[...]]§/\verb§[[[...]]]§ style annotated formulas).\\

To see the result of the expansion, use command-line switch \hyperref[cmdline:showexpansion]{\texttt{--showexpansion}}.\\

Example (1):

\begin{verbatim}
person(1).
person(2).
person(3).
#domain person(P).

[[wght]] asthma(P) :- wght(50000/2).
\end{verbatim}

...expands to:

\begin{verbatim}
[0.25] asthma(1) :- true.
[0.25] asthma(2) :- true.
[0.25] asthma(3) :- true.
\end{verbatim}

Example(2):

Weight annotations can also be computed in a procedural or functional style using a scripting language. Here we show how the desired distribution of weights can be specified using the scripting language \textit{Lua} \cite{Lua} which is integrated in Gringo and Clingo. This is particularly useful if some specific prior distribution over weights are required in background knowledge for which no built-in random generators exist. Besides numerical computations, Lua can also be used for interfacing the ASP grounder with a database. For details please refer to the Gringo/Clingo/Lua documentation. \\
\noindent Gringo 4 and thus \prasp also support Python scripts if Gringo was compiled with the appropriate \\
option (see \verb§http://potassco.sourceforge.net/§).

\begin{verbatim}
#volat
#begin_lua

function binomial(n, k)  
    if k > n then return nil end
    if k > n/2 then k = n - k end 
    numer = 1
    denom = 1
    for i = 1, k do
        numer = numer * (n - i + 1)
        denom = denom * i
    end
    return numer / denom
end

function binDistr(k, p, n)
  return binomial(n, k) * math.pow(p, k) * math.pow(1-p, n-k)
end

function binDistrA(k,n)
  p = 0.5
  return binDistr(k, p, n) * 100000
end

#end_lua.

person(1).
person(2).
person(3).

#endVolat

[[_wgt]] blob(X,Y) :- _wgt(@binDistrA(#INDEX,#NUMBER)), person(X), person(Y).
\end{verbatim}

...expands to (after simplification):

\begin{verbatim}
[0.01758] blob(1,1).
[0.07031] blob(1,2).
[0.16406] blob(1,3).
[0.24609] blob(2,1).
[0.24609] blob(2,2).
[0.16406] blob(2,3).
[0.07031] blob(3,1).
[0.01758] blob(3,2).
[0.00195] blob(3,3).
\end{verbatim}

What happens here is that each ground formula (i.e., the expansion of\\
\verb|[[_wgt]] blob(X,Y) :- _wgt(@binDistrA(#INDEX,#NUMBER)), person(X), person(Y)|) is annotated with a weight computed using Lua-function \verb#binDistrA()#. Also, everything from \verb§_wgt§ on is replaced by \verb§true§. In effect, the ground instances of \verb§blob/2§ are annotated according to the binomial distribution $B(\mathtt{\#NUMBER}, 0.5)$.\\

For each ground instance, \verb|#INDEX| is replaced by a unique numerical index. The respective index is then passed as argument to \verb#binDistrA()#, together with the total number of ground instances \verb|#NUMBER|. The instances of \verb|#INDEX| are simply the numbers 1,2,3,..., assigned in the reverse order in which the grounder provides the ground instances $f_i$ of $f$ (so you can rely on their uniqueness, but the pairings (index, $f_i$) depend on the grounder-specific grounding algorithm).

\item[\texttt{[\textit{pred}|c] f}] (analogous to above. Note that it is not possible to obtain the weight from some occurrence of \textit{pred\texttt{}} within the condition $c$.)
\item[\texttt{[[\textit{pred}|c]] f}] (analogous to above)
\item[\texttt{[[[\textit{pred}|c]]] f}] (analogous to above)

\item[\texttt{[wx;wy] f}]\index{Probability intervals} where $wx$ and $wy$ are numbers between 0 and 1 (inclusive) and $wy \geq wx$ (if $wy = wx$, the annotation corresponds to \verb§[wx]§).\\
This annotation specifies that the probability of \verb§f § lies in interval $[wx,wy]$ \footnote{Not to be confused with \textit{Bayesian probability intervals} (\textit{credible intervals}) or \textit{confidence intervals}. In \prasp, a probability interval is a higher-order probability defined as a range of probabilities between a lower and an upper boundary. In contrast to credible intervals, these boundaries are probabilities. }. See Sect. \ref{intervalweights}.\\

\item[\texttt{[wx;wy|c] f}] where $wx$ and $wy$ are numbers between 0 and 1 (inclusive). Analogously to \texttt{[w|c] f}, with the difference that we specify that $wx \leq Pr(f|c) \leq wy$.

\item[\texttt{[[:]] f}] expands to the list of all ground instances of $f$, each with weight $1/n$, where $n$ is the number of ground instances.\\

The statement about the grounding-context for \texttt{[[w]] f} above also applies here (and other \verb§[[...]]§/\verb§[[[...]]]§ style annotated formulas).\\

To see the result of the expansion, use command-line switch \hyperref[cmdline:showexpansion]{\texttt{--showexpansion}}.\\

Example: 

\begin{verbatim}
face(1..6).

[[:]] faceObserved(F) :- face(F).
\end{verbatim}

...expands to

\begin{verbatim}
[0.16666666666666666] faceObserved(1).
[0.16666666666666666] faceObserved(2).
[0.16666666666666666] faceObserved(3).
[0.16666666666666666] faceObserved(4).
[0.16666666666666666] faceObserved(5).
[0.16666666666666666] faceObserved(6).
\end{verbatim}

\item[\texttt{[[[:]]] f}] (analogous to above)
\item[\texttt{[[:|c]] f}] (analogous to above)
\item[\texttt{[[[:|c]]] f}] (analogous to above)

\item[\texttt{[\#rnd] f}] Analogous to \texttt{[w] f}, but with uniformly distributed pseudo-random weight.
\item[\texttt{[[\#rnd]] f}] Analogous to \texttt{[[w]] f} and above.
\item[\texttt{[\#rnd|c] f}] Analogous to \texttt{[w|c] f} and above.
\item[\texttt{[[\#rnd|c]] f}] Analogous to \texttt{[[w|c]] f} and above.
\item[\texttt{[[[\#rnd|c]]] f}] Analogous to \texttt{[[[w|c]]] f} and above.

\item[\texttt{[\#rndn($\mu$,$\sigma$)] f}] Analogous to \texttt{[w] f} and \texttt{[\#rnd] f}, but with normally (Gaussian) distributed pseudo-random weight. $\mu$ is the mean and $\sigma$ is the variance of the normal distribution (given as plain numbers, not as symbols).

\item[\texttt{[[\#rndn($\mu$,$\sigma$)]] f}] Analogous to \texttt{[w] f} and above.
\item[\texttt{[\#rndn($\mu$,$\sigma$)|c] f}] Analogous to \texttt{[w|c] f} and above.
\item[\texttt{[[\#rndn($\mu$,$\sigma$)|c]] f}] Analogous to \texttt{[[w|c]] f} and above.
\item[\texttt{[[[\#rndn($\mu$,$\sigma$)|c]]] f}] Analogous to \texttt{[[[w|c]]] f} and above.

\item[\texttt{[.] f}] represents $f \vee \neg f$, i.e., the spanning formula for $f$. It can be used to express that neither the truth value of $f$ nor its probability is known, but that possible worlds should be generated which cover both $f$ and $\neg f$ (unweighted nondeterminism).\\
 $\neg$ stands for default negation, or for classical negation if command-line option \hyperref[cmdline:strongnegbelief]{\texttt{--strongnegbelief}} is used.\\

Notice that this does not enforce a probability of 0.5 of \verb|f|. If you would like to formalize an uncertain statement with ``unbiased'' probability, consider \verb§[0.5] f§ which has a different semantics.\\

Use double-square syntax \verb|[[.]] f.| to specify that \verb#(fg#$_i$ \verb#| not fg#$_i$\verb|)|  for each ground instance \verb|fg|$_i$ of \verb|f|. \\
In contrast, \verb|[.] f.| for a non-ground formula \verb|f| specifies that \verb#(f) | not (f)# holds for \verb#f# as a whole (i.e.,  \verb#(fg1 & fg2 & ...) | not (fg1 & fg2 & ...)# if the  \verb|fg|$_i$ are the ground instances of  \verb|f|).

\item[\texttt{[[.]] f}] (see above)

\item[\texttt{[?] f}] Analogous to \texttt{[w] f}, but in query files (i.e., we are looking for the probability of query formula \verb§f§).
\item[\texttt{[[?]] f}] Analogous to above.
\item[\texttt{[?|c] f}] Analogous to above.
\item[\texttt{[[?|c]] f}] Analogous to above.
\item[\texttt{[[[?|c]]] f}] Analogous to above.

\end{description}

\subsection{Spanning formulas and programs}
\label{spanGen}

The \textit{spanning program} (Sect. \ref{semantics}) consists of the \textit{spanning formulas} and the unweighted formulas in the background knowledge. Each weighted formula in the background knowledge has exactly one corresponding spanning formula. Informally, the spanning formula reflects that the respective weighted formula holds or does not hold, i.e., that it represents an uncertain event. The answer sets of the spanning program are the possible worlds of the \prasp program.\\
A single spanning formula \cite{prasp12,prasp13} can actually consist of multiple ASP rules.\\

Generation of the ASP rules in spanning formulas might introduce unsafe variables (variables not bound by domain predicates, leading to grounding errors later). With Gringo 3, the domains of variables in weighted formulas can be declared using \verb§#domain§ statements. Where these are absent or if Gringo 4 or higher is being used, \prasp automatically attempts to determine the domains of variables and bind them properly, more precisely: it identifies domain predicates in the program which bind these variables and adds them to the bodies of the generated ASP rules of which the spanning formula consists. This requires that the user has already bound these variables using domain predicates in the original formula. To identify which of the predicates in the original formula are domain predicates, \prasp uses a heuristics: if the respective predicate is defined in the program and never occurs in a uncertain position (e.g., head of a rule), it is considered as domain predicate. Use switch \verb§--verbose§ to see which variables bindings are added to the spanning formula. \\

\noindent You can influence how spanning formulas are generated using command line option \hyperref[cmdline:spangenconf]{\texttt{--spangenconf}}. \\

\noindent [More tbw.]

\subsection{About grounding}
\label{grounding}

Grounding of formulas means that a formula with variables (a so-called \textit{non-ground} formula) is converted into an equivalent set of formulas without variables (the ground instances). Grounding happens at various steps in the inference pipeline. The most important cases:

\begin{itemize}
	\item Grounding of formulas annotated with \verb§[[...]]§ or \verb§[[[...]]]§. This is done as a preparation step before the spanning program is being generated. Each such formula is grounded individually in its so-called \textit{grounding precontext}. See Sect. \ref{weights} for details.
	
	\item Grounding of \verb§[...]§-weighted non-ground formulas in order to determine whether it holds in a certain possible world. This happens when \prasp computes the probabilities of possible worlds. The context of this type of grounding is the spanning program. 
\end{itemize}

You can influence how grounding is performed using command line option \hyperref[cmdline:groundingconf]{\texttt{--groundingconf}}.\\

\noindent [More tbw.]

\subsection{Meta-statements}
\label{meta}

In addition to those meta-statements supported by the ASP grounder which is being used (such as \verb|#domain|), \prasp introduces
a few further meta-statements. All of them can be used in \verb§.prasp§-files and some of them also in other kinds of input files, as detailed below.\\

\noindent Note the absence of a dot in the following meta-statements which is in contrast to Gringo-style meta-statements like \verb§#domain§.

\begin{description}
	
\item[Most grounder meta-statements] such as Gringo/Lparse's \verb§#external§ or Gringo's \verb§#script§ can be used with \prasp too, with the following restrictions:

\begin{itemize}
	\item Scripts must be placed at the beginning of a file. There cannot be more than one Lua or Python script per file.
	\item \verb§#compute§ and \verb§#program§ are not supported (there might be support for \verb§#program§ in the future though).
	\item Some meta-statements are supported natively by \prasp, e.g., \verb§#include§ is processed by \prasp and not submitted to the grounder.
\end{itemize} 

\item[\texttt{\#include "}\textit{file}\texttt{"}] inserts the contents of \textit{file} at this place. Any \texttt{\#include} statements in  \textit{file} are recursively resolved, provided \textit{file} ends with \verb§.prasp§, \verb§.hypoth§, \verb§.query§, or \verb§.examples§. \texttt{\#include} is handled by \prasp directly, not by the grounder (although Gringo also understands \texttt{\#include}). It can be used in .prasp, .query, .hypoth and .examples files.\\
\texttt{\#include "}\textit{file}\texttt{"} must occupy a line of text of its own. Note that in contrast to Gringo's \verb§#include§ there is no dot at the end of the line.

\item[\texttt{\#def} \textit{name} \texttt{=} \textit{content}] defines a non-parametric text substitution macro \textit{name} with content \textit{content}. Everywhere below the \verb§#def§ statement in the same file, all occurrences of word \textit{name} are replaced with \textit{content}, with the exception of meta-keywords (e.g., \verb§#def§ or \verb§#domain§) and the \verb§name§ in other macro definitions.\\
\textit{name} is a sequence of one or more alphanumerical characters or underscores. A  non-parametric macro can be redefined later in the file, although this is not recommended (\prasp issues a warning). A  non-parametric macro definition reaches until the end of the line or the start of a comment (i.e., character \%), whichever comes first.\\

 Non-parametric macro definitions can be placed in all kinds of input files (background knowledge, query files, and files with learning examples or hypotheses).\\ However, they don't automatically track over from one file to another: if you want to use the same definition in two files, you need to place it in both (or use \verb§#include§). 

Remark: \prasp itself does not limit the length of lines or formulas. However, external tools such as Gringo or F2LP might impose a maximum size of lines of text, which automatically also limits the length of formulas (please see the documentation of these tools for details).\\

\noindent Example:

\begin{verbatim}
#def evidence1 = 2#count{sm(2),if(1,2)}2

	[?|evidence1] sm(1).
	[?|evidence1] sm(2).
	[?|evidence1] sm(3).
	[?|evidence1] sm(4).
	
#def evidence2 = evidence1 & sm(4)
	
	[?|evidence2] ast(2).
	[?|evidence2] ast(3).
	[?|evidence2] ast(4).  
\end{verbatim}

...expands to:

\begin{verbatim}
[?|2#count{sm(2),if(1,2)}2] sm(1).
[?|2#count{sm(2),if(1,2)}2] sm(2).
[?|2#count{sm(2),if(1,2)}2] sm(3).
[?|2#count{sm(2),if(1,2)}2] sm(4).
[?|2#count{sm(2),if(1,2)}2 & sm(4)] ast(2).
[?|2#count{sm(2),if(1,2)}2 & sm(4)] ast(3).
[?|2#count{sm(2),if(1,2)}2 & sm(4)] ast(4).
\end{verbatim}

Note(1): you could likewise place the evidence atoms directly in the query file.\\
Note(2): in this and most other examples, we use Gringo 3 syntax for aggregates. If you want to use them with Gringo/Clingo 4, please refer to the Clingo 4 manual.\\

\item[\texttt{\#scala...\#endScala}] allows the user to include a number of Scala method definitions which can be applied during parsing of background knowledge or other \prasp files (\textit{parametric Scala-based macros}). The results of the Scala method calls replace pieces of text in the file which contains this meta-statement. This meta-statement can be used in all kinds of \prasp files.\\

The included Scala methods are applied in the remaining file using \$\textit{fnName}\verb§(text)§ where \textit{fnName} is the name of the respective user-defined Scala method.\\

The signature of each method defined in \texttt{\#scala...\#endScala} must be\\ 
\textit{fnName}\verb§(text: String, startPos: Int, endPos: Int, context: String): String§, where \textit{fnName} is a new Scala method name, parameter \textit{text} contains the text piece which should be replaced, \textit{startPos} contains the start index of \$\textit{fnName}\verb§(text)§ within the overall PrASP file (after processing of any \verb§#include§ statements and comments removal), \verb§endPos§ contains the end index, and \verb§context§ contains the overall file contents. \$\verb§fnName(text)§ is replaced with the result of the method call, which must be of Scala type \verb§String§. In the method definitions, the \prasp API can be used.\\

Scala-based macros are resolved after removal of comments and processing of \verb§#include§, and before any other modifications (even before any \verb§#def§'s are processed).\\ 

The compiled Scala function definitions are cached, that is, any call of a certain such function (regardless of its context) will use the same function body to evaluate its arguments.\\

There can be only one \texttt{\#scala...\#endScala} section per file.\\

The main use case of this feature is the inclusion of experimental parsing features ``on-the-fly'' without the need to manually invoke the Scala compiler or to build a jar-file. The Scala compiler and runtime library necessary to resolve such macros are included with \prasp, so no extra installation is required. However, this feature comes at the price that parsing the file which contains \texttt{\#scala...\#endScala} might take a few seconds (because the Scala compiler needs to be launched), depending on the speed of your computer.\\ 

Scala-based macros are not related to Gringo's Lua and Python scripting support mentioned earlier. \\

\noindent Example:\\

\PrASP already understands Annotated Disjunctions (ADs) out-of-the-box (see Sect. \ref{annotatedDisjunctions}), but let's assume that, for some reason, you would like to provide your own approach to the de-sugaring of ADs, implemented as a small Scala script. The following example adds a scripted implementation of Annotated Disjunctions to a \prasp program (with \verb§my::-§ as the rule operator, to distinguish these constructs from the built-in ADs):

\begin{verbatim}
#scala	  
#include "examples/annotatedDisjunctions.scala"
#endScala

[0.12] b1.
:- b2.
[0.8] b3.

$ad( [0.1] h1; [0.33] h2 my::- b1, b2, b3. )  % an annotated disjunction, 
%   translated into a set of regular PrASP formulas using Scala-method ad()

\end{verbatim}

File \verb§annotatedDisjunctions.scala§ contains the Scala source code of user-defined function \verb§ad§ which handles the statement within  \texttt{\$ad(...)}:

\begin{verbatim}
def ad(rr: String, pos: Int, endPos: Int, prog: String): String = {

    ... (see examples/annotatedDisjunctions.scala for the full code)

}
\end{verbatim}

\item[\texttt{\#volat...\#endVolat}] in background knowledge files has the effect that everything between \texttt{\#volat} and \texttt{\#endVolat}
is removed from the background knowledge after double and triple square brackets have been expanded to ground formulas and 
\textit{computed weights} have been determined (see Sect. \ref{weights} for examples).

\item[\texttt{\#dontExternalize}\texttt{...\#endDontExternalize}] in background knowledge specifies that all predicates in the heads of annotated formulas between \verb§#dontExternalize§ and \verb§#endDontExternalize§ should be exempt from being excluded from being simplified away during grounding. To selectively externalize certain predicates (within or outside \texttt{\#dontExternalize}\texttt{...\#endDontExternalize}), manually add \verb§#external§ statements to background knowledge.\\

\noindent Example:

\begin{verbatim}
#dontExternalize

[[0.72]] p(X), q(Y) :- dp(X), dq(Y).
#external p(X) : dp(X).

#endDontExternalize
\end{verbatim}

In this example, only predicate p/1 will be externalized. Grounding of formulas which contain \verb§p§ (but not \verb§q§) will therefore not assume that \verb§p§ is undefined and could be omitted.

\item[Independence declarations] Independence declarations are optional - they are not required for inference or learning, irrespectively of the presence or absence of actual probabilistic independences among formulas (i.e., random events). Depending on the concrete sampling and inference approach, \prasp can discover or assume independence, or provide solutions which are correct even if weighted formulas are not mutually or pairwise independent. However, independence declarations provide\prasp with extra information which allows some approaches (\hyperref[cmdline:mod0]{\texttt{--mod0}}) to perform better (faster and/or giving less biased solutions), not unlike the independence constraints imposed by the structures of graphical models. Algorithms might be a magnitude faster if event independence can be exploited. On the other hand, there are algorithms, in particular \hyperref[cmdline:simanneal]{\texttt{--simanneal}} and linear system solving and linear programming, which currently don't profit from declared independence but slow down (as they need to consider additional constraints - in this regard, see switch \hyperref[cmdline:noindepconstrs]{\texttt{--noindepconstrs}}).\\

Conditional as well as non-conditional independence of events (formulas) can also be specified using conditional probabilities (such that $Pr(x) = Pr(x|y)$), however, explicitly declared independence is processed faster.\\

\prasp can also discover independence automatically. Informally, this works by assuming that any formulas which are not related logically (via rules) are independent from each other. However, the benefits of declared independence over automatically discovered independence are that the latter is not necessarily complete (for performance reasons it uses a heuristic approach) and that it requires time for analyzing the background knowledge. \\

If independence is neither declared or discovered (nor indirectly specified using formula weights) for a set of formulas, their (in)dependence is considered undefined and query results which combine such events are in a range of possible probabilities (if entropy is not maximized).\\  

There are several types of independence which can be declared in \prasp:
\begin{description}
	\item[Unconditional mutual independence] See below
		\item[Unconditional pairwise independence] See below
			\item[Conditional independence] Needs to be specified using conditional probabilities.
		
\end{description}
 \begin{description}

\item[\texttt{\#indep...\#endIndep}]-sections in background knowledge files declare that the set $FG$ of all(*) formulas between \verb§#indep§ and \verb§#endIndep§ are \textit{mutually} and unconditionally independent (in the probabilistic sense). \\

(*)Restriction: these formulas need to have a \textit{specific weight} (or probability 1), i.e., annotation with \verb§[.]§ is not allowed within an independence declaration. This restriction might disappear in future versions of \prasp.\\

If linear programming is used, a \texttt{\#indep...\#endIndep} section imposes the following constraint:\\
\label{indep}

$\forall \{SF_i\} \subseteq FG, |\{SF_i\}| \geq 2: Pr(\bigwedge {SF_i}) = \prod Pr(SF_i)$.\\

The $SF_i$ must be distinct (\prasp does not check this). Observe that any formulas for which the above holds are also uncorrelated and that $FG$ is also pairwise independent (but pairwise independent formulas are not necessarily mutually independent!). If you require pairwise independence only, use \texttt{\#pIndep...\#endPIndep} (please see below).\\

Naturally, the tractability of \texttt{\#indep...\#endIndep} and similar meta-statements is prone to combinatorial explosion. Ways to cope with this are to use \hyperref[cmdline:limitindepcombs]{\texttt{--limitindepcombs}} or the application of entropy maximization approaches (in connection with near-uniform sampling of initial models) instead of independence declarations (see \ref{intervalresults}). \\

With \hyperref[cmdline:ignoredeclindeps]{\texttt{--ignoredeclindeps}}, \texttt{\#indep...\#endIndep} and similar meta-statements are entirely ignored.\\

Independence declarations can be useful because by default, \prasp does \textit{not} assume independence of all events (propositions). More precisely, by default,
\prasp uses the dependency graph emitted by the ASP grounder to automatically identify ground atoms which are presumed to be independent from all other atoms. For the set of these facts, mutual independence is assumed, as well as mutual / pairwise independence of these facts and any user-declared mutually / pairwise independent formulas. \\
But if the independence of a set of events is neither manually specified nor automatically discovered, \prasp assumes that the kind of dependence or independence of this set is \textit{unspecified}. Furthermore, the automated discovery of independent formulas is
deliberately lightweight (a precise discovery process would take a lot of time) and might fail to discover all independences.\\

Events whose interdependence is unspecified can lead to an under-determination of the probability distribution over possible worlds - queries can then have \textit{ranges of values} instead of point probabilities as results. In that case, \prasp returns either \textit{one} of the possible values from the respective range (and optionally looks for a possible worlds' probability distribution with maximum entropy), or it returns the entire interval of valid solutions for this query, or an approximation thereof. See \ref{intervalresults} for details.\\ 

%Besides their influence on query and learning results, declared as well as automatically discovered independence 
%also massively influences inference and learning speed. In certain cases, this influence is positive (see \ref{performance}), because combinations of mutually independent formulas can be simplified in ways which are not possible for combinations of other formulas. On the other hand, it is very costly to enforce event independence using additional constraints derived from #indep meta-statements.\\

Apart from their independence, the formulas within \texttt{\#indep...\#endIndep} are treated as regular background knowledge (and they have normally weights attached). They can be facts as well as rules.\\
A \verb§.prasp§ file can contain multiple \texttt{\#indep...\#endIndep} segments.\\ 

Conditional probability constraints (i.e., annotated formulas where the weight contains the form
$[w|condition]$) are not allowed within \texttt{\#indep...\#endIndep}. Typically (but not necessarily), independent formulas carry a probability annotation (\prasp shows a warning if the weight is missing, as this is probably a typo).\\

In case command-line switches \hyperref[cmdline:mod0]{\texttt{--mod0}} or \hyperref[cmdline:mod1]{\texttt{--mod1}} are provided, only those formulas are considered which have ``survived'' the background knowledge simplifications triggered by these switches.\\

Example:

\begin{verbatim}
coin(1..3).
#indep
[0.6] coin_out(1,heads).
[[0.5]] coin_out(N,heads) :- coin(N), N != 1.
#endIndep
1{coin_out(N,heads), coin_out(N,tails)}1 :- coin(N).
n_win :- coin_out(N,tails), coin(N).
win :- not n_win.
[0.8|win] happy.
:- happy, not win.
\end{verbatim}

In this model of a coin flipping game, the block \verb§#indep§ ... \verb§#endIndep§ declares that
the outcomes of the three coins shall be mutually independent of each other.

\item[\texttt{\#indepGroups...\#endIndepGroups}] declares that each number ($\geq 2$) of ground instances of each possible combination of different formulas between \verb§#indepGroups§ and \verb§#endIndepGroups§ are unconditionally independent. With $GF^1=\{f^1_i\}$, ..., $GF^n=\{f^n_j\}$, ... being the sets of ground instances of the formulas $F^1$ ... $F^n$ within \texttt{\#indepGroups...\#endIndepGroups}, this declaration imposes the following constraints:\\ 

$\forall sel \in 2^{\{1,...,n\}}, |sel| \geq 2, \mathit{sf} = \{ GF^i: i \in sel \}, sg =  \mathit{sf}_1 \times ... \times \mathit{sf}_{|\mathit{sf}|},$\\
$ \forall (sg_1,...,sg_{|\mathit{sf}|}) \in sg: Pr(\bigwedge_{k=1}^{|\mathit{sf}|} sg_k) = \prod_{k=1}^{|\mathit{sf}|}Pr(sg_k)$\\

Apart from that, formulas $F^1$ ... $F^n$ are treated as regular background knowledge. Normally, the enclosed formulas have weights attached.  \\
Each \verb§.prasp§ file can contain multiple \texttt{\#indepGroups...\#endIndepGroups} segments.\\

Otherwise, \texttt{\#indepGroups...\#endIndepGroups} behaves like \\
\texttt{\#indep...\#endIndep}.

\item[\texttt{\#pIndep...\#endPIndep}] declares that the set of all formulas between \verb§#pIndep§ and \verb§#endPIndep§ is \textit{pairwise} independent (in the probabilistic sense). This imposes the following constraint:\\

$\forall f_i, f_j: Pr(f_i \wedge f_j) = Pr(f_i)Pr(f_j)$ for all formulas $f_i$ and $f_j$ in \texttt{\#pIndep...\#endPIndep}. All $f_i$ need to be distinct from each other (\prasp does not check this).\\

Observe that each two formulas for which the above holds are also uncorrelated. However, the entire set of  formulas $f_i$ is not necessarily \textit{mutually independent}. If you require mutual independence instead of pairwise independence, use \texttt{\#indep...\#endIndep}.\\

Apart from that, \texttt{\#pIndep...\#endPIndep} behaves like \texttt{\#indep...\#endIndep}.

\item[\texttt{\#pIndepGroups...\#endPIndepGroups}] declares that each two ground instances of each possible \textit{pair} of different (non-ground or ground) formulas between \verb§#pIndepGroups§ and \verb§#endPIndepGroups§ are independent. With $GF^1=\{f^1_i\}$, ..., $GF^n=\{f^n_j\}$, ... being the sets of ground instances of the formulas $F^1$ ... $F^n$ within \texttt{\#pIndepGroups...\#endPIndepGroups}, this declaration imposes the following constraints:\\ 

$\forall 1 \leq i, j \leq n, i \neq j\ \ \ \forall f^i_m \in GF^i, f^j_n \in GF^j: Pr(f^i_m \wedge f^j_n) = Pr(f^i_m)Pr(f^j_n)$\\

Otherwise, \texttt{\#pIndepGroups...\#endPIndepGroups} behaves like \\
\texttt{\#indep...\#endIndep}.

If you require \textit{mutual independence} instead of pairwise independence, use\\ \texttt{\#indepGroup...\#endIndepGroup}.\\

Example:

\begin{verbatim}
face(1..6).		
dice(1..2).

#pIndepGroups

[[:]] result(1,F) :- face(F).
[[:]] result(2,F) :- face(F).

#endPIndepGroups

1{ result(1,F): face(F) }1. 
1{ result(2,F): face(F) }1. 
win :- result(1,6), result(2,6).
[0.8|win] happy.
:- happy, not win.

\end{verbatim}

The \verb§#pIndepGroups§ segment generates two groups of ground formulas:

Group 1:
\begin{verbatim}
result(1,6).
result(1,5).
result(1,4).
result(1,3).
result(1,2).
result(1,1).
\end{verbatim}
Group 2:
\begin{verbatim}
result(2,6).
result(2,5).
result(2,4).
result(2,3).
result(2,2).
result(2,1).
\end{verbatim}

For each pair $(x \in Group1, y \in Group2)$, \prasp adds the constraint $Pr(x \wedge y) = Pr(x)Pr(y)$. E.g.,
$Pr(result(1,6) \wedge result(2,5)) = Pr(result(1,6))Pr(result(2,5))$. In contrast, the formulas
\emph{within} each group are \underline{not} necessarily independent of each other.

\item[\texttt{\#indepVolat...\#endIndepVolat}]
\ \newline and \textbf{\texttt{\#pIndepVolat...\#endPIndepVolat}}\\
 have the same effect as \texttt{\#indep...\#endIndep} / \texttt{\#pIndep...\#endPIndep}, with the exception that the formulas within 
this section do not become part of the background knowledge (they are only used to generate probabilistic independence constraints). 

\item[\texttt{\#indepGroupsVolat...\#endIndepGroupsVolat}]
 \ \newline and \textbf{\texttt{\#pIndepGroupsVolat...\#endPIndepGroupsVolat}}\\
 have the same effect as  \texttt{\#indepGroups...\#endIndepGroups} /\\ \texttt{\#pIndepGroups...\#endPIndepGroups}, with the exception that the formulas within this segment do not become part of the background knowledge (they are only used to generate probabilistic independence constraints). 

\item[\texttt{\#gIndep...\#endGIndep}] declares that the set $FG$ of all formulas between \verb§#gIndep§ and \verb§#endGIndep§ are mutually independent (in the probabilistic sense). Additionally, these formulas are mutually independent from \textit{all} other formulas which are declared or discovered mutually independent. Furthermore, they are pairwise independent from all formulas which have been declared pairwise independent. In other words, formulas in \texttt{\#gIndep...\#endGIndep} behave like those formulas \prasp detects automatically (both sets might overlaps, in which case duplicates are ignored).

\end{description}
\end{description}

\newpage \section{Semantics \& Core Algorithms}
\label{corealgos}
%\begin{figure}
%\centering
%\includegraphics[width=1.4\linewidth, scale = 1, trim=5mm 5mm 0 10mm]{PrASP_Framework_Overview_ver2.pdf}
%\caption{\prasp inference core - overview} 
%\label{fig:inferenceCore}
%\end{figure}

%\subsection{Preprocessing \& Simplification Stage}

%Spanning formula generation, \hyperref[cmdline:mod0]{\texttt{--mod0}}, \hyperref[cmdline:mod1]{\texttt{--mod1}}, etc.\\
%\noindent {\color{Red} (tbw.) }

\subsection{\prasp Semantics \& Inference Based on Linear Programming (\textit{\prasp Vanilla})}
\label{semantics}

Without any command-line switches, \prasp solves a system of linear equations or inequalities (the latter in case there are any interval weights). This approach also defines the ``official'' semantics of \prasp inference. The actual solving approach depends on the nature of the (in)equation system (whether it is under-determined or not, whether there are probability intervals in the background knowledge) and is typically approximating in order to be tractable. In the following, we assume linear programming is being used.\\

Given a \prasp program $\kb$, the inference semantics is defined based on a probability distribution over a finite set of possible worlds in form of \textit{answer sets} of the \textit{spanning program}\cite{prasp15,prasp12,prasp13,prasp14} $\dmh(\kb)$. \\
Informally, $\dmh(\kb)$ is a non-probabilistic disjunctive program\footnote{\prasp's default ASP grounder/solver Clingo additionally allows for function symbols, but for simplicity we ignore this in this section.} generated from $\kb$ by removing all weights and transforming each formerly weighted formula $f$ into a spanning formula, as follows:
If $f$ is in FOL syntax, its spanning formula is the disjunction $f|\neg\ f$ (where $\neg$ stands for default negation). If $f$ is in ASP syntax (under stable model semantics), rules are interpreted as FOL rules and the spanning formula is also $f|\neg\ f$. If $f$ is an atom, the spanning formula can be expressed as a so-called \textit{choice rule} \cite{asp}, and is equivalent to \verb§f :- not not f§ (or \verb§f :- {not f}0§ in Gringo 3 syntax). Choice rules stemming from weighted atoms are the main (albeit not the only) vehicle for expressing uncertainty in \prasp. We assume that all predicates which occur in $\kb$ are defined (they occur in the head of at least one rule). Technical details about spanning formula generation can be found in Sect. \ref{spanGen}. After transforming each weighted formula into its spanning formula, the resulting FOL program under stable model semantics is transformed into ASP syntax according \cite{f2lp} which gives the spanning program from which later the possible worlds are generated in form of its answer sets. Note that $f|\neg\ f$ does not guarantee that answer sets are generated for weighted formula $f$. \\

We define $\as(a)$ as set of all answer sets of answer set program $a$, i.e., the set of possible worlds deemed possible according to existing belief $\dmh(\kb)$ is denoted as $\as(\dmh(\kb))$.\\

We define the parameterized probability distribution $\mu(\kb, \Theta)$ over a set $\Theta = \{ \theta_i \in \Theta \}$ of answer sets (possible worlds) as a randomly selected solution $\{ Pr(\theta_i): \theta_i \in \Theta \}$ of the following system of inequalities. If multiple (that is, an infinite number of) solutions exist, we compute a user specifiable number of solutions (command-line argument\verb§ --ndistrs n§) and/or select the one with (approximately) maximum entropy (see Sect. ref{intervalresults}). \\
 
We define the parameterized probability distribution $\mu^l(\kb, \Theta, q)$ over a set $\Theta = \{ \theta_i \in \Theta \}$ of answer sets (possible worlds), a \prasp program $\kb = \{ ([p_i] f_i)\} \cup \{ ([p_i|c_i] f^c_i)\} \cup \{ \mathit{indep}(\{ f_1^i,..., f_k^i \}) \}$ and query formula $q$ as the maximum entropy solution $\{ Pr(\theta_i): \theta_i \in \Theta \}$ of the following system of inequalities (\textit{constraints}) such that $Pr^l(q) = \sum_{\theta_i \in \Theta: \theta_i \vDash_{\kb} q}{Pr(\theta_i)}$ is minimized (analogously, $\mu^u$ denotes the maximizing distribution). The result for query \verb§[?] q§ is defined as the interval $[Pr^l(q),Pr^u(q)]$ (analogously for conditional queries\verb§[?|c] f§, where we compute $Pr(f|c)$ as $Pr(f \wedge c) / Pr(c)$).\\
For small systems, \prasp can compute minimizing and maximizing distributions using linear programming or SMT solving\footnote{If there are no intervals, the system can be approximately solved using a Non-Negative Linear Least Squares method, see \hyperref[cmdline:linsolveconf]{\texttt{--linsolveconf}}.} and a maximum entropy solution amongst a number of candidate distributions (solutions of an under-determined system) can be discovered using gradient descent. However, to make distribution finding scalable, we need to use other algorithms, see next section. 

{ \begin{gather}
	\label{linearSystem}
	l(f_1) \leq \sum_{\theta_i \in \Theta: \theta_i \vDash_{\kb} f_1}{Pr(\theta_i)}   \leq u(f_1) 
	\end{gather}
	\vspace{-0.2cm}
	\centerline{\textbf{$ \cdots$}}
	\vspace{-0.4cm}
	\begin{gather}
	l(f_n) \leq \sum_{\theta_i \in \Theta: \theta_i \vDash_{\kb} f_n}{Pr(\theta_i)}   \leq u(f_n)\\		
	%	\prod_{f_i^1}{l(f_i^1)} \leq \sum_{\theta_i \in \Theta: \theta_i \vDash_{\kb} \bigwedge f_i^g}{Pr(\theta_i)}   \leq \prod_{f_i^1}{u(f_i^g)}
	%	\end{gather}
	%	\vspace{-0.2cm}
	%	\centerline{\textbf{$ \cdots$}}
	%	\vspace{-0.4cm}
	%	\begin{gather}
	%	\prod_{f_i^g}{l(f_i^g)} \leq \sum_{\theta_i \in \Theta: \theta_i \vDash_{\kb} \bigwedge f_i^g}{Pr(\theta_i)}   \leq \prod_{f_i^g}{u(f_i^g)}	\\	
	% \sum_{\theta_i \in \Theta}{Pr(\theta_i)\nu(\theta_i,f^c_1 \wedge c_1)} +  
	%\sum_{\theta_i \in \Theta}{-l(f^c_1|c_1)Pr(\theta_i)\nu(\theta_1,c_1)} > 0\\
	%		 \sum_{\theta_i \in \Theta}{Pr(\theta_i)\nu(\theta_i,f^c_1 \wedge c_1)} +  
	%		 \sum_{\theta_i \in \Theta}{-u(f^c_1|c_1)Pr(\theta_i)\nu(\theta_1,c_1)} < 0
	%	\end{gather}
	%				\vspace{-0.2cm}
	%				\textbf{\centerline{$ \cdots$}}
	%				\vspace{-0.4cm}
	%	\begin{gather}
	%	 \sum_{\theta_i \in \Theta}{Pr(\theta_i)\nu(\theta_i,f^c_m \wedge c_m)} +  
	%	 \sum_{\theta_i \in \Theta}{-l(f^c_m|c_m)Pr(\theta_i)\nu(\theta_1,c_m)} > 0\\
	%	 \sum_{\theta_i \in \Theta}{Pr(\theta_i)\nu(\theta_i,f^c_m \wedge c_m)} +  
	%	 \sum_{\theta_i \in \Theta}{-u(f^c_m|c_m)Pr(\theta_i)\nu(\theta_i,c_m)} < 0 \\
	\sum_{\theta_i \in \Theta}{\theta_i = 1}\\
	\forall \theta_i \in \Theta: 0 \leq Pr(\theta_i) \leq 1 
	\end{gather}}

\noindent At this, $l(f_i)$ and $u(f_i)$ denote the endpoints of the probability interval (weight) of unconditional formula $f_i$ (analogous for endpoints $l(f^c_i|c_i)$ and $u(f^c_i|c_i)$ of conditional probabilities). \\
It is noteworthy that the $f_i$ are \textit{not} necessarily ground in our approach.\\

\noindent We define

$\nu(\theta,f) = \begin{cases}
1, & \text{if}\ \theta \vDash_{\kb} f \\
0, & \text{otherwise}
\end{cases}$\\
In addition to the constraints above, $indep$-declaration of the form $\mathit{indep}(F)$ for non-conditional mutual independence in the program add further constraints of the following form to the linear system for each subset $\{ f_1^i,..., f_r^i \} \subseteq F$, $r > 1$:\\
$	\prod_{f_{k=\{1..r\}}^i}{l(f_k^i)} \leq \sum_{\theta_i \in \Theta: \theta_i \vDash_{\kb} \bigwedge f_{k=\{1..r\}}^i}{Pr(\theta_i)} \leq \prod_{f_{k=\{1..r\}}^i}{u(f_k^i)}$\\
and a conditional probability formula $[p_i|c_i] f^c_i)$ in the program induces constraints\\
$ \sum_{\theta_i \in \Theta}{Pr(\theta_i)\nu(\theta_i,f^c_i \wedge c_1)} +  
\sum_{\theta_i \in \Theta}{-l(f^c_i|c_i)Pr(\theta_i)\nu(\theta_1,c_i)} > 0\\
\sum_{\theta_i \in \Theta}{Pr(\theta_i)\nu(\theta_i,f^c_i \wedge c_i)} +  
\sum_{\theta_i \in \Theta}{-u(f^c_i|c_i)Pr(\theta_i)\nu(\theta_1,c_i)} < 0$\\

Note that the linear system of constraints above can often not be used directly for inference (except for very small systems), but it serves as a means in order to define the formal semantics of \prasp programs and queries.\\

\noindent We compute the probability $Pr(\phi)$ for a query formula $\phi$ using
{\small \begin{align}
	Pr(\phi) = 
	\sum_{\{\theta' \in \Theta': \theta' \models_{\kb} \phi\}} {Pr(\theta')} 
	\end{align}}

\noindent Like with all inference approaches built into \prasp, using the above system, we can compute a probability distribution over all answer sets of the spanning program or just a sampled subset of its models (\textit{initial sampling}), see command-line options \ref{commandline}.\\

\noindent The way \prasp performs Answer Set Checking ($\models$) is specified using command-line option \hyperref[cmdline:ascheckmode]{\texttt{--ascheckmode}}. Please see \ref{commandline} for details.

\subsection{Inference Using Weight Conversion and Model Counting}

With this approach \cite{prasp11}, weighted ASP formulas are converted into unweighted formulas with additional ``helper atoms''. The resulting ordinary ASP program $\dmh^{wc}(\kb)$ takes the place of the spanning program. Afterwards, (marginal) query probabilities are computed by determining the percentage of those answer sets in which the respective query formula holds (ignoring any helper atoms introduced in the conversion step). There is no need for computing an explicit probability distribution over possible worlds with this approach. 
However, computing models of $\dmh^{wc}(\kb)$ is very costly and inference results are only valid if the uncertain formulas in $\kb$ are mutually independent (which is not checked by \prasp}). Consider the approaches 
described under Section \ref{merecountingwithweights} which combine \hyperref[cmdline:nosolve]{\texttt{--nosolve }} with \hyperref[cmdline:initsample]{\texttt{--initsample}} as likely faster alternatives.\\

In $\dmh^{wc}(\kb)$, each formula $[w]\ f$ in $\kb$ is replaced with formulas $1\{ h_1, ..., h_n\}1$, $f \leftarrow h_1|...|h_m$ and $not\ f \leftarrow not\ (h_1|...|h_m)$ (that is, their ASP equivalents). At this, the $h_i$ are new names (``helper atoms''), $\frac{m}{n} = w$ and $m < n$.\\

This inference method is selected using \hyperref[cmdline:weights2cc --nosolve ]{\texttt{--weights2cc --nosolve }}on the commandline.
 
\subsection{Inference Using Simulated Annealing}
\label{simanneal}

Algorithm \ref{alg:simanneal} shows the approach \prasp uses for approximate inference using a parallel form of simulated annealing. For the sampling steps at each iteration ($\Call{sampleStep}{}$), this algorithm uses near-uniform sampling (e.g., using XOR-constraints based on the approach presented in \cite{xor}, see function $\Call{xorSample}{}$) or one of the other sampling approaches available with command-line option \hyperref[cmdline:simanneal]{\texttt{--simanneal}}. The initial list of samples $initSamples$ is computed according to the approach specified with command-line option \hyperref[cmdline:initsample]{\texttt{--initsample}}.

\def\multiset#1#2{\left(\!\left({#1\atopwithdelims..#2}\right)\!\right)}

\newcommand{\argmin}{\operatornamewithlimits{argmin}}

\begin{algorithm}
	\caption{Inference by parallel simulated annealing Part 1.\newline(We show only the basic variant for non-conditional formulas with point weights. The extension for conditional probabilities and intervals is straightforward.)
		\label{alg:simanneal}}
	\begin{algorithmic}[1]
		\Require{$maxTime$, $maxEnergy$, $initTemp$, $initSamples$ (see \hyperref[cmdline:initsample]{\texttt{--initsample}}). $initSamples$ is a multiset which encodes a probability distribution via frequencies of models), $\mathit{frozen}$, $\mathit{degreeOfParallelization}$, $\alpha$, $F$ (set of weighted formulas), $\mathit{samplingMethod}$, $\kb$ (\prasp program)}
		\Statex
		\Let{$s$}{$initSamples$}  
		\Let{$k$}{0} 
		\Let{$temp$}{$initTemp$}
		\Let{$e$}{$\Call{energy}{s}$}
		
		\While{$k \le maxTime \wedge temp \ge \mathit{frozen}$ } 
		
			%\Let{$s''$}{$s$} 
		
			 \ParFor{$i\gets 1, \mathit{degreeOfParallelization}$}
				 \Let{$s''_i$}{$s'' \uplus \Call{sampleStep}{\mathit{samplingMethod}}$}
			
			\EndParFor			
							
			\Let{$s'$}{$\argmin_{s''}(\Call{energy}{s''_1},...,\Call{energy}{s''_n})$}
						
			\Let{$e'$}{$\Call{energy}{s'}$} 

			\If{$e' < e \vee random^1_0 < e^{-(e' - e) / temp)}$} 
			
				\Let{$s$}{$s'$}
				 
				\Let{$e$}{$e'$}
			
			\EndIf
			
			\Let{$temp$}{$temp\cdot \alpha$}
			          
			\Let{$k$}{$k+1$}

		\EndWhile
		
		\Statex
		
		\Ensure Multiset $s = (pw, \mathit{frq}) = \mu_{approx}(\kb)$ approximates the probability distribution $\mu(\kb) = Pr(\as(\dmh(\kb)))$  over the set $pw = \{ pw_i \} = \as(\dmh(\kb))$ of possible worlds by $\{ Pr(pw_i) \approx \frac{\mathit{frq}(pw)}{|s|} \}$. $\mu(\kb)$ is as defined in Sect. \ref{semantics}.\\
		Computation of query probabilities using $\mu_{approx}(\kb)$ is also done as described in Sect. \ref{semantics}.
				
			\Statex
		
		\Function{energy}{$s$} 
		\Comment (without entropy maximization criterion, tbw.)
		
			\ParFor{$f_i \in |F|$}
			\Let{$\mathit{freq}_{f_i}$}{{\Large $\frac{|\{\{ s' \in s: s' \models_{\kb} f_i \}\}|}{|s|}$}}
			\EndParFor
			
	\hspace{-0.25cm} \Return $\sqrt{\sum_{f_i \in F}{(\mathit{freq}_{f_i} - \mathit{weight}_{f_i})^2}}$ 
		
		\EndFunction
		
\State \Comment (Continued in Part 2 below)
\algstore{simanneal}
	
		\end{algorithmic}
	\end{algorithm}
	
\begin{algorithm}
	\addtocounter{algorithm}{-1}
	\caption{Inference by simulated annealing Part 2. Handling of conditional formulas omitted (straightforward)}
	\begin{algorithmic}[1]
		
		\algrestore{simanneal}

		\Function{stepSample}{$\mathit{samplingMethod}$} 
	
			\If{$\mathit{samplingMethod}$ = 0}  \Comment{see \texttt{--simanneal}}
% uniform 

				\Return $\Call{xorSample}{\dmh(\kb)}$  \Comment{$\dmh(\kb)$ is the spanning program of background knowledge (\prasp program) $\kb$}
	
			\ElsIf{$\mathit{samplingMethod}$ = 1}
		
		% rejection with use of sets of independent formulas
			
				\Let{$cs$}{$\Call{xorSample}{\dmh(\kb)}$}  \Comment{Alternative uniform sampling methods can be specified using \texttt{--unisample}}
				
				\Let{$FS_{supported}$}\\
				\hspace{1.5cm}  { $\{ F' \subseteq F: independent(F') \wedge \forall f' \in F': cs \models_{\kb} f' \}$ }
				
				%\Let{$prods$}{$\{ \prod_{f \in F_{supported}}{weight_{f}}: F_{supported} \in FS_{supported} \}$} 				

				\Let{$F_{supported}$}{$\argmin_{F \in FS_{supported}}{\prod_{f \in F}{weight_{f}}}$}
				
				\Let{$prob_{min}$}{$\prod_{f \in F_{supported}}{weight_{f}}$}
				
				\If{$|cs|=|F_{supported}|$}
				\Let{$r$}{$prob_{min}$}	
				\Else
				\Let{$r$}{$random^1_0 \cdot prob_{min}$}
				\EndIf
				
				\If{$random^1_0 \leq r$}
				\Return{$cs$}
				\Else
				\Return{$\{\}$}
				\EndIf
			
			\ElsIf{$\mathit{samplingMethod}$ = 2}
			
				% as samplingMethod 1, but now _all_ weighted formulas are assumed to be mutually independent
			
				\Let{$cs$}{$\Call{xorSample}{\dmh(\kb)}$} \Comment{Alternative uniform sampling methods can be specified using \texttt{--unisample}}
				
				\Let{$F_{supported}$}{ $\{ f \in F: cs \models_{\kb} f \}$ }
				
				\Let{$prod$}{$\prod_{f \in F_{supported}}{weight_{f}}$} 
	
				\If{$|cs|=|F_{supported}|$}
					 \Let{$r$}{$prod$}	
				\Else
					 \Let{$r$}{$random^1_0 \cdot prod$}
				\EndIf
				
				\If{$random^1_0 \leq r$}
					\Return{$cs$}
				\Else
					\Return{$\{\}$}
				\EndIf
			
			\ElsIf{$\mathit{samplingMethod} \geq 3$} \Comment{if $\geq 3$, the $\mathit{samplingMethod}$ numbers have the same meaning as the sampling method numbers used with \texttt{--initsample}. E.g., $\mathit{samplingMethod}$ = 4 means stratified weighted flip-sampling, see Algorithm \ref{alg:flipsamp}	}		
				\State {\LARGE ...}				
				% flipsamp				
%				\Let{$F'$}{$\{\}$}
%				
%				 \For{$f_i \in |F|$}
%				 
%					 	\If{$random^1_0 < weight_{f_i}$}
%						 	\Let{$F'$}{$F' \cup \{ f_i \}$}
%						\Else
%							\Let{$F'$}{$F' \cup \{ \neg f_i \}$}
%						\EndIf
%						 					
%				\EndFor
%				
%				%\Let{$F''$}{$\{ f \in F: weight_f = 1\}$}
%				
%				\Return $\Call{xorSample}{F'}$
%			
			\EndIf

		\EndFunction
		
		%\State \Comment (Continued in Part 3 below)
		%\algstore{simanneal}
		
			\end{algorithmic}
		\end{algorithm}
	
\subsection{Flip-Sampling}
\label{flipsamp} 

	{\scriptsize 	\begin{algorithm}
			\addtocounter{algorithm}{0}
			\caption{Weighted stratified flip-sampling for sampling from all answer sets of the spanning program (\textbf{simplified version} - see Algorithm \ref{alg:flipsamp} for details). 
			%	It is related to \textit{distribution semantics} \cite{prism0,icl} where the semantics of a probabilistic program is defined in terms of a probability distribution over a set of \textit{normal logic programs}.
			}
			\label{alg:flipsampsimple}	
			\begin{algorithmic}[1]	
				
				\Require{maximum number of samples $n$, \prasp program $\kb$, set of uncertain formulas $\mathit{uf} = \{\mathit{uf}_i: [w(\mathit{uf}_i)] \mathit{uf}_i \in \kb \wedge w(\mathit{uf}_i) < 1\}$, set of certain formulas $\mathit{uf} = \{\mathit{uf}_i: [w(\mathit{uf}_i)] \mathit{uf}_i \in \kb \wedge w(\mathit{uf}_i) = 1\}$}
				\Let{$i$}{$1$} 
				\While{$i \leq |\mathit{uf}|$}
				\Let{$r^i$}{random element of $\mathit{Sym}(\{1,...,n\})$ (permutations of $(1,...,n)$)}
				\Let{$i$}{$i+1$}
				\EndWhile
				\Let{$m$}{$\{\}$}
				%\Let{$j$}{1}
				\ParFor{$j \in \{1,...,n\}$}
				\Let{$p$}{$\{\}$}
				\Let{$k$}{1}
				\While{$k \leq |\mathit{uf}|$}
				\If{$r^k_j \leq n\cdot w(\mathit{uf}_k)$} %due to "stratification", this is not a Bernoulli trial
				\Let{$p$}{$p \cup \mathit{uf}_k$}
				\Else
				\Let{$p$}{$p \cup \neg \mathit{uf}_k$}
				\EndIf
				\Let{$k$}{$k+1$}
				\EndWhile
				\Let{$s$}{model sampled uniformly from models of program $\mathit{cf} \cup p$} (if models exist, otherwise $\emptyset$)
				\State \Comment This step can be influenced using \hyperref[cmdline:unisample]{\texttt{--unisample}}
				\Let{$m$}{$m \uplus\{s\}$}
				\EndParFor
				\Ensure Multiset $m$ contains samples from all answer sets of spanning program such that\\ $\forall \mathit{uf}_i: w(\mathit{uf}_i) \approx$ {\large $\frac{|\{ s \in m: s \models\mathit{uf}_i\}|}{|m|}$} \textit{iff} set $\mathit{uf}$ mutually independent, i.e., subset of some $F$, $\mathit{indep}(F)$.
			\end{algorithmic}
		\end{algorithm}
	}
	
	{\scriptsize 	\begin{algorithm}
			\addtocounter{algorithm}{0}
			\caption{Weighted flip-sampling for sampling from all answer sets of the spanning program (for non-conditional probabilities - version for conditional probabilities tbw.). % Flip-sampling is a form of rejection sampling where attached weights define the prior distribution.
			}
			\label{alg:flipsamp}
			\begin{algorithmic}[1]	 			
				\Require{maximum number of samples $n$, \prasp program $\kb$, set of uncertain formulas $\mathit{uf} = \{\mathit{uf}_i: [w^l(\mathit{uf}_i);w^u(\mathit{uf}_i)]\ \mathit{uf}_i \in \kb \wedge w(\mathit{uf}_i) < 1\}$, set of certain formulas $\mathit{uf} = \{\mathit{uf}_i: [w(\mathit{uf}_i)] \mathit{uf}_i \in \kb \wedge w(\mathit{uf}_i) = 1\}$, declared or automatically discovered sets $F$ of independent formulas available via function $\mathit{independent}(F)$, $\mathit{ignoreindependence}$, $\mathit{stratify}$}
				\Let{$i$}{$1$} 
				\While{$i \leq |\mathit{uf}|$}
				\If{$\mathit{stratify}$}
				\Let{$r^i$}{random element of $\mathit{Sym}(\{\frac{1}{n},...,\frac{n}{n}\})$ (permutations of $(\frac{1}{n},...,\frac{n}{n})$)}
				\Else
				\Let{$r^i$}{$(rand_0^1, ..., rand_0^1) \in \mathbb{R}^n$}
				\EndIf
				\Let{$i$}{$i+1$}
				\EndWhile
				\Let{$m$}{$\{\}$}
				%\Let{$j$}{1}
				\ParFor{$j \in \{1,...,n\}$}
				\Let{$p$}{$\{\}$}
				\Let{$k$}{1}
				\While{$k \leq |\mathit{uf}|$}
				\If{$r^k_j \leq w^l(\mathit{uf}_k) + (w^u(\mathit{uf}_k) - w^l(\mathit{uf}_k)) \cdot rand_0^1 \wedge (\mathit{independent}(p \cup \mathit{uf}_k) \vee \mathit{ignoreindependence})$}
				\Let{$p$}{$p \cup \mathit{uf}_k$}  				\Comment{as for the remote version, see \textbf{(*)}}
				\ElsIf{$\mathit{independent}(p \cup \neg \mathit{uf}_k) \vee \mathit{ignoreindependence}$} 
				\Let{$p$}{$p \cup \neg \mathit{uf}_k$}
				\EndIf
				\Let{$k$}{$k+1$}
				\EndWhile
				\Let{$s$}{model sampled uniformly from models of program $\mathit{cf} \cup p$} (if models exist, otherwise $\emptyset$)
				\Statex \Comment This step can be influenced using \hyperref[cmdline:unisample]{\texttt{--unisample}}
				\Let{$m$}{$m \uplus\{s\}$}
				\EndParFor
				\Ensure Multiset $m$ contains samples from all answer sets of spanning program such that\\ $\forall \mathit{uf}_i: w^{avg}(\mathit{uf}_i) \approx$ {\large $\frac{|\{ s \in m: s \models\mathit{uf}_i\}|}{|m|}$} \textit{iff} set $\mathit{uf}$ mutually independent, i.e., subset of some $F$, $\mathit{indep}(F)$.
				\Statex
				\Statex \Comment \textbf{(*)} In the remote version (script \verb§clingo4Sampling.lp§), extra statements of the form \verb§#external hp__atom_switch_uf_k. 1{atom_uf_k; not hp__atom_switch_uf_k}1.§ are added to the spanning program for each weighted atom \verb§atom_uf_k§. Instead of assembling $p$ using $p \leftarrow p \cup \mathit{uf}_k$ and $p \leftarrow p \cup \neg \mathit{uf}_k$, we take the entire spanning program and switch on/off each atom $\mathit{uf}_k$ by assigning truth values to the respective external atoms. Please see \verb§clingo4Sampling.lp§ for details.
			\end{algorithmic}
		\end{algorithm}
	}
	
See Algorithms \ref{alg:flipsamp} and \ref{alg:flipsampsimple}. The random choice made by flip-sampling over the two ``branches'' of each spanning formula resembles choice in distribution semantics-based approaches \cite{sato} such as ProbLog, however, besides the fact that flip-sampling is a sampling approach, selection in flip-sampling is not just over probabilistic facts.\\

There is also \textit{uniform flip-sampling}, which is implemented as a special case of weighted flip-sampling where the weight of each uncertain formula is assumed to be $0.5$.
	
\subsection{Near-uniform sampling using XOR-Constraints}

Near-uniform sampling is a foundational approach in \prasp. It is used in various places, e.g., as an option for initial sampling. There are various ways to do this, as specified by switch \hyperref[cmdline:unisample]{\texttt{--unisample}} and further configurable using \hyperref[cmdline:unisample]{\texttt{--unisample}}, \hyperref[cmdline:xorconf]{\texttt{--xorconf}}, \hyperref[cmdline:sirndconf]{\texttt{--sirndconf}} and \hyperref[cmdline:flipsampconf]{\texttt{--flipsampconf}}. The following algorithm (Algo. \ref{alg:xor}) is the most interesting approach to this (although it is not the default approach anymore since \prasp 0.8), namely near-uniform sampling using so-called \textit{XOR streamlining constraints} (XOR-sampling, switch \hyperref[cmdline:xorconf]{\texttt{--xorconf}}), an approach introduced in \cite{xor}.   

	\begin{algorithm}[H]				
			\caption{Near-Uniform Sampling using XOR-Constraints \cite{xor}}	
			\label{alg:xor}
			\addtocounter{algorithm}{0}
		\begin{algorithmic}[1]
	\Function{xorSample}{$\dmh(\kb)$} 
	\Statex
	\Comment The following procedure \cite{xor} computes a near-uniformly drawn random sample $\gamma$ from the set $\as(\dmh(\kb))$ of all answer sets of ASP program $\dmh(\kb)$ .\\
	\Comment $q_1$ and $q_2$ are used-defined parameters which influence the quality of the result.
	\Statex
	\State $\psi_g \leftarrow$ ground($\dmh(\kb)$)
	\State $ga \leftarrow atoms(\psi_g) \cup \mathtt{true}$
	\Let{$V$}{$\{\  \mathtt{:\hyphen\ \#even\{ a_1, ..., a_n \}}: a_i \in ga, 1 \le n \le |ga| \}$}  \Comment XOR constraints
	\State $\mathit{xors_{sampled}} \leftarrow \{ xor_1, ..., xor_{q_2} \} \subseteq V$, such that each $xor_i$ is drawn independently at random with probability 0.5 and includes $\mathtt{true}$ with probability 0.5. \Comment \prasp uses $q_2 = log_2(|S|)$ by default (where $|S|$ is the number of symbols in $\kb$).
	\State $\gamma \leftarrow $ any randomly selected answer set from $M \subseteq \as(\dmh(\kb) \cup \mathit{xors_{sampled}})$ with $|M| = q_1$  \Comment $q_1$ needs to be $|\as(\dmh(\kb) \cup \mathit{xors_{sampled}})|$ for near-uniform sampling\\
	\Return $\gamma$
	\EndFunction
	\end{algorithmic}
\end{algorithm}

%See \cite{xor} for the rationale behind this algorithm. 

\subsection{Iterative Refinement}
\label{itrefinement}

This inference algorithm (Algorithm \ref{alg:itrefine}) - which we call Iterative Refinement\footnote{Not related to the Iterative Refinement method in solving systems of linear equations.} - is based on the entropy maximizing inference approach introduced in \cite{spirit} which makes use of the Kuhn-Tucker theorem. For usage instructions, see switch \hyperref[cmdline:itrefinement]{\texttt{--itrefinement}}. The algorithm either starts, as in \cite{spirit}, from the uniform distribution over the possible worlds delivered by the initial sampling step (with duplicates removed), or from the (possibly non-uniform) probability distribution obtained by the initial sampling step with duplicates retained. In the former case, the Kullback$\textrm{-}$Leibler divergence to the uniform distribution is minimal (and thus the final distribution's entropy is maximal). However, this is relative to the number of models sampled by initial sampling - if initial sampling provides only a subset of all models of the spanning program (a concept which doesn't appear in \cite{spirit}), entropy will also be lower compared to using the full set of possible worlds as input for iterative refinement as in \cite{spirit}. In the latter case (retaining of duplicates), the counts of worlds within the multiset of possible worlds specifies the possible world probabilities which are then iteratively refined.  

\begin{algorithm}[H]
	\caption{Maximum entropy inference by iterative refinement (command-line switch \texttt{--itrefinement}). The algorithm below (which is based on the algorithm presented in \cite{spirit}) uses $\mathit{initSamples}$ as a multiset; the original form of this algorithm without multiple instances of the same possible world is obtained by using $\mathit{initSamples}$ as a set.
		\label{alg:itrefine}}
	\begin{algorithmic}[1]
		\Require{$\mathit{maxIterations}$, $\epsilon$, $\mathit{initSamples}$ (see \hyperref[cmdline:initsample]{\texttt{--initsample}}), $\kb$ (a \prasp program), set of uncertain conditional probability formulas with (point) weights $\mathit{ufw} = \{(\mathit{uf}^f_i, \mathit{uf}^c_i, w_i): [w_i|\mathit{uf}^c_i] \mathit{uf}^f_i \in \kb \wedge w_i < 1\}$} (the non-conditional case can simply be obtained with $\mathit{uf}^c_i = true$), convergence threshold $\epsilon$
		\Statex
		\State $Pr^0(pw_i) = \frac{\mathit{frq}(pw_i)}{|\mathit{initSamples}|} \Leftrightarrow \mathit{initSamples} = (pw, \mathit{frq}) $    \Comment $\mathit{frq}(pw_i)$ is the count of answer set (possible world) $pw_i$ within multiset $\mathit{initSamples}$
	%	\Let{$pd^k$}{$\{ Pr(pw_i): pw_i \in \mathit{initSamples}\}$}			
		\Let{$k$}{1} 
		\Repeat
				\For{$(\mathit{uf}^f_i, \mathit{uf}^c_i, w_i) \in \textit{uf}$}
					\For{$pw_i \in \mathit{initSamples}$}
					\Let{$Pr^{k-1}(\phi)$}{$\sum_{\{pw_i \in \mathit{initSamples}: pw_i \models_{\kb} \phi\}} {Pr^{k-1}(pw_i)}$ 
					\State \ \ \ \ \ for all $\phi \in \{ \mathit{uf}^c_i \wedge \mathit{uf}^f_i, \mathit{uf}^c_i \wedge \neg \mathit{uf}^f_i, \neg \mathit{uf}^c_i\}$}		
				\Let{$b$}{$Pr^{k-1}(\mathit{uf}^c_i \wedge \mathit{uf}^f_i)^{w_i} Pr^{k-1}(\mathit{uf}^c_i \wedge \neg \mathit{uf}^f_i)^{1-w_i}$}			
					\Let{$a$}{{\Large $\frac{b}{b + Pr^{k-1}(\neg \mathit{uf}^c_i){w_i^{w_i}} (1-w_i)^{1-w_i}}$}}
					\State $Pr^{k}(pw_i) = \begin{cases}
						Pr^{k-1}(pw_i){\Large \frac{1-a}{Pr^{k-1}(\neg \mathit{uf}^c_i)}}  & \text{if }pw_i \models_{\kb} \neg \mathit{uf}^c_i\\
						Pr^{k-1}(pw_i)\frac{(1-w_i)a}{{\Large Pr^{k-1}(\mathit{uf}^c_i \wedge \neg \mathit{uf}^f_i)}}  & \text{if }pw_i \models_{\kb} \mathit{uf}^c_i \wedge \neg \mathit{uf}^f_i\\
						Pr^{k-1}(pw_i)\frac{w_i a}{{\Large Pr^{k-1}(\mathit{uf}^c_i \wedge \mathit{uf}^f_i)}}  & \text{if }pw_i \models_{\kb} \mathit{uf}^c_i \wedge \mathit{uf}^f_i\\
					\end{cases}	$				
					\EndFor
				\EndFor
				\Let{$d$}{$\sqrt{\sum_{pw_i \in \mathit{initSamples}}{(Pr^k(pw_i) - Pr^{k-1}(pw_i))^2}}$}
				
	    \Let{$k$}{$k+1$} 
		\Until{$k > \mathit{maxIterations} \vee d \leq \epsilon$}
		
		\Statex
		
		\Ensure $\{ Pr^k(pw_i) \} = \mu_{approx}(\kb)$ approximates the probability distribution $\mu(\kb) = Pr(\as(\dmh(\kb)))$  over the set $\as(\dmh(\kb))$ of possible worlds. $\mu(\kb)$ is as defined in Sect. \ref{semantics}. Importantly, $Pr^k, Pr^{k+1}, ...$ converges to the probability distribution with maximum entropy among all distributions over $\{pw_i\}$ where the constraints imposed by the formula weights hold, \textit{provided} $Pr^0$ is the uniform distribution \cite{spirit}.\\
		Computation of query probabilities using $\mu_{approx}(\kb)$ is done as described in Sect. \ref{semantics}.		
	\end{algorithmic}
\end{algorithm}

\subsection{PrASP's MaxWalkSAT variant}
\begin{algorithm}[H]
\prasp with switch \hyperref[cmdline:maxwalksat]{\texttt{--maxwalksat}} uses a slightly modified version of the algorithm presented in \cite{kautz}. All formulas need to be ``simple formulas'' in the sense of Sect. \ref{simpleFormulas}. $w(f)$ denotes the weight of formula $f$. However, if the weight is an interval, $w(f)$ gives the mean value of the interval boundaries. $\textit{randomUNSATFormula}(m)$ is a formula chosen randomly from all formulas of which $m$ is not a model. If the weight (or its lower bound) is $\geq 1$, $w(f)$ denotes 1000.\\
As for lazy variants of this algorithm (e.g., LazySAT \cite{mln2}), see considerations under \hyperref[cmdline:maxwalksat]{\texttt{--maxwalksat}}.
\label{alg:maxwalksat}
\begin{algorithmic}
\Let{$atoms$}{$allAtomsIn(formulas)$} 	
\For{$i$ $\leftarrow$ $1$ to $maxTries$}
\Let{$m$}{$randomSubset(atoms)$}
\For{$j$ $\leftarrow$ $1$ to $maxFlips$}
\Let{$cost$}{\Call{cost}{$m$}}
\If{$cost \leq acceptableCost$}
\Let{$topModels$}{$topModels \cup m$}
\If{$|topModels| \geq n$} \Return{$topModels$} \Else\ \textbf{break} \EndIf
\EndIf
\Let{$usf$}{$\textit{randomUNSATFormula}(m)$}
\If{$\textit{random}_0^1 < p$}
\Let{$flipAtom$}{$\textit{randomAtomFrom}(allAtomsIn(usf))$}
\Else
\For{$a \in allAtomsIn(usf)$}
\Let{$fm$}{\Call{flip}{$m$, $a$}}
\Let{$newCost(a)$}{\Call{cost}{$fm$}}
\EndFor
\Let{$flipAtom$}{Atom with lowest $newCost(a)$}
\EndIf
\Let{$m$}{\Call{flip}{$m$, $flipAtom$}}
\EndFor % j
\EndFor % i
\State
\Return{Max number of tries exceeded, return $topModels$}
\State
\Function{flip}{$m, a$} 
\If{$a \in m$}
\Return{$m \setminus \{a\}$}
\Else\ \Return{$m \cup \{a\}$}
\EndIf
\EndFunction
\Function{cost}{$m$} 
\Return{$\sum_{f_i \in \{ f: f \in formulas, m \nvDash f\}}w(f_i)$}
\EndFunction
\end{algorithmic}
\end{algorithm}

%\subsection{Parameter Learning Using Stochastic Gradient Ascent}
%
%tbw.
%
\subsection{Parameter Learning Using Barzilai-Borwein Algorithm}

Let $H = \{f_1, ..., f_n\}$ be a given set of formulas and a vector $w = (w^1,...,w^n)$ of (unknown) weights of these formulas. Using the Barzilai and Borwein method \cite{Barzilai} (a variant of the gradient descent approach with possibly superlinear convergence), we seek to find $w$ such that $Pr(E | H_w \cup B)$ is maximized ($H_w$ denotes the formulas in $H$ with the weights $w$ such that each $f_i$ is weighted with $w^i$). % in dependence of vector $w$.
Any existing weights of formulas in the background knowledge ar not touched, which can significantly reduce learning complexity if $H$ is comparatively small. Probabilistic or unobservable examples are not considered.

\noindent The learning algorithm \cite{Barzilai} is as follows:\\

\noindent Repeat for $k=0,1,...$ until convergence:\\
\indent Set  $s_k = $ {\Large$\frac{1}{\alpha_k}$} $\triangledown(Pr(E|H_{w_k} \cup B))$\\
\indent Set $w_{k+1} = w_{k} + s_k$\\
\indent Set $y_{k} = \triangledown(Pr(E|H_{w_{k+1}} \cup B)) - \triangledown(Pr(E|H_{w_k} \cup B))$\\

\indent Set $\alpha_{k+1} = $ {\Large$\frac{s^T_k y_k}{s^T_k s_k}$}\\

At this, the initial gradient ascent step size $\alpha_0$ and the initial weight vector $w_0$ can be chosen freely. 
%let $w^{s+1}$ is the next estimation of $w$ using current estimation $w^{s}$, and $\alpha$ is the learning rate. 
$Pr(E|H_w \cup B)$ denotes $\prod_{e_i \in E}Pr(e_i | H_w \cup B)$ inferred using vector $w$ as weights for the hypothesis formulas, and
{\small \begin{gather}
	\bigtriangledown(Pr(E|H_w \cup B)) =\\
	(\frac{\partial}{\partial w^1} Pr(E|H_w \cup B), ..., \frac{\partial}{\partial w^n} Pr(E|H_w \cup B))
	\end{gather}}

Since we usually cannot practically express $Pr(E|H_w \cup B)$ in dependency of $w$ in closed form, at a first glance, the above formalization appears to be not very helpful. However, we can still resort to numerical differentiation and approximate
{\small \begin{gather}
	\bigtriangledown(Pr(E|H_w \cup B)) =\\
	(\lim_{h \rightarrow 0} \frac{Pr(E|H_{(w^1+h, ..., w^n)} \cup B)-Pr(E|H_{(w^1, ..., w^n)} \cup B)}{h},
	\end{gather}
	\centerline{{\large ...,}}
	\begin{gather}
	\lim_{h \rightarrow 0} \frac{Pr(E|H_{(w^1, ..., w^n+h)} \cup B)-Pr(E|H_{(w^1, ..., w^n)} \cup B)}{h})
	\end{gather}}
by computing the above vector (dropping the limit operator) for a sufficiently small $h$ (in our prototypical implementation, $h = \sqrt{\epsilon}w_i$ is used, where $\epsilon$ is an upper bound to the rounding error using the machine's double-precision floating point arithmetic).\\
This approach has the benefit of allowing in principle for any maximization target (not just $E$). In particular, any unweighted formulas (unnegated and negated facts as well as rules) can be used as (positive) examples.

%\subsection{Searching for a Maximum Entropy Distribution Using Parallel Stochastic Gradient Descent}
%\label{linSystemSolvingMaxEntropy}
%tbw.
%
%Parallelized Stochastic Gradient Descent

\newpage 

% NOPE:
%\subsection{Inference using Metropolis-Hastings Sampler}
%
%The inference algorithms described so far are as accurate as the underlying arithmetic and linear algebra routines allow,
%but they are slow and thus only suitable for rather small inference tasks. In the following, we describe an  MCMC (Markov Chain Monte Carlo) sampling algorithm which computes approximate solutions and which is typically much faster than the previous algorithms.\\
%
%An obvious way to approach MCMC in \prasp would be to sample from the joint probability $Pr(x_1, ...,x_D) $ of all atoms $x_i$. However, this would result in a large number of ASP solver calls, since we would need to compute samples component-wise for each dimension $1..D$ (i.e., each atom) in turn, holding the others fixed. We can reduce the number of solver calls by sampling component-wise 
%
%
%
%
%In the special case that all atoms are independent and their probabilities are given, we could compute the probability of each possible world directly as the product of atom probabilities.\\
%
%
%
%This algorithm is akin to MC-SAT \cite{mln} but is based on ASP and also differs in a few other details.

\section{Configuration and startup options}
\label{confopts}

\subsection{Starting \prasp}
\label{starting}

\noindent Run \prasp using \verb|sh \prasp.sh <arguments>|
(Linux, Mac OS X)\\
or \verb|\prasp.bat <arguments>| (Windows).\\

%\begin{changemargin}{1cm}{1cm} 
There has to be exactly one background knowledge file (``\prasp program'') per command-line. All inference and learning tasks triggered by a certain \prasp call share the same background knowledge.\
Unless optional command-line arguments  \hyperref[cmdline:bgk]{\texttt{--bgk}}, \hyperref[cmdline:query]{\texttt{--query}},  \hyperref[cmdline:learn]{\texttt{--learn}} or  \hyperref[cmdline:examples]{\texttt{--examples}} are provided (see below), tasks are determined by file name extensions, as follows:\\

Use file name ending \verb|.prasp| for the background knowledge file (= rules and facts whose probabilities are known already), \verb|.query| for a file with queries (= formulas whose probabilities are unknown but can be \textit{inferred} from the background knowledge),
\verb|.hypoth| for a file with hypotheses formulas (= formulas whose probabilities should be \textit{learned} from example data), and \verb|.examples| for a file with learning example data.\\

If you provide a \verb|.prasp| and at least one \verb|.query| file, an \textit{inference} task will be performed (see description of command-line argument \hyperref[cmdline:query]{\texttt{--query}} for performance-related implications of specifying multiple query-files).
If you provide a \verb|.prasp| and a \verb|.hypoth| and an \verb|.examples| file, a weight \textit{learning} task will be performed. \\

As additional inference-related tasks, you can let \prasp print \verb§n§ samples from the probability distribution over possible worlds (switch \hyperref[cmdline:pwsamples]{\texttt{--pwsamples}}) or print this distribution and its entropy (switch \hyperref[cmdline:pwdistr]{\texttt{--pwdistr}}).\\

By default, all command-line switches are ``off''. Switches refer to all tasks, regardless of their position in the command-line.\\

\noindent Example:\\
\verb|sh \prasp.sh examples/example1.prasp examples/example1.query|\\
\noindent This command-line specifies an inference task (query answering).\\

\noindent Further examples can be found at page \pageref{basic}ff.

%\end{changemargin}

\subsection{Influencing the Inference Pipeline}
\label{pipeline}

Inference consists of several subsequent steps (sub-tasks) selectable and configurable using command-line arguments.\\

Figure \ref{fig:pipeline} shows an overview of the various alternative successions of steps and the most important command-line arguments which activate or configure the respective step.\\

\noindent In this regard, also see Section \ref{performance}. \\

\noindent The general (configurable) sequence of inference steps is as follows:

\begin{enumerate}
	\item The background knowledge (\prasp program) is being parsed. At this stage, meta-statements (e.g., event independence declarations) are being resolved and non-ground weights are computed. Formulas annotated with double- or triple-square brackets (but not necessarily other non-ground formulas) are grounded (please see for Sect. \ref{syntax} for details).
	\item The so-called \textit{spanning program} is generated. This program represents the uncertainty found in the background knowledge but without probabilistic weights. Basically, each weighted formula (soft constraint) $[p] f$ with $p < 1$ is converted into a disjunction $f \vee \neg f$, whereas each unweighted formula (hard constraint) is copied directly into the spanning program. Details are presented in Sect. \ref{semantics}.  
	\item The spanning program is converted into ASP syntax (if FOL syntax was used in the \prasp program)
	\item The result is optionally simplified (see switches \hyperref[cmdline:mod0]{\texttt{--mod0}} and \hyperref[cmdline:mod1]{\texttt{--mod1}} in Sect. \ref{commandline})
	\item Using a variety of alternative sampling algorithms, an initial random sample of models is computed. This so-called initial sample is a multiset or set of possible worlds. Each possible world is an answer set (stable model) of the simplified spanning program. The initial sampling methods are described informally in Sect. \ref{commandline} (switch \hyperref[cmdline:initsample]{\texttt{--initsample}}) and formally in Sect. \ref{corealgos}.\\
	It is also possible to let \prasp generate \textit{all} models of the spanning problem, although this is normally too time consuming unless the system is small.
	\item In the following inference stage, the probabilities of the sampled possible worlds are computed, either using linear system solving (default), or linear programming (if interval results are requested, see \hyperref[cmdline:intervalresults]{\texttt{--intervalresults}}), or using approximation algorithms - currently, these are \textit{simulated annealing} (\hyperref[cmdline:simanneal]{\texttt{--simanneal}}) or \textit{iterative refinement} (\hyperref[cmdline:itrefinement]{\texttt{--itrefinement}}).\\
	An important additional command-line switch useful in this regard is \hyperref[cmdline:nosolve]{\texttt{--nosolve}}. With this switch, linear system solving (or linear programming) is deactivated. If this switch is omitted, the result of simulated annealing is fed into the linear solver/optimizer (which is usually redundant and not desirable). However, iterative refinement replaces linear system solving / linear programming without \hyperref[cmdline:nosolve]{\texttt{--nosolve}} and can thus theoretically be combined with simulated annealing (the reason for the different treatment of both algorithms is that simulated annealing is formally a sampling algorithm, whereas iterative refinement computes probabilities and is thus seen as a solver).\\
	It is also possible to use \hyperref[cmdline:nosolve]{\texttt{--nosolve}} without any inference algorithms - in that case, query probabilities are directly deduced from the initial sample, which is possible if certain assumptions hold (see Sect. \ref{merecounting}).
	\item Finally, the probability distribution over possible worlds is used to compute the probabilities of the annotated formulas (e.g., \verb§[?] f§ or \verb§[?|c] f§) in the \verb§.query§-file.
\end{enumerate}

%\newgeometry{left=2.5cm,bottom=4cm}
\vspace{-1cm}
\begin{figure}[H]
	\centering
	\caption{\prasp v0.9.0 inference pipeline alternatives (see Fig. \ref{fig:parsingOrder} for details on first stage)} 
	\label{fig:pipeline}
	\includegraphics[width=1.3\linewidth, trim=60mm 0 0 0mm, keepaspectratio]{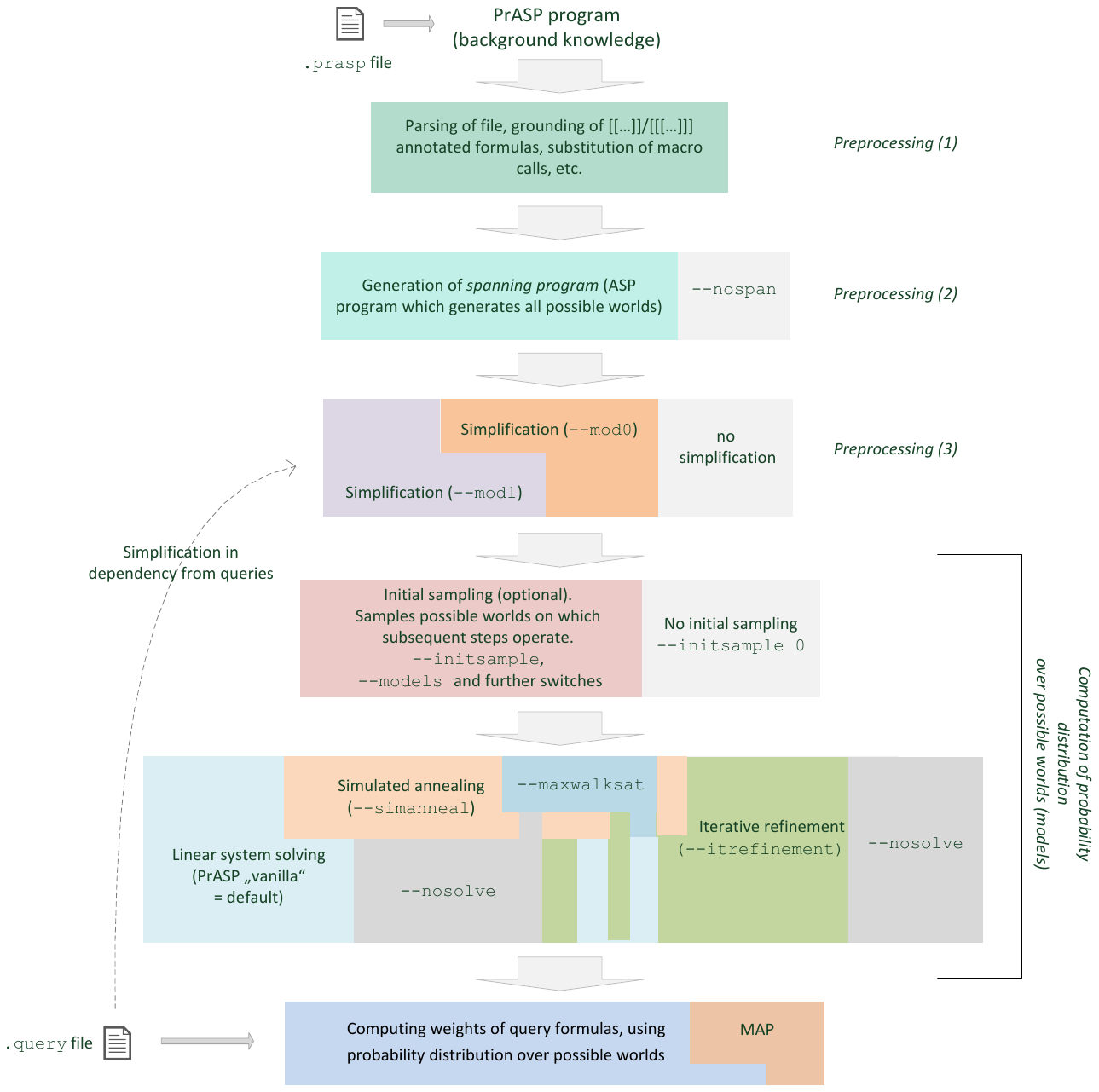}
	
\end{figure}

%\restoregeometry % doesnt work
%\newgeometry{left=3.5cm,bottom=6cm}

\subsection{Command-line arguments}
\label{commandline}

\prasp \version\ is configured mainly using command-line options. Most of the switches and other arguments listed below are \underline{not} required for simple use cases. However, it is suggested that you make yourself familiar at least with the inference pipeline (Fig. \ref{fig:pipeline}) and optimization command-line switches ( \hyperref[cmdline:o1]{\texttt{-o1}},  \hyperref[cmdline:o1asp]{\texttt{-o1asp}}, \hyperref[cmdline:o2]{\texttt{-o2}}, etc). Basic knowledge about alternative approximate solving approaches \hyperref[cmdline:itrefinement]{\texttt{--itrefinement}} and \hyperref[cmdline:simanneal]{\texttt{--simanneal}} is also useful. This is because
in ``vanilla mode'' (the default), settings are very conservative. More concretely, using default settings, \prasp allows for maximum expressiveness in knowledge and query files but performs larger-scale tasks rather slowly. Therefore large inference or learning tasks require the use of optimization switches. \\

Please observe that inference-related command-line options also affect learning, since learning relies on iterated inference.

%\\textbf(\{[^}]+\})
%\subsection*{\normalsize\1}

{\parindent0pt 
	\subsection*{\normalsize{\texttt{--help}}} for a short version of this subsection.
	
	\addcontentsline{toc}{subsection}{\texttt{\texttt{--help}}} \label{cmdline:help}
	
	\subsection*{\normalsize{\texttt{--verbose}}} shows additional information about inference and learning tasks. This information
	might be useful for the optimization of input files. Further information- and debugging-related switches are \hyperref[cmdline:check]{\texttt{--check}}, \hyperref[cmdline:pwdistr]{\texttt{--pwdistr}}, \hyperref[cmdline:pwsamples]{\texttt{--pwsamples}}, \hyperref[cmdline:showentropy]{\texttt{--showentropy}}, \hyperref[cmdline:debug]{\texttt{--debug}}, \hyperref[cmdline:showindeps]{\texttt{--showindeps}}, \hyperref[cmdline:showexpansion]{\texttt{--showexpansion}} and \hyperref[cmdline:showspan]{\texttt{--showspan}}, which are all described further below. 
	
	\addcontentsline{toc}{subsection}{\texttt{\texttt{--verbose}}} \label{cmdline:verbose}
	
	\subsection*{\normalsize{\texttt{--bgk file}} \normalfont{\ or\ } \textbf{\texttt{-b file}}}  is used to specify background knowledge (that is, a \prasp program). If \verb|--bgk| and \verb|-b| are omitted but a file name is provided (which is not bound by any other file name accepting command-line argument), the background knowledge file is identified based on its file name extension \verb|.prasp|. Exactly one background knowledge file needs to be specified per \prasp call.
	
	\addcontentsline{toc}{subsection}{\texttt{\texttt{--bgk}}} \label{cmdline:bgk}
	\addcontentsline{toc}{subsection}{\texttt{\texttt{-b}}}
	
	\subsection*{\normalsize{\texttt{--query file\_1 ... file\_n}} \normalfont{\ or\ } \textbf{\texttt{-q file\_1 ... file\_n}}} to approximate the probabilities of given query formulas within each of these files (\textit{probabilistic inference}). If \verb|--query| and \verb|-q| are omitted but a file name is provided, the query file is identified based on its file name extension \verb|.query|\\
	Additionally, a background knowledge file is required (see \hyperref[cmdline:bgk]{\texttt{--bgk}}).\\
	It is possible to specify multiple \hyperref[cmdline:query]{\texttt{--query}}'s per command-line, each with multiple file names as arguments. All query files in the command-line share the same background knowledge. However, for each \hyperref[cmdline:query]{\texttt{--query}} occurrence (and also for each query file which is not bound by any explicit \hyperref[cmdline:query]{\texttt{--query}} argument but identified by its name ending), the background knowledge is processed anew. This way, \prasp can consider different optimization strategies for different sets of query files. 
	
	\addcontentsline{toc}{subsection}{\texttt{--query}}  \label{cmdline:query}
	\addcontentsline{toc}{subsection}{\texttt{\texttt{-q}}}
	
	\subsection*{\normalsize{\texttt{--learn file}} \normalfont{\ or\ } \textbf{\texttt{-l file}}} to learn weights of given hypothesis formulas within file \verb|file|. If \hyperref[cmdline:learn]{\texttt{--learn}} and \verb|-l| are omitted but a file name is provided (which is not bound by any other file name accepting command-line argument), the hypothesis file is identified based on its file name extension \verb|.hypoth|. Maximally one hypothesis file can be specified with each \prasp call. If \hyperref[cmdline:learn]{\texttt{--learn}} or \verb|-l| is specified, also \hyperref[cmdline:examples]{\texttt{--examples}} or \verb|-e| needs to be specified.\\
	Additionally, a background knowledge file is required (see \hyperref[cmdline:bgk]{\texttt{--bgk}}).
	
	\addcontentsline{toc}{subsection}{\texttt{--learn}}  \label{cmdline:learn}
	\addcontentsline{toc}{subsection}{\texttt{\texttt{--l}}}
	
	\subsection*{\normalsize{\texttt{--examples file}} \normalfont{\ or\ } \textbf{\texttt{-e file}}} to specify learning examples (together with \hyperref[cmdline:learn]{\texttt{--learn}} or \verb|-l|). If \hyperref[cmdline:examples]{\texttt{--examples}} and \verb|-e| are omitted but a file name is provided (which is not bound by any other file name accepting command-line argument), the learning examples file is identified based on its file name extension \verb|.examples|. Maximally one examples file can be specified with each \prasp call.
	
	\addcontentsline{toc}{subsection}{\texttt{--examples}}  \label{cmdline:examples}
	\addcontentsline{toc}{subsection}{\texttt{\texttt{-e}}}
	
	\subsection*{\normalsize{\texttt{--pwsamples n}}} prints \verb§n§ sampled possible worlds (uniformly sampled with replacement).\\
	Notice that \hyperref[cmdline:pwsamples]{\texttt{--pwsamples n}} has no influence on inference or learning results, it is a purely informative task with no connection to initial sampling or \hyperref[cmdline:models]{\texttt{--models}}.
	
	\addcontentsline{toc}{subsection}{\texttt{--pwsamples}}  \label{cmdline:pwsamples}
	
	\subsection*{\normalsize{\texttt{--pwdistr}} n} prints the probability distribution over possible worlds and its entropy as obtained using initial sampling (see \hyperref[cmdline:initsample]{\texttt{--initsample}}). If additionally \hyperref[cmdline:ndistrs n]{\texttt{--ndistrs n}} is specified and the default inference approach is being used (linear system solving), \verb§n§ distributions and their entropies are printed.\\
	If \verb§n§ is specified, the number of printed worlds is limited to \verb§n§. Since the results are ranked, \hyperref[cmdline:pwdistr]{\texttt{--pwdistr}} can be used for MAP inference if the specified sampling or inference approach produces a probability distribution over possible worlds (or the entire set of possible worlds are top-ranked in terms of their probabilities, e.g., \hyperref[cmdline:maxwalksat]{\texttt{--maxwalksat}}). 
	
	\addcontentsline{toc}{subsection}{\texttt{--pwdistr}}  \label{cmdline:pwdistr}
	
	\subsection*{\normalsize{\texttt{--showentropy}}} prints the entropy of the distribution over possible worlds. In case \hyperref[cmdline:mod1]{\texttt{--mod1}} was used, multiple entropies are printed. No entropies are printed if interval results have been requested using \hyperref[cmdline:intervalresults]{\texttt{--intervalresults}}.\\
	
	\addcontentsline{toc}{subsection}{\texttt{--showentropy}}  \label{cmdline:showentropy}
	
	\noindent \underline{\textbf{The following optimization switches}} provide shortcuts for frequently combined command-line arguments with an influence on inference performance and accuracy. A switch with a higher number after the ``\verb§o§'' is  not ``more optimizing'' but just different from the other \verb§-o§ switches:
	
	\noindent \subsection*{\normalsize{\texttt{-o1}}} is a shortcut for \hyperref[cmdline:noindepconstrs]{\texttt{--noindepconstrs}}\hyperref[cmdline:noautoindeps]{\texttt{--noautoindeps}}\hyperref[cmdline:itrefinement]{\texttt{ --itrefinement}}
	
	\addcontentsline{toc}{subsection}{\texttt{-o1}}  \label{cmdline:o1}
	
	\noindent \subsection*{\normalsize{\texttt{-o1asp}}} is a shortcut for \hyperref[cmdline:o1]{\texttt{ -o1}}\hyperref[cmdline:folconv]{\texttt{ --folconv none}}
	
	\addcontentsline{toc}{subsection}{\texttt{-o1asp}}
	
	\noindent \subsection*{\normalsize{\texttt{-o2}}} is a shortcut for 
	\hyperref[cmdline:noindepconstrs]{\texttt{ --noindepconstrs}}
	\hyperref[cmdline:noautoindeps]{\texttt{--noautoindeps}}
	\hyperref[cmdline:initsample]{\texttt{ --initsample 4}}
	\hyperref[cmdline:itrefinement]{\texttt{ --itrefinement}}
	
	\addcontentsline{toc}{subsection}{\texttt{-o2}}\label{cmdline:o2}
	
	\noindent \subsection*{\normalsize{\texttt{-o2asp}}} is a shortcut for 
	\hyperref[cmdline:o2]{\texttt{ -o2}}
	\hyperref[cmdline:folconv]{\texttt{ --folconv none}}
	
	\addcontentsline{toc}{subsection}{\texttt{-o2asp}}
	If your input files are in pure ASP syntax, this switch should be used instead of \verb§-o2§
	
	\noindent \subsection*{\normalsize{\texttt{-o3}}} is a shortcut for 
	
	\hyperref[cmdline:mod1]{\texttt{ --mod1}}
	\hyperref[cmdline:simanneal]{\texttt{ --simanneal}}
	\hyperref[cmdline:nosolve]{\texttt{ --nosolve}}
	\hyperref[cmdline:ignoredeclindeps]{\texttt{ --ignoredeclindeps}}
	\hyperref[cmdline:noindepconstrs]{\texttt{ --noindepconstrs}}
	\hyperref[cmdline:noautoindeps]{\texttt{ --noautoindeps}}
	
	\addcontentsline{toc}{subsection}{\texttt{-o3}}\label{cmdline:o3}
	
	\noindent \subsection*{\normalsize{\texttt{-o3asp}}} is a shortcut for 
	\hyperref[cmdline:o3]{\texttt{ -o3}}
	\hyperref[cmdline:folconv]{\texttt{ --folconv none}}\\
	If your input files are in pure ASP syntax, this switch should be used instead of \verb§-o3§
	
	\addcontentsline{toc}{subsection}{\texttt{-o3asp}}
	
	\subsection*{{\normalsize \texttt{-o4}}} is a shortcut for\\
	\hyperref[cmdline:mod1]{\texttt{ --mod1}}
	\hyperref[cmdline:noautoindeps]{\texttt{ --noautoindeps}}
	\hyperref[cmdline:noindepconstrs]{\texttt{ --noindepconstrs}}
	\hyperref[cmdline:ignoreentropy]{\texttt{ --ignoreentropy}}
	
	\addcontentsline{toc}{subsection}{\texttt{-o4}}\label{cmdline:o4}
	
	\subsection*{\normalsize{\texttt{-o4asp}}}  is a shortcut for \hyperref[cmdline:o4]{\texttt{ -o4}} \hyperref[cmdline:folconv]{\texttt{ --folconv none}}
	
	If your input files are in pure ASP syntax, this switch should be used instead of \verb§-o4§
	
	\addcontentsline{toc}{subsection}{\texttt{-o4asp}}
	
	\subsection*{\normalsize{\texttt{--initsample m}}} specifies the method used for \textit{initial sampling}, that is, the way \prasp generates a list of random models (possible worlds) of the spanning program used by the subsequent solving or annealing step. These initial models are passed on to the next step in the inference pipeline (e.g., they might serve as the initial distribution for simulated annealing or they are used in the linear system's equations/inequalities). The chosen initial sampling approach and the number of initial samples (set with \hyperref[cmdline:models]{\texttt{--models}}) strongly influence inference and learning speed and accuracy (in the ideal case, \textit{all} models would be computed, but this is often computationally intractable).\\
	
	Initial sampling either produces a set or a multiset (where the element frequencies are interpreted as a possibly non-uniform probability 
	distribution over possible worlds). Whether the additional information carried by the multiset (vs. a mere set) is used by the following algorithm in the inference pipeline depends on the respective next algorithm and its parameters (e.g., simulated annealing can use the result of initial sampling as a multiset, whereas the default inference approach (linear system solving) discards information about model frequencies).\\
	
	Algorithms used for initial sampling are presented in Sect. \ref{corealgos}.\\
	
	If argument \verb§m§ or \hyperref[cmdline:initsample]{\texttt{--initsample}} are omitted, \prasp heuristically chooses an initial sampling approach based on the information it gathered from the command-line and the background knowledge. Use \hyperref[cmdline:verbose]{\texttt{--verbose}} to see the selected approach (which is not guaranteed to be the optimal one!).\\
	
	The indicative number \textit{n} of sampled models is specified by \hyperref[cmdline:models]{\texttt{--models}}. If \hyperref[cmdline:models]{\texttt{--models}} is omitted, a heuristically determined number of models are generated.\\
	
	It is recommended to compare the sets or possible worlds and their entropies achieved with different approaches (in combination with different settings for \hyperref[cmdline:unisample ]{\texttt{--unisample}}and other sampling-related parameters) in order to identify the optimal approach for the respective use case. See \hyperref[cmdline:pwdistr]{\texttt{--pwdistr}}. Also, switch \hyperref[cmdline:check]{\texttt{--check}} might be useful in this regard.\\
	
	\texttt{m} = 0: empty set,\\
	
	\texttt{m} = 1: $\approx$ \textit{n} random models sampled with replacement from a near-uniform distribution. The concrete sampling approach can optionally be specified using \hyperref[cmdline:unisample]{\texttt{--unisample}}, otherwise the approach is chosen by \prasp (use \hyperref[cmdline:verbose]{\texttt{--verbose}} to see which one).\\
	
	\texttt{m} = 2: any \textit{n} models  (random or non-random). In practice, this simply gives the first \textit{n} answer sets in the stream of answer sets computed by the ASP solver. If $n = 0$ (i.e., \hyperref[cmdline:models]{\texttt{--models 0}}), \textit{all} models of the spanning program will be generated (intractable unless the system is small). \\
	
	\texttt{m} $\geq$ 3: $\approx$ \textit{n} random models sampled from a distribution of models such that the frequencies of models of uncertain formulas correspond approximately to the given weights of these models \textit{if} certain assumptions are met (e.g., mutual independence). So in contrast to \texttt{m} = 2 the formula weights are not ignored here.  \\
	The model frequencies are represented by means of the number of occurrences of the respective model in the initial sampling result list. Whether any duplicates in the resulting sample will be retained or discarded depends on the next inference stage.\\
	
	Remark: most sampling approaches work significantly faster if Clingo 4 is used as ASP solver (``remote sampling'') (see Sect. \ref{installation}). With Clingo 4, remote sampling is active by default where the selected sampling approach can use it. See \hyperref[cmdline:noremotesampling]{\texttt{--noremotesampling}} for a discussion and requirements.\\
	
	Note that even weighted sampling approaches need to sample from a uniform or near-uniform distribution at some point of the sampling algorithm, that is, even with approach $\geq$ 3, sampling is influences by the approach specified with \hyperref[cmdline:unisample]{\texttt{--unisample}}. The selected sampling method massively influences inference and learning performance and quality.\\
	
	Methods \texttt{m} $\geq$ 3 in detail:\\
	
	\texttt{m} = 3: a distribution is generated using random sampling with replacement such that the frequencies of sampled models fully reflect the products of the probabilities of the mutually independent weighted formulas which hold in these models. To make this
	tracktable, the number of considered combinations of independent formulas might has to be limited using \hyperref[cmdline:limitindepcombs]{\texttt{--limitindepcombs}}. \\
	
	\verb§m§ = 4: all models are randomly sampled (\textit{stratified weighted flip-sampling}, with replacement) \textit{as if} weighted formulas were mutually independent (regardless of actual or declared independence). See Algorithm \ref{alg:flipsamp} for details.\\% Flip-sampling is a form of rejection sampling where attached weights define a prior distribution.\\
	If this independence assumption holds and the target number of samples is not too large, queries can be solved  quite efficiently with \verb§m§ = 4 and \hyperref[cmdline:nosolve]{\texttt{--nosolve}}. Otherwise, using this option may have little benefit, or you might get no result at all (if the initial sampling process generates a lot of inconsistent sample programs). %This approach cannot be used if there are
	%any conditional probabilities in background knowledge. 
	Note that the aforementioned independence assumption 
	does \textit{not} necessarily mean that with the resulting approximate distribution $Pr(a \wedge b) = Pr(a)Pr(b)$ holds for any actually independent events $a$ and $b$, as this also depends from the size of the set of samples (specified with \verb§--models§). The same is true for entropy: if approach 1 is used, the entropy of the resulting distribution depends on the number of samples even if in reality all events are actually mutually independent. Caveat: with \verb§--models 0§, PrASP computes the set of all models of the spanning program (without replacement), which would not be useful in combination with \verb§--initsample 4§.\\
	
	\verb§m§ = 5: models are randomly sampled (\textit{stratified weighted flip-sampling with consideration of declared/discovered independence}, with replacement).
	%
	%generating random programs as follows: To the spanning program (see Sect. \ref{semantics}) a randomly selected subset of mutually independent weighted formulas is added such that the selection probability of each such subset is the product of the weights of these weighted formulas. For each formula not part of this subset its negation is added to the generated program. One model of the generated program (chosen among all models using near-uniform sampling) is added to the set of sampled possible worlds. This continues until the target number of sampled possible worlds has been reached.\\ 
	This approach is similar to approach 4 above, but uses declared or discovered event (formula) independence instead of making an assumption about universal independence. See Algo. \ref{alg:flipsamp} for a detailed description of this approach.\\
	%In our experiments, we found this sampling approach particularly useful in connection with iterative refinement (\hyperref[cmdline:itrefinement]{\texttt{--itrefinement}}).\\
	
	Remark: flip-sampling can cause Clingo messages of the form ``warning: x is never defined''. If the respective message doesn't turn up without sampling activated then you can ignore it.\\
	
	\verb§m§ = 6: Sampling according Algorithm \ref{alg:flipsamp} (\textit{unstratified weighted flip-sampling}). Same as approach \verb§m§ = 4 with the exception that random numbers are not stratified - see Algorithm \ref{alg:flipsamp} for details.\\
	
	\verb§m§ = 7: Sampling according Algorithm \ref{alg:flipsamp} (\textit{unstratified weighted flip-sampling with consideration of declared/discovered independence}). Same as approach \verb§m§ = 5 with the exception that random numbers are not stratified - see Algorithm \ref{alg:flipsamp} for details.
	
	\addcontentsline{toc}{subsection}{\texttt{--initsample}}  \label{cmdline:initsample}
	
	\subsection*{\normalsize{\texttt{--models n}}} specifies the target number of \textit{initial models} (possible worlds) computed during initial sampling using the approach specified using \hyperref[cmdline:initsample]{\texttt{--initsample}}. This number and the sampling approach strongly influence inference and learning accuracy and duration.\\
	
	\hyperref[cmdline:models]{\texttt{--models}} is typically used together with approximate inference approaches whereas limiting the number of models in order to speed up the default inference approach is usually not a good idea (results might be highly inaccurate). It is even possible that a lower number of models increases overall inference  duration, if the smaller number of models makes the solving algorithm require more time to converge to a solution for the linear system.\\
	
	\verb§n§ is just a target - depending on the chosen sampling approach, \verb§n§ cannot be guaranteed (actual number of sampled models might be lower - use \hyperref[cmdline:verbose]{\texttt{--verbose}} to see this number).\\
	
	Whether initial sampling is used or not is specified by \hyperref[cmdline:initsample]{\texttt{--initsample}}, \hyperref[cmdline:models]{\texttt{--models}} just specifies the number of models sampling should generate. Exception: if \verb§n§ is 0 and the sampling approach set with \hyperref[cmdline:initsample]{\texttt{--initsample}} is not 0, all models will be computed without sampling. \\
	If \hyperref[cmdline:models]{\texttt{--models}} is omitted, the target number of sampled models is heuristically determined.\\
	% as follows:\\
	%In case the sampling approach specified by \hyperref[cmdline:initsample]{\texttt{--initsample}} is 1 or 5,\\
	% $100 \left({\#models_{estim}}^{0.18}-1\right))$ is used, unless $\#models_{estim} \leq 150$ (in which case 100 is used). At this, $\#models_{estim}$ is a rough estimation of the total number of models of the spanning program, computed using an approximate model counting approach \cite{emc}).\\
	%Please note that the computation of $\#models_{estim}$ is a rather costly task in itself.\\
	In case the sampling approach specified by \hyperref[cmdline:initsample]{\texttt{--initsample}} is 3 or 4, the default number of samples is 
	either as for approaches 1 and 5 (with \hyperref[cmdline:maxentropy]{\texttt{--maxentropy}}) or 20 (otherwise).\\
	The reason for the small default target number for approaches 3 and 4 without \hyperref[cmdline:maxentropy]{\texttt{--maxentropy}} is that these methods
	are often able to compute valid initial distributions even for a very small number of possible worlds. Increasing this number is then only required if the distribution entropy should be increased or if there are additional constraints to be observed by later inference steps (such as independence constraints). In the latter case, a following simulated annealing step in the inference pipeline automatically enhances the initial list of sampled models, provided \hyperref[cmdline:simanneal]{\texttt{--simanneal}} is also specified.\\
	
	The number of actually computed samples and the estimated total number of spanning program models $\#models_{estim}$ are printed if \hyperref[cmdline:verbose]{\texttt{--verbose}} is provided. \\
	Also see \hyperref[cmdline:unisample]{\texttt{--unisample}}, \hyperref[cmdline:xorconf]{\texttt{--xorconf}}, \hyperref[cmdline:flipsampconf]{\texttt{--flipsampconf}} and \hyperref[cmdline:sirndconf]{\texttt{--sirndconf}}.
	
	\addcontentsline{toc}{subsection}{\texttt{--models}}  \label{cmdline:models}
	
	\subsection*{\normalsize{\texttt{--simanneal energy minTemp tempDecr samplingMethod initTemp samplesPerStep targetAll}}} uses simulated annealing as approximative inference method. It uses the models generated by initial sampling (see \hyperref[cmdline:initsample]{\texttt{--initsample}}) as a starting point. The initial samples are treated as a multiset where the normalized count of a certain model within the multiset is interpreted as the initial approximation of the probability of that model (which is then refined by the simulated annealing algorithm). A formal description can be found in Sect. \ref{simanneal}.\\
	
	Because simulated annealing in \prasp relies heavily on sampling, it should be used with Clingo 4 (see Sect. \ref{installation} - however, it works with Clingo 3 too).\\
	
	All arguments of \hyperref[cmdline:simanneal]{\texttt{--simanneal}} are optional.\\
	Defaults: \verb§energy=0.05, minTemp=1e-150, tempDecr=0.95,§ \\
	\verb§samplingMethod=0, initTemp=5, samplesPerStep=1, targetAll=true§ \\
	
	Example: \hyperref[cmdline:simanneal 0.001 1e-150 0.95 1 5 1 false]{\texttt{--simanneal 0.001 1e-150 0.95 1 5 1 false}} \\
	
	\verb§energy§ ($\geq 0$) is the acceptable inaccuracy. Lower values mean more accurate results.
	More precisely, this parameter specifies the maximum acceptable euclidean distance between the vector of given formula weights and the vector of weights according to the current solution candidate. \\
	
	\verb§minTemp§ is the temperature at which the algorithm stops even if the desired accuracy has not been reached yet.\\
	
	\verb§tempDecr§ is the factor by which the current temperature is multiplied in each step.\\
	
	\verb§samplingMethod§ specifies the approach simulated annealing generates new samples until it found a distribution which approximates the given constraints well enough. This should not to be confused with the approach used to compute \textit{initial models} (\hyperref[cmdline:initisample]{\texttt{--initisample}}) (the set of random models simulated annealing uses as a starting point).\\ 
	
	If \verb§samplingMethod§ is 0 (default), at each simulated annealing step, models are sampled from a near-uniform distribution. Use \hyperref[cmdline:unisample]{\texttt{--unisample}} to specify the concrete approach used here.\\
	If \verb§samplingMethod§ is 1, simulated annealing samples are at each step obtained using a sampling approach which takes into account declared or discovered formula independence assumptions.\\
	\verb§samplingMethod§ 2 works like method 1, but makes the assumption that \textit{all} weighted formulas are mutually independent, regardless of meta-statements. If this is not actually the case, this
	sampling approach is very slow.\\
	Methods with \verb§samplingMethod§ $\geq$ 3 are those one can specify with \hyperref[cmdline:initsample]{\texttt{--initsample}} (with the difference of course that they are used for simulated annealing here, and that at each simulated annealing step only \verb§samplesPerStep§ random models are computed).\\
	%If \verb§samplingMethod§ is 3, a heuristic sampling approach is used which switches at each step randomly the truth values of weighted formulas. \\ % akin to initsample method 4 !
	
	Details about these sampling step methods can be found in Sect. \ref{simanneal} and Sect. \ref{simannealspeedtipps}.\\
	
	\verb§initTemp§ is the initial temperature, and \verb§samplesPerStep§ denotes the number of samples computed per step.\\ %(ignored if sampling method is rejection sampling).\\
	
	If \verb§targetAll§ is \verb§false§, only formulas with a weight $< 1$ are targeted when minimizing the energy, whereas otherwise all formulas are targeted (default: \verb§false§). ``Targeted'' means here that in the calculation of the energy, the frequency of the respective formula in the current best set of sampled models is included.\\
	
	% In other words, if \verb§targetAll§ is \verb§false§, simulated annealing doesn't care whether formulas with given weight 1 also would have probability 1 in a query result. This setting only makes sense if the weight target for formulas with weight 1 is already (approximately) met in the initial set of samples and the sampling steps performed by simulated annealing don't change the approximated weights of these formulas much (you can check these criteria using \hyperref[cmdline:check]{\texttt{--check}}). In that case, \verb§false§ speeds up simulated annealing greatly, otherwise you get wrong results in case your query formulas depend on truth values of formulas with weight 1.     \\
	
	\hyperref[cmdline:simanneal]{\texttt{--simanneal}} is typically used together with \hyperref[cmdline:initsample]{\texttt{--initsample}}, \hyperref[cmdline:nosolve]{\texttt{--nosolve}}\\ and \hyperref[cmdline:noindepconstrs]{\texttt{--noindepconstrs}}. If one of these is omitted, accuracy is increased but processing time (massively) increased. \prasp shows a warning in such cases. Also consider combining \hyperref[cmdline:simanneal]{\texttt{--simanneal}} with further switches such as \hyperref[cmdline:mod1]{\texttt{--mod1}}. The sampling step performed by simulated annealing can be further configured using \hyperref[cmdline:xorconf]{\texttt{--xorconf}}, \hyperref[cmdline:flipsampconf]{\texttt{--flipsampconf}} and \hyperref[cmdline:sirndconf]{\texttt{--sirndconf}}.\\
	
	\textit{Important}: \hyperref[cmdline:simanneal]{\texttt{--simanneal}} works best if all formulas (including query formulas) are so-called
	``simple formulas'', see \ref{simpleFormulas}. If this is not possible and inference is too slow, consider specifying an alternative model checking algorithm using \hyperref[cmdline:ascheckmode]{\texttt{--ascheckmode}}. Also, \hyperref[cmdline:simanneal]{\texttt{--simanneal}} profits from mutually independent formulas.\\
	
	If \hyperref[cmdline:maxentropy]{\texttt{--maxentropy}} is specified, simulated annealing takes the entropy (\ref{intervalresults}) into account when computing the current energy, providing a heuristic approach to the \textit{principle of maximum entropy}. However, in contrast to some other inference approaches in \prasp (like \hyperref[cmdline:itrefinement]{\texttt{--itrefinement}}) simulated annealing can generally \textit{not guarantee} a maximum entropy solution. Also, \hyperref[cmdline:maxentropy]{\texttt{--maxentropy}} influences the initial set of models (see \hyperref[cmdline:initsample]{\texttt{--initsample}} and \hyperref[cmdline:models]{\texttt{--models}}).\\
	\hyperref[cmdline:maxentropy]{\texttt{--maxentropy}} is incorporated into the ``energy'' $e$ using $e = d + \frac{1}{entropy}$ (where $d$ is the euclidean distance between current and target weights, as described above). However, even with \hyperref[cmdline:maxentropy]{\texttt{--maxentropy}}, parameter \verb§energy§ refers only to $d$,
	whereas the acceptable entropy is specified using a parameter of command-line option \hyperref[cmdline:maxentropy]{\texttt{--maxentropy}}.\\
	
	For more information about \hyperref[cmdline:simanneal]{\texttt{--simanneal}}, please refer to Sect. \ref{alg:simanneal} and Sect. \ref{simannealspeedtipps}. 
	
	\addcontentsline{toc}{subsection}{\texttt{--simanneal}}  \label{cmdline:simanneal}

	\subsection*{\normalsize{\texttt{--nospan}}}
	With this switch, no spanning program is being created. Useful, e.g., in connection with   \hyperref[cmdline:maxwalksat]{\texttt{--maxwalksat}}.  
	
	\addcontentsline{toc}{subsection}{\texttt{--nospan}}  \label{cmdline:nospan}
	
	\subsection*{\normalsize{\texttt{--maxwalksat costtarget maxit maxtries p repl}}}
	(experimental) uses the MaxWalkSAT (weighted maxsat) algorithm \cite{kautz,mln}, a weight-aware variant of the WalkSAT local search satisfiability solver, to compute a number of possible worlds with larger probabilities than the rest of the possible worlds (for MAP/MPE-style inference). In contrast to normal \prasp terminology, ``possible worlds'' here means \textit{any} satisfying truth assignments for the conjunction of the formulas in background knowledge, not answer sets of the spanning program.\\
	
	All formulas (annotated or not) need to be ``simple formulas'' in the sense of Section \ref{simpleFormulas} (you can still use non-ground or even FOL formulas with quantifiers too, but you need to make sure they are ``simple'' before the MaxWalkSAT phase during inference is reached, by annotating them appropriately, i.e., with \verb§[[§ or \verb§[[[§ weights).\\ 
	
	\verb§costtarget§ specifies the acceptable cost (sum of weights of formulas not satisfied in the current candidate model at which the algorithm stops; default: 1 - you might need to provide a different value here), \verb§maxit§ specifies the maximum number of local search iterations per trial (default: 100000), \verb§maxtries§ is the maximum number of trials (default: 10000), \verb§p§ is the probability of taking random steps (default: 0.3), and \verb§repl§ (\verb§true§ or \verb§false§) specifies whether discovered models are generated with or without replacement (default: \verb§true§, only relevant if you want to use MaxWalkSAT to produce more than one possible world).\\
	
	MaxWalkSAT\footnote{and also the other Markov Logic-like inference approach built into \prasp, see \texttt{--mln} and Section \ref{markov}} fits well into the overall \prasp inference pipeline (Sect. \ref{pipeline}), but in one respect it is exotic among the \prasp inference approaches - it treats formulas as propositional formulas and solves them in the sense of SAT solving (which is closely related to but not identical with ASP solving). It does not follow stable model semantics and it does not use an ASP solver, provided it is used with \hyperref[cmdline:nospan]{\texttt{--nospan}}. An external ASP grounder is required if any of the formulas in background knowledge or query file are non-ground (but in that case they must be annotated with double- or triple-square brackets, in order to produce ground formulas before the MaxWalkSAT stage!) and cannot be grounded internally. As a side note, observe that ASP grounding using an external grounded is already ``lazy'' in the sense that it omits ground instances about which it knows that they have no influence on models. \\	
	
	With \verb§--maxwalksat§, PrASP can be used without ASP solver, provided the query formulas are also ``simple formulas''. To enable this, specify additional switches \verb§--nospan --assumegroundersolver 0§\\. If your background knowledge is ground already, you also don't need an ASP grounder. \\
	
	MaxWalkSAT can be combined with \hyperref[cmdline:initsample]{\texttt{--initsample}}. There are two major ways to do this:
	
	\begin{itemize}
		\item specify \verb§--initsample 0 --models n --pwsamples --nosolve --nospan§\\
		\verb§ --noindepconstrs --ignoredeclindeps --noautoindeps§ where \verb§n§$\geqslant 1$. In that case, no initial samples are generated and MaxWalkSAT generates \verb§n§ possible worlds (approximately the ones with the highest probability) all by itself. If \verb§n§ = 1, this emulates Markov Logic's MAP inference.\\
		To project the model to certain literals, provide these literals as individual query formulas - where the probability is 1, the literal is true in the model (if \verb§n§$= 1$). Alternatively, use \hyperref[cmdline:pwsamples]{\texttt{--pwsamples}} to see the entire generated set of possible worlds.
		\item specify some \hyperref[cmdline:initsample]{\texttt{--initsample}} approach other than 0 (empty set) and, using \hyperref[cmdline:models]{\texttt{--models}}, a number of models $\geqslant 1$. In that case, the initial models are used as seeds for MaxWalkSAT: for each MaxWalkSAT try, a randomly picked initial model is used as the initial ``guess'' (which would be a random truth assignment instead if there are no initial models). This way, MaxWalkSAT replaces the $n$ sampled initial models with $n$ new models. 
	\end{itemize} 
	
	Although probably not advisable, it is possible to use the output of MaxWalkSAT as input (subset of the set of all possible worlds) for subsequent inference stages, such as iterative refinement (\hyperref[cmdline:itrefinement]{\texttt{--itrefinement}}). %With this, MaxWalkSAT would serve as a kind of sampling approach whose output provides a subset of possible worlds for the following inference approach.\\  %??: This doesn't make MaxWalkSAT a real ``sampling algorithm'' - for using \prasp for Markov Logic-style $Pr(x|c)$ inference, use \verb§--mln§ together with the MC-SAT sampling algorithm instead.\\ 
	
	It is also possible to let \prasp perform preceding steps besides \hyperref[cmdline:initsample]{\texttt{--initsample}}, such as \hyperref[cmdline:mod1]{\texttt{--mod1}} for program simplification.\\
	
	In some cases (see above) MaxWalkSAT is used together with \hyperref[cmdline:nospan]{\texttt{--nospan}}, as MaxWalkSAT doesn't require or use a spanning program.\\
	
	Caveats/issues:
	
	\begin{itemize}
		\item Under \prasp semantics, formula weights denote point probabilities or probability intervals. In order to emulate weights of ``hard constraints'' (using MLN terminology), our MaxWalkSAT variant counts formulas annotated with weight 1 as well as unannotated formulas as if they had weight 1000 (instead of the infinite weights of hard rules in MLN).  
		\item There is currently no MaxWalkSAT support for conditional probabilities or independence constraints in background knowledge, however, conditional probabilities in queries are fine.
		\item MaxWalkSAT maximizes the sum of given formula weights. It doesn't take into consideration at weight maximization any implied weights of formulas which don't appear explicitly in background knowledge. Switch \hyperref[cmdline:addnegf]{\texttt{--addnegf}} might be helpful in this regard.
		\item MaxWalkSAT is efficient only for MAP/MPE-style inference, i.e., to compute a single (or a very small) number of possible worlds with maximum (or close to maximum) weight. 
		\item All of the currently supported schemes for translating FOL formulas into ASP syntax (such as F2LP)  assume stable model semantics, that is, if used with a non-stable model-based approach to model generation (such as MaxWalkSAT), unexpected query results are to be expected.
		\item For Markov Logic Network-style $Pr(x|c)$ inference, see \verb§--mln§.
	\end{itemize}
	
	The algorithm of PrASP's MaxWalkSAT variant is presented in Sect. \ref{alg:maxwalksat}.\\
	
	 Remark: there is currently no LazySAT in \prasp, as LazySAT would be incompatible with \prasp's approach to grounding. However, notice that if \prasp employs the external grounder for grounding, this grounder typically applies simplifications. 
	
	\addcontentsline{toc}{subsection}{\texttt{--maxwalksat}} \label{cmdline:maxwalksat}
	
	\subsection*{\normalsize{\texttt{--itrefinement epsilon n retainCounts§}}} uses the inference approach introduced in \cite{spirit} (called \textit{iterative refinement} in \prasp terminology) to compute a possible worlds probability distribution with maximum entropy (where the entropy is bound by the initial set of possible worlds provided as input for this approach. With smaller sets, the entropy typically also decreases).\\
	
	Iterative refinement is often a good first choice if the system is just too large for the default ``vanilla'' inference approach. However, it requires a ``good'' set of initial samples - if the set of sampled models is too small or not random enough, you will get no or inaccurate results. Use of \hyperref[cmdline:check]{\texttt{--check}} is advised.\\
	If \hyperref[cmdline:models]{\texttt{--models}} isn't provided, \prasp samples a heuristically determined  number of random models (which is not guaranteed to be optimal).\\
	
	Iteration stops if the maximum number of iterations \verb§n§ (default: 1000) is reached or if the euclidean distance between the probability distribution computed in the current step and the distribution obtained in the previous step falls below threshold \verb§epsilon§ (default: 0.02). \\
	
	To make this approach tractable for complex background knowledge, \texttt{--itrefinement} can be combined with \hyperref[cmdline:models]{\texttt{--models}} and \hyperref[cmdline:initsample]{\texttt{--initsample}} in order to reduce the number of models on which this approach operates  (but a smaller number of sampled models also decreases entropy). Actually, if these arguments are not provided, \prasp heuristically selects a sampling approach and a number of samples. If you want to enforce that iterative refinement should use the full set of models of the spanning program (for more accurate results, possible with small systems), provide arguments \hyperref[cmdline:initsample 2 --models 0]{\texttt{--initsample 2 --models 0}}\\
	
	Iterative refinement starts by default with the uniform distribution over all models generated by the initial sampling step (discarding any duplicates if the initial samples are a multiset). Alternatively (if \verb§retainCounts§ is \verb§true§), it also uses the frequencies of models generated by initial sampling as a starting point. The initial samples are then treated as a multiset where the normalized count of a certain model within the multiset is interpreted as the initial approximation of the probability of that model. However, there are no entropy guarantees anymore in that case.\\
	
	The algorithm is presented in Section \ref{itrefinement}.\\ 
	
	Providing switch \hyperref[cmdline:maxentropy]{\texttt{--maxentropy}} has no direct influence on the \hyperref[cmdline:itrefinement]{\texttt{--itrefinement}} algorithm (which intrinsically maximizes entropy, given a certain number of samples). However, \hyperref[cmdline:maxentropy]{\texttt{--maxentropy}} might influence the number of initial samples computed with \hyperref[cmdline:initsample]{\texttt{--initsample}} and might thus indirectly also influence the entropy of the solution distribution.\\
	
	If specified, iterative refinement replaces the linear system solving approach in the inference pipeline, so \hyperref[cmdline:nosolve]{\texttt{--nosolve}} should normally \underline{not} be specified together with \hyperref[cmdline:itrefinement]{\texttt{--itrefinement}} (it is possible to use \texttt{--itrefinement} with \hyperref[cmdline:nosolve]{\texttt{--nosolve}}, but it is unclear yet if this combination provides any benefits). \\
	
	A major benefit of this approach is that the entropy of the computed solution depends on the 
	number of initial samples, i.e., you can increase the entropy by providing a larger number
	of sampled models or by changing the initial sampling approach. But of course, generating a larger, uniformly distributed number of models increases computation time too. Maximum entropy can be achieved by using \textit{all} models of the spanning program using initial sampling approach 2 (which obtains \textit{all} answer sets of the spanning program), but this is of course intractable for all but the most simple background knowledge programs.\\ 
	
	Notice that convergence of iterative refinement (in the way it is used in \prasp) does not guarantee that the query results reflect your constraints (formula weights, rules, etc). If the list of models obtained using initial sampling is too small or biased, query results are wrong regardless of \verb§epsilon§.\\
	It is thus recommended to use \hyperref[cmdline:check]{\texttt{--check}} together with \hyperref[cmdline:itrefinement]{\texttt{--itrefinement}}.\\
	
	\hyperref[cmdline:itrefinement]{\texttt{--itrefinement}} can theoretically be combined with \hyperref[cmdline:simanneal]{\texttt{--simanneal}} as a sampling approach (in that case, \hyperref[cmdline:nosolve]{\texttt{--nosolve}} is also omitted although \hyperref[cmdline:simanneal]{\texttt{--simanneal}} is given), but this combination has no known benefits.\\
	
	\hyperref[cmdline:itrefinement]{\texttt{--itrefinement}} currently doesn't work with \hyperref[cmdline:intervalresults]{\texttt{--intervalresults}} (but intervals can be specified in background knowledge).\\
	
	Independence declarations and discovered independence are not enforced by the iterative refinement algorithm (but the fact that the algorithm maximizes entropy compensates for this to some degree). However, independence declarations or discovered independence might still be considered during initial sampling, depending on the initial sampling approach (see \hyperref[cmdline:initsample]{\texttt{--initsample}}).
	
	\addcontentsline{toc}{subsection}{\texttt{--itrefinement}}  \label{cmdline:itrefinement}
	
	\subsection*{\normalsize{\texttt{--ascheckmode m}}} specifies the approach to be used for answer set checking. 
	\prasp performs \textit{Answer Set Checking} (ASC, a form of model checking) at various points. ASC determines whether
	$m \models p$ holds for a model $m$ and some ASP program $p$. When \prasp checks a model for a single formula $f$ in the background knowledge, $p = \dmh(\kb) \cup f$ where  $\dmh(\kb)$ is the spanning program of $\kb$. We sometimes write $\models_{\kb}$ instead of $\models \dmh(\kb) \cup ...$.\\
	
	When \prasp checks a model for a single query formula $q$,  $p =  \dmh(\kb) \cup uQ \cup q$ where $uQ$ is the program consisting of all unannotated formulas in the respective query file.\\
	
	This switch specifies the approach being used if $p$ is not a so-called ``simple formula'' \ref{simpleFormulas} in the context of the spanning program.\\
	
	With m = -1 (default), the approach is automatically chosen using a heuristics depending from the number of computed models of the spanning programming.\\
	
	m = 0:  $m \models^C p$ \textit{iff} $p \cup \mathtt{n\{}m'_1,...,m'_n\mathtt{\}n}$ satisfiable for $m' = m \cup \{ \mathtt{not\ } sr_i, sr_i \in \mathit{Atoms}(p)\setminus m \}$ (checks consistency with model).\\
	
	m = 1: $m \models p$ \textit{iff} $m \in \mathit{AS}(p)$ where $\mathit{AS}(p)$ is the set of all answer sets of $p$.\\
	
	m = 2: like m = 1, but only a uniformly sampled subset of $n$ answer sets of $p$ is used where $n$ is the number of models specified using \hyperref[cmdline:models]{\texttt{--models}} (generally unsound).\\
	
	For small systems, m = 1 is typically fastest but intractable for larger systems for which m = 0 works better.
	
	\addcontentsline{toc}{subsection}{\texttt{--ascheckmode}}  \label{cmdline:ascheckmode}
	
	\subsection*{\normalsize{\texttt{--mod0}}} (\textit{experimental!}) simplifies certain ASP ground rules whose bodies consist entirely of literals which are declared probabilistically mutually unconditionally independent. \hyperref[cmdline:mod0]{\texttt{--mod0}} requires a special design of the background knowledge to be useful in practice, otherwise the time required for the extra analysis caused by \hyperref[cmdline:mod0]{\texttt{--mod0}} exceeds any performance gain. Requires that weighted formulas with single bracket weights (i.e., of form \verb§[w] f§) are ground. Typically used in connection with \hyperref[cmdline:mod1]{\texttt{--mod1}}. See \ref{mod0} for details.\\
	
	Note that \texttt{--mod0} currently considers mutual independence only. Furthermore, in contrast to default setting, \texttt{--mod0} requires formulas to be fully parsed locally (i.e., by \prasp itself, not just by the ASP grounder), which currently induces limitations wrt. Gringo4 formula syntax supported with \texttt{--mod0}.\\
	
	\addcontentsline{toc}{subsection}{\texttt{--mod0}}  \label{cmdline:mod0}
	
	\subsection*{\normalsize{\texttt{--mod1}}} (\textit{experimental!}) simplifies background knowledge with the aim to remove all parts which have no influence on the result of queries. This filtering step might make a normally intractable task solvable (even with precise inference approaches). The aim of this feature is remotely similar to the identification of the minimal Markov network required for answering a query in Markov Logic \cite{mln}.\\
	
	See \ref{mod1} for details. Requires that weighted formulas with single bracket weights (i.e., of form \verb§[w] f§) are ground.\\
	
	\verb§--mod1§ only has a positive effect if the gain achieved by the simplification is bigger than its cost (time required for analyzing and simplifying the spanning program). \\
	
	\hyperref[cmdline:mod1]{\texttt{--mod1}} does not comprise \hyperref[cmdline:mod0]{\texttt{--mod0}}, both switches are independent from each other and can be used together or alone.\\
	
	Note that in contrast to default setting, \texttt{--mod1} requires formulas to be fully parsed locally (i.e., by \prasp itself, not just by the ASP grounder), which currently induces limitations wrt. Gringo4 syntax supported with \texttt{--mod1}.\\
	
	\addcontentsline{toc}{subsection}{\texttt{--mod1}}  \label{cmdline:mod1}
	
	\subsection*{\normalsize{\texttt{--noautoindeps}}} disables the automated identification of mutually independent atoms. Manually declared independent formulas are not affected by this switch. Use this switch in case you know which formulas are independent and you have declared these yourself independent in the background knowledge (using meta-statements such as \verb§#indep§, etc), or in case the automated identification includes atoms you wish to treat as non-independent (use \hyperref[cmdline:showindeps]{\texttt{--showindeps}} to see which atoms \prasp considers to be independent),
	or in order to increase processing speed in case the constraints introduced in order to enforce
	independence among these atoms slow down inference (but see \hyperref[cmdline:mod0]{\texttt{--mod0}}). For more details about independence handling see \ref{indep} (meta-statement \texttt{\#indep}) and \ref{performance}.
	
	\addcontentsline{toc}{subsection}{\texttt{--noautoindeps}}  \label{cmdline:noautoindeps}
	
	\subsection*{\normalsize{\texttt{--ignoredeclindeps}}} ignores all declared independence (meta-statements such as \verb§#indep§). Observe that even with this switch, certain sampling methods might make implicit independence assumptions (e.g., \verb§--initsample 4§).\\
	
	Also consider \hyperref[cmdline:noindepconstrs]{\texttt{--noindepconstrs}} and \hyperref[cmdline:noautoindeps]{\texttt{--noautoindeps}}.
	
	\addcontentsline{toc}{subsection}{\texttt{--ignoredeclindeps}}  \label{cmdline:ignoredeclindeps}
	
	%\textbf{\texttt{--extiidanalysis| (\textit{internal use only}) activates an extended (bolder) analysis for the discovery of mutually independent atoms.\\ % still works, but very dubious
	
	\subsection*{\normalsize{\texttt{--noindepconstrs}}} specifies that \prasp should not generate probability constraints for mutually or pairwise independent atoms (automatically discovered as well as manually declared ones). If this switch is not present, and event independence has been declared or discovered, inference algorithms observe constraints which enforce that $Pr(a \wedge b) = Pr(a)Pr(b)$ for any two independent events $a$, $b$. The algorithms affected by this are currently linear system solving, linear programming and simulated annealing. Any other use of formula independence (e.g., with \hyperref[cmdline:mod0]{\texttt{--mod0}}) is \textit{not} affected by this switch.\\
	It is important to not confuse independence \textit{enforcement} with the independence \textit{assumption} made by some (but not all) algorithms in \prasp; \hyperref[cmdline:noindepconstrs]{\texttt{--noindepconstrs}} only disables the enforcement of independence relations. For further details about independence handling see \ref{indep} (meta-statement \texttt{\#indep}) and \ref{performance}.\\
	
	\texttt{--noindepconstrs} does not imply/is not implied by \hyperref[cmdline:noautoindeps]{\texttt{--noautoindeps}} or \hyperref[cmdline:ignoredeclindeps]{\texttt{--ignoredeclindeps}}.
	
	\addcontentsline{toc}{subsection}{\texttt{--noindepconstrs}}  \label{cmdline:noindepconstrs}
	
	\subsection*{\normalsize{\texttt{--limitindepcombs n}}} limits computation of combinations $f_1, ..., f_n$ of mutually independent formulas to length \verb§n§$ \geq 2$. Leads to wrong results for $Pr(f_1 \wedge ... \wedge f_m)$ if results depend on the assumption that $f_1, ..., f_m$ are independent for any $m>n$.
	
	\addcontentsline{toc}{subsection}{\texttt{--limitindepcombs}}  \label{cmdline:limitindepcombs}
	
	\subsection*{\normalsize{\texttt{--check}}} makes \prasp print information useful for estimating the accuracy of inference results. Concretely, it shows the differences between the given weights of formulas in background knowledge and the inferred probabilities for these formulas. Constraints imposed on probabilities due to independence declarations and initial sample entropy are not checked (but see \hyperref[cmdline:showentropy]{\texttt{--showentropy}} and \hyperref[cmdline:pwdistr]{\texttt{--pwdistr}}).\\
	
	A different kind of check is provided using \hyperref[cmdline:checkconsistency]{\texttt{--checkconsistency}}
	
	\addcontentsline{toc}{subsection}{\texttt{--check}}  \label{cmdline:check} 
	
	\subsection*{\normalsize{\texttt{--checkconsistency}}} lets PrASP check the logical and probabilistic consistency of the specified background knowledge. Note that this slows down inference. A different kind of check is provided using \hyperref[cmdline:check]{\texttt{--check}}
	
	\addcontentsline{toc}{subsection}{\texttt{--checkconsistency}}  \label{cmdline:checkconsistency}
	
	\subsection*{\normalsize{\texttt{--showindeps}}} shows the set of formulas which \prasp considers to be logically independent of each other.
	
	\addcontentsline{toc}{subsection}{\texttt{--showindeps}}  \label{cmdline:showindeps}
	
	\subsection*{\normalsize{\texttt{--cacheinfsetup}}} enables a cache for inference setups. Significant speed gains if inference and learning tasks comprise repetition (in particular with streams which contain repetitions of the same floating windows), but might make inference and learning slower in other cases.
	
	\addcontentsline{toc}{subsection}{\texttt{--cacheinfsetup}}  \label{cmdline:cacheinfsetup}
	
	\subsection*{\normalsize{\texttt{--unisample m}}} specifies the approach used for sampling from a (near-)uniform distribution of models. Choosing a sampling approach which is appropriate for the problem at hand can make a dramatic difference wrt. inference and learning speed.\\
	
	\hyperref[cmdline:unisample]{\texttt{--unisample}} affects almost all places where uniform sampling is performed, including where uniform sampling is a helper task for weighted sampling. This affects most of the initial sampling approaches but also sampling which is part of inference approaches like simulated annealing. An exception are those places where flip-sampling needs itself to call a uniform-sampling approach (see \hyperref[cmdline:flipsampconf]{\texttt{--flipsampconf}}). 
	
	\verb§m§ = -1: auto (let \prasp decide, use \hyperref[cmdline:verbose]{\texttt{--verbose}} to see selected approach), 0: random models from full set of models (high quality but can be very slow), 1: (uniform) flip-sampling (see Section \ref{flipsamp}; where applicable, which is where the sampled models are answer sets of the spanning program), 2: XOR-sampling (Algorithm \ref{alg:xor}), 3: ASP solver-internal randomization (if applicable; see \hyperref[cmdline:sirndconf]{\texttt{--sirndconf}} for details), 4: like 1, but only formulas which are known to be independent are flipped. \\ 
	These methods can be further configured using \hyperref[cmdline:xorconf]{\texttt{--xorconf}}, \hyperref[cmdline:flipsampconf]{\texttt{--flipsampconf}} and \hyperref[cmdline:sirndconf]{\texttt{--sirndconf}}.\\
	
	If there are dependencies between the events in your program, try XOR-sampling first. If you know that all events are mutually independent, try flip-sampling first. Method 3 is typically fastest, but of low sampling quality (large bias).\\
	
	Uniform flip-sampling (which can be seen as a meta-sampling approach) itself needs to obtain uniformly sampled models from a randomly modified variant of the spanning program, and it does so using some other approach to sampling, namely the approach specified using \hyperref[cmdline:flipsampconf]{\texttt{--flipsampconf}} (or a default method). Again, this choice influences sampling quality.\\  
	
	Observe that \textbf{\texttt{--unisample}} does not ``activate'' (near-)uniform sampling, it merely specifies the approach used for (near-)uniform sampling.
	
	\addcontentsline{toc}{subsection}{\texttt{--unisample}}  \label{cmdline:unisample}
	
	\subsection*{\normalsize{\texttt{--xorconf q1 q2}}} specifies the quality of near-uniform sampling using XOR streamlining constrains \cite{xor}. This sampling method is underlying most sampling tasks in \prasp, including some of the initial sampling approaches and simulated annealing.\\
	\verb§q1§ is the number of models requested with each ASP solver call (default: 100). The larger \verb§q1§, the better the sampling quality (uniformity) and the slower the computation of samples. With \verb§q1§ = 0, the samples requested will
	be chosen randomly from \textit{all} models computed using XOR constraints (i.e., the random subset of all models of the respective XOR-constrained ASP program. Best quality, slowest).\\
	\verb§q2§ is the number of XOR constraints generated per sampling step (if \verb§q2§ is omitted, a heuristics is being used). See Algo. \ref*{alg:xor} for details.\\
	
	Note that this command-line argument does not ``activate'' XOR-sampling, it just configures it.\\
	
	Sampling using XOR streamlining constraints is formally described as Algorithm \ref{alg:xor}.
	
	\addcontentsline{toc}{subsection}{\texttt{--xorconf}}  \label{cmdline:xorconf}
	
	\subsection*{\normalsize{\texttt{--sirndconf o}}} configures sampling without use of XOR streamlining constraints, using ASP solver-internal randomization instead. This is faster but often obtains significantly lower degree of uniformity compared with XOR-sampling, i.e., less accurate results or slower convergence rate if combined with sampling-based inference such as simulated annealing.\\
	\hyperref[cmdline:sirndconf]{\texttt{--sirndconf}} affects all sampling tasks in \prasp where solver-internal randomization can be used, including initial sampling with \hyperref[cmdline:initsample]{\texttt{--initsample}}.\\
	$o$ is an optional multiplier which influences the size of the set of models from which samples are chosen randomly. The size of each set is $o \cdot n$ where
	$n$ is the requested number of samples. The higher $o$, the better is the sample quality, except for the following: With $o=0$, and if the respective sampling task is without replacement, samples are selected randomly from a uniform random distribution of \textit{all} models (very slow for large systems). Default is 100.\\
	
	Note that this command-line argument does not ``activate'' solver-internal randomization, it just provides an opportunity to configure it when solver-internal randomization is selected with, e.g., \hyperref[cmdline:initsample]{\texttt{--initsample}} or \hyperref[cmdline:flipsampconf]{\texttt{--flipsampconf}}.\\
	
	Alternative ASP solver parameters for randomization can be specified using \hyperref[cmdline:groundersolver]{\texttt{--groundersolver}}.
	
	\addcontentsline{toc}{subsection}{\texttt{--sirndconf}}  \label{cmdline:sirndconf}
	
	\subsection*{\normalsize{\texttt{--flipsampconf r}}} configures flip-sampling (both uniform and weighted). If \verb§r§ is \verb§0§ (default), each time the flip-sampling algorithm requests a model sampled from the uniform distribution, it uses the first model generated by the solver with solver-internal randomization activated. This is much faster but less ``uniform'' than the other available approaches here, where the uniformly-sampled model is either obtained randomly from all models of the spanning program with flipped formulas (r = 1) or using XOR-sampling (r = 2). If \verb§r§ is 0, solver-internal randomization quality can be further configured using \hyperref[cmdline:sirndconf]{\texttt{--sirndconf}} or \hyperref[cmdline:xorconf]{\texttt{--xorconf}}. These three approaches can also be used directly (without flip-sampling).\\
	
	Note that \hyperref[cmdline:flipsampconf]{\texttt{--flipsampconf}} does not ``activate'' flip-sampling, it just allows to specify its parameters.\\
	
	Flip-sampling is formally described in Section \ref{flipsamp}.
	
	\addcontentsline{toc}{subsection}{\texttt{--flipsampconf}}  \label{cmdline:flipsampconf}
	
	\subsection*{\normalsize{\texttt{--noremotesampling}}} enforces that always the older (pre v0.8) approach to invoke the ASP solver for sampling is being used even where the more recent Clingo4-based \textit{remote sampling} would be possible.\\
	
	Remote sampling means that Clingo 4 is used with Lua-script \verb§clingo4Sampling.lp§ for certain sampling approaches. Remote sampling is typically much faster than the older approach. Unless \texttt{--noremotesampling} is specified, \prasp automatically chooses in which situations \verb§clingo4Sampling.lp§ is used. At this, \prasp maintains a standing connection to the external ASP solver, with communication with \prasp via TCP/IP. Sampling via (stratified or unstratified) weighted or unweighted flip-sampling algorithms is performed on client side, which makes sampling massively faster. Furthermore, weakly uniform sampling using solver-internal randomization is supported.  
	
	\verb§clingo4Sampling.lp§ is a Lua-script, so if you like you can provide your own sampling approach ``on-the-fly'' without having to change \prasp itself.\\
	The algorithm for remote flip-sampling currently implemented by the Lua script is, for technical reasons, somewhat different from the \prasp-internal one, see Algo. \ref{alg:flipsamp}.
	
	There are currently a few restrictions in connection with remote sampling:\\
	
	- flip-sampling using \verb§clingo4Sampling.lp§ requires that only atoms have weights attached (or rules where \prasp is able to simplify the body away during grounding if annotated with \verb§[[...]]/[[[...]]]§). This is not a serious restriction for most use cases, since Annotated Disjunctions can still be used (they are de-sugared into unweighted rules and weighted atoms).\\
	
	- The combined grounder-solver needs to be Clingo 4 (version 4.5.3 or higher). Other ASP grounders/solvers are not yet supported with remote sampling.
	
	\addcontentsline{toc}{subsection}{\texttt{--noremotesampling}}  \label{cmdline:noremotesampling}
	
	%\textbf{\texttt{--sampleformulamodels}} (experimental) makes PrASP sample from a near-uniform distribution if the models of a formula are required. Otherwise (default), \prasp compute \textit{all} models of a certain formula in its context (the spanning program) if at some point during inference models of that formula are needed.  \\
	
	\subsection*{\normalsize{\texttt{--nosolve}}} omits possible world weight computation by linear equation system solving or linear programming (optimization) when doing inference. Query probabilities are determined by direct counting of models. \hyperref[cmdline:nosolve]{\texttt{--nosolve}} requires that at least \textit{one} of the following prerequisites holds in order to obtain approximately correct results:
	\begin{itemize}
		\item An initial sampling approach which immediately provides a valid possible worlds distribution.
		This is doable if all weighted formulas are mutually independent. See \hyperref[cmdline:initsample]{\texttt{--initsample}} for details.
		%\item Switch \hyperref[cmdline:itrefinement]{\texttt{--itrefinement}}
		\item Switch \hyperref[cmdline:simanneal]{\texttt{--simanneal}} is provided
		\item Switch \hyperref[cmdline:weights2cc]{\texttt{--weights2cc}} is given together with background knowledge where weighted formulas represent mutually independent events. 
		\item Unweighted background knowledge which is manually designed such that the models (obtained using initial sampling) of its spanning program
		represent the desired probability distribution (not always possible). Weights are entirely ignored here.
	\end{itemize}
	
	\addcontentsline{toc}{subsection}{\texttt{--nosolve}}  \label{cmdline:nosolve}
	
	%DOESN'T WORK PROPERLY YET: \textbf{\texttt{--infapproxcount numberofsamples}} performs inference by approximate model counting. If \hyperref[cmdline:weights2cc]{\texttt{--weights2cc}} is not given, weighted model counting will be used. Requires \hyperref[cmdline:nosolve]{\texttt{--nosolve}} and \hyperref[cmdline:initsample 0]{\texttt{--initsample 0}}.
	%\verb§numberofsamples§ (optional) is the number of samples used in each step (higher = more accurate result). If omitted, \prasp determines this value heuristically.\\
	
	\subsection*{\normalsize{\texttt{--mlns}}} 
	(experimental!) emulates Markov Logic Network \cite{mln} semantics - to some degree in the style of \cite{mlnasp}, i.e. combined with stable model semantics.\\ 
	Weighted rules where the weight is below 1 are interpreted as soft rules whereas other rules become hard rules in the MLN sense (i.e., weight 1 corresponds to MLN's infinite weight). Weights in background knowledge do \textit{not} denote probabilities anymore, whereas weights resulting from queries are still probabilities.\\
	
	This switch needs to be combined with --\hyperref[cmdline:noindepconstrs]{\texttt{--noindepconstrs}}. Also, it doesn't support conditional probabilities in background knowledge (they are simply ignored). However, conditional probabilities can still be used in query files.\\
	 
	\noindent See Sect. \ref{inconsistency} for an example for how to use this switch.\\
	
	Also see \hyperref[cmdline:maxwalksat]{\texttt{--maxwalksat}}.
	
	\addcontentsline{toc}{subsection}{\texttt{--mlns}}  \label{cmdline:mlns}
	
	\subsection*{\normalsize{\texttt{--intervalresults}}} \index{Probability intervals} makes \prasp return for each query formula a \textit{range} of probabilities for this query (lower and upper bounds), i.e., query results become probability intervals. \\
	If no other inference approach is specified, this makes \prasp use a simplex solver (linear programming) to compute a probability interval for each query formula. The solver can be configured using \hyperref[cmdline:linoptimconf]{\texttt{--linoptimconf}}.\\
	Note: \texttt{--intervalresults} is not required for the specification of probability intervalsin background knowledge.\\
	See \ref{intervalresults} for further details about probability intervals.
	
	\addcontentsline{toc}{subsection}{\texttt{--intervalresults}}  \label{cmdline:intervalresults}
	
	\subsection*{\normalsize{\texttt{--linoptimconf epsilon fpcompULP cutoff maxit}}} configures the linear programming approach used by \prasp.
	\verb§epsilon§ specifies threshold for algorithm convergence (default: 0.001). \verb§fpcompULP§ (deprecated) is the Units in the Last Place for floating point comparison (default: 10), \verb§cutoff§ (deprecated) is the cut-off value for pivot value selection in the simplex method (elements smaller than \verb§cutoff§ will be treated as zeros; default: 1.0E-10), and \verb§maxit§ is the maximum number of iterations before the algorithm gives up (default: 100000).\\
	\verb§fpcompULP§ and \verb§cutoff§ are deprecated and their values are ignored with the current linear programming solver. 
	
	\addcontentsline{toc}{subsection}{\texttt{--linoptimconf}}  \label{cmdline:linoptimconf}
	
		\subsection*{\normalsize{\texttt{--linsolveconf approach regularizer}}} 
		
		With the ``vanilla'' approach to inference, \prasp computes the probabilities of possible worlds by calculating or approximating the solution of a system of linear equations $Ax=b$ (see Sect. \ref{linearSystem}). If there are no intervals in background knowledge and also no queries with intervals, the following approaches are currently provided. \\
			
		Note that while some of the approaches below (in particular native and CUDA solvers and iterative approaches to LS) are very fast, the true bottleneck with the ``vanilla'' approach to inference is not the system solver or LS method but the time required to compute a very large number of models (possible worlds). \\ 
		
		\prasp uses either (NN)LS (with SMT as fall-back) or linear programming. (NN)LS is configured using this switch whereas linear programming is configured with \hyperref[cmdline:linoptimconf]{\texttt{--linoptimconf}} and enforced using \hyperref[cmdline:intervalresults]{\texttt{--intervalresults}}. \\ 
		
		\verb§approach§ is a number which specifies the concrete Linear Least Squares method. By default (-1), \prasp uses for its ``vanilla'' inference approach the Biconjugate Gradient Stabilized Method with regularizer 0.001 (unless the system size is below 170x170, then SVD is used by default). Exceptions if \verb§--enforceSMT§ or \verb§--checkconsistency§ or \verb§--maxentropy§ are specified: with \verb§--checkconsistency§, Singular Value Decomposition (SVD) with one additional SVD call is enforced and with \verb§--maxentropy§, we use a single SVD with a subsequent gradient descent search for entropy maximizing distributions (search is not guaranteed to succeed - consider \verb§--itrefinement§ as a truly entropy maximizing alternative).
		
		\verb§regularizer§ (optional) is a number between 0 and 1 which specifies the regularization constant applied to increase the robustness of the solver. The larger this value, the more robust the solver becomes (more often finds a solution) and the less accurate are inference results.  
		
		\begin{description}
			\item[Non-Negative Linear Least Squares (NNLS)] (\verb§approach§ = 0) using Lawson and Hanson's method \cite{NNLS} (as with \verb§lsqnonneg§ in MATLAB) backed up with a QR-based non-constrained least-squares solver (native or on GPU, where available). While Lawson and Hanson is not the fastest approach to NNLS, it is quite robust and faster than the also robust L-BFGS-B. 	
			\item[Fast Non-Negative Linear Least-Square (FNNLS)] (\verb§approach§ = 1) using the algorithm proposed in \cite{FNNLS} (variant of Lawson and Hanson's method).   
			\item[Non-constrained Linear Least Squares (LS)] based on non-iterative or iterative approaches. Non-constrained LS is not a general inference approach in \prasp but \textit{if} it finds a solution it does so faster than the NNLS approaches. 
			\begin{description}
				\item[Non-iterative approaches] provided are SVD-decomposition, QR-decomposition or Cholesky-decomposition. Cholesky is fastest but least robust. SVD is typically most accurate but slowest and QR is on the middle-ground between the others. These approaches are typically faster than NNLS, however, they don't necessarily compute a probability distribution. If no valid solution is returned, \prasp attempts to use the invalid solution and the linear system's kernel basis (if available) as a starting point to find valid solutions, provided the system is underdetermined.\\
				\verb§approach§ = 10: QR decomposition using CUDA \\
				\verb§approach§ = 11: Cholesky decomposition using CUDA\\
				\verb§approach§ = 12: QR using LAPACK/BLAS\\
				\verb§approach§ = 13: SVD using LAPACK/BLAS\\
				\verb§approach§ = 14: QR using Apache Math\\
				\verb§approach§ = 15: SVD using Apache Math\\					
				For more precise results, it is advisable to use SVD (\verb§approach§ = 13 or 15).
						 
				\item[Iterative approaches] in \prasp are the Biconjugate Gradient Stabilized method (BiCGstab) and the Generalized Minimal Residual method (GMRES). The latter is particularly useful with sparse systems. Given that rules tend to be true in all possible worlds (and they always are if they are not weighted), we can often increase sparsity by flipping the coefficients of matrix A (which denote whether the respective row formula holds in the respective possible world) and negating the row formulas (i.e., replacing their weights with 1-weight). \\
				\verb§approach§ = 20: BiCGstab based on LAPACK/BLAS\\
				\verb§approach§ = 21: GMRES based on LAPACK/BLAS				  
			\end{description}

		\end{description}
		
		SVD non-constrained LLS is used with subsequent search for further solutions using the system's null space (kernel). This does not guarantee a solution but is quite fast if it finds one. Also, it allows to find a maximum entropy solution in case of an underdetermined system.\\
		
		Matrix decompositions can be computed, where available, using CUDA (i.e., on the NVIDIA graphics processor) or using an external native LAPACK library. Performance of CUDA naturally depends on the speed and memory of the Nvidia GPU in your machine.\\
		
		LAPACK/BLAS native linear algebra libraries need to be specified on the command line, using JVM \verb§-D§ parameters. If these are missing, Java implementations are used, which are naturally slower.\\
			Various alternative libraries can be specified here (commercial as well as free ones), with different performance characteristics. To enforce the use of a particular implementation, modify \verb§prasp.sh §or \verb§prasp.bat§. Example:
		\begin{verbatim}		
		-Dcom.github.fommil.netlib.BLAS=com.github.fommil.netlib.NativeRefBLAS
		-Dcom.github.fommil.netlib.LAPACK=com.github.fommil.netlib.NativeRefLAPACK
		-Dcom.github.fommil.netlib.ARPACK=com.github.fommil.netlib.NativeRefARPACK
		\end{verbatim}	

		To \textit{de}activate the use of native BLAS, LAPACK and ARPACK libraries, specify
		
		\begin{verbatim}
		-Dcom.github.fommil.netlib.BLAS=com.github.fommil.netlib.F2jBLAS
		-Dcom.github.fommil.netlib.LAPACK=com.github.fommil.netlib.F2jLAPACK
		-Dcom.github.fommil.netlib.ARPACK=com.github.fommil.netlib.F2jARPACK
		\end{verbatim}
		
		The (NN)LS approaches minimize the 2-norm $||Ax-b||_2$, so they return a probability distribution (and thus query results) even where the linear system does not have a solution (i.e., if your background knowledge is inconsistent).  \\
		
		For handling of intervals, see \hyperref[cmdline:intervalresults]{\texttt{--intervalresults}} and \ref{intervalresults}.
		The basic difference to the approaches listed above is that with intervals, \prasp uses linear programming as default approach. In contrast to (NN)LS, linear programming minimizes the 1-norm instead of the 2-norm of $Ax-b$, where the latter is otherwise preferable (as it is differentiable). Alternatively, \prasp can also use the SMT solver as fallback solution to interval handling (but this is much slower).  
					
		\addcontentsline{toc}{subsection}{\texttt{--linsolveconf}}  \label{cmdline:linsolveconf}
	
	\subsection*{\normalsize{\texttt{--ndistrs n}}} performs inference using \verb|n| probability distributions over possible worlds (\verb|n |$ \geq 1$). Default: \verb|n| = 1. \texttt{--ndistrs n} works only with the default inference method (linear system solving).\\
	Each query formula might then result in multiple alternative probabilities (not necessarily different from each other).
	For unconditional query formulas, these probabilities are separated by ``,' '  and listed in the order corresponding to the sequence
	of probability distributions which have been used. \\ 
	Consider using \hyperref[cmdline:intervalresults]{\texttt{--intervalresults}} instead of \hyperref[cmdline:ndistrs]{\texttt{--ndistrs}}. For details, see Sect. \ref{intervalresults}.
	
	\addcontentsline{toc}{subsection}{\texttt{--ndistrs}}  \label{cmdline:ndistrs}
	
	\subsection*{\normalsize{\texttt{--maxentropy p nsc st r}}} (\textit{experimental!}) activates heuristic options in inference and sampling algorithms so that the entropy of the computed probability distribution over possible worlds is increased compared to the default settings. How much the entropy can actually be increased depends on the chosen sampling and inference approaches. \texttt{--maxentropy} is currently effective on the default inference approach (linear system solving with point probability results) and - to a limited degree - for simulated annealing.\\
	\texttt{--maxentropy} also influence the sampling of initial models, see \hyperref[cmdline:models]{\texttt{--models}} and \hyperref[cmdline:initsample]{\texttt{--initsample}}. \hyperref[cmdline:maxentropy]{\texttt{--maxentropy}} has no direct influence on \hyperref[cmdline:itrefinement]{\texttt{--itrefinement}} but might indirectly influence the entropy of the iterative refinement result via the number of initial models. \\
	Note that a general limiting factor for entropy is the number of initial models set by \hyperref[cmdline:models]{\texttt{--models}} (more models lead to higher entropy).\\
	
	Optionally, and if linear system solving is used (\prasp ``vanilla''), you can specify the desired precision \verb§p§ wrt. entropy of the distribution search algorithm (lower value means higher precision and longer computation time). Default precision is $1\ensuremath{\times 10^{-4}}$.\\
	If parameter \verb§nsc§ is \verb§true§, the algorithm doesn't stop if the search direction changes (default: \verb§false§). \\
	If parameter \verb§st§ is \verb§true§, Stochastic Gradient Descent is used (vs Batch Gradient Descent). Default is \verb§false§. \\
	\verb§r§ is a value by which the algorithm replaces encountered zeros during the computation of the derivative of the Kullback$\textendash$Leibler divergence during Stochastic Gradient Descent. If you encounter ``sign change'' errors, try a smaller value than the default of 1e-6.\\
	
	These parameters are only available with linear system solving.\\
	
	If inference method is simulated annealing, p has a different meaning: it specifies the maximum acceptable coefficient of variation {\Large $\frac{\sigma}{\mu}$} (relative standard deviation) of the list of all entropies (of the candidate distributions computed by the simulated annealing algorithm). If the coefficient has not yet fallen below \verb§p§, the search for a distribution with higher entropy continues. Default value for p is here 0.15. \verb§nsc§ and \verb§st§ are ignored.\\
	
	For other inference methods, all parameters of \texttt{--maxentropy} are ignored (and it depends on the inference approach whether \texttt{--maxentropy} has any effect at all). For details, please see Sect. \ref{intervalresults}.
	
	\addcontentsline{toc}{subsection}{\texttt{--maxentropy}}  \label{cmdline:maxentropy}
	
	\subsection*{\normalsize{\texttt{--ignoreentropy}}} makes \prasp ignore the entropies of candidate probability distributions over possible worlds if inference is performed using linear system solving. Can be used to obtain faster results (which might carry an unwanted information bias though) or
	as an alternative to \hyperref[cmdline:intervalresults]{\texttt{--intervalresults}} in connection with \hyperref[cmdline:ndistrs]{\texttt{--ndistrs}} (here \hyperref[cmdline:ignoreentropy]{\texttt{--ignoreentropy}} computes a number of biased solution whose minimum and maximum values approximate the interval of all valid solutions).\\
	
	In the absence of both \hyperref[cmdline:maxentropy]{\texttt{--maxentropy}} and \hyperref[cmdline:ignoreentropy]{\texttt{--ignoreentropy}} in connection with linear system solving as inference method, \prasp computes a small number of sample distributions and chooses the one with the highest entropy. This obviously does not guarantee maximum entropy but merely aims at avoiding minimum entropy.\\
	
	Note that \hyperref[cmdline:ignoreentropy]{\texttt{--ignoreentropy}} does not mean that the resulting probability distribution has a low entropy, just that \PrASP does not apply any entropy-increasing measures.
	
	\addcontentsline{toc}{subsection}{\texttt{--ignoreentropy}}  \label{cmdline:ignoreentropy}
	
	\subsection*{\normalsize{\texttt{--enforceSMT}}} tells \prasp to always use the external SMT solver in place of the internal linear system solver.
	
	\addcontentsline{toc}{subsection}{\texttt{--enforceSMT}}  \label{cmdline:enforceSMT} 
	
	\subsection*{\normalsize{\texttt{--omitSMT}}} causes that the external SMT solver is only called in case the internal solver could not find at least one probability distribution. Otherwise, the SMT solver might be called even in case a distribution was already found by the internal solver, in order to generate additional possible worlds.
	
	\addcontentsline{toc}{subsection}{\texttt{--omitSMT}}  \label{cmdline:omitSMT}  
		
	\subsection*{\normalsize{\texttt{--weights2cc}}} (\textit{experimental!}) replaces the ordinary spanning program with a conversion of mutually independent formula weights into conditions with choice constructs. Experimental feature. Use in connection with \hyperref[cmdline:nosolve]{\texttt{--nosolve}} or \hyperref[cmdline:models]{\texttt{--models}}. 
	
	\addcontentsline{toc}{subsection}{\texttt{--weights2cc}}  \label{cmdline:weights2cc}
	
	\subsection*{\normalsize{\texttt{--strongnegbelief}}} (\textit{experimental!}) uses strong (classical) negation for modeling opposite beliefs $1-Pr(...)$ and creation of possible worlds from uncertain knowledge. Default is use of default negation (negation as failure). Also note that due to the way F2LP operates, the strong negation of complex formulas (vs. atoms) might be identical to the default negation.\\
	But generally, inferred probabilities of formulas $not\ f$ and $-f$ might differ from each other. If you want to ensure that they are identical, you can e.g. enforce a strong closed world assumption for some predicates by adding rules of the form \verb|-p :- not p|. However, note that classical negation is essentially just a syntactic feature of \textit{literals} in the input language of ASP grounders/solvers such as Gringo/Clingo. 
	
	\addcontentsline{toc}{subsection}{\texttt{--strongnegbelief}}  \label{cmdline:strongnegbelief}
	
	\subsection*{\normalsize{\texttt{--addnegf}}} generates additional constraints for negated weighted formulas. E.g., in addition to \verb§[0.3] f§, formula \verb§[0.7] not f§ is generated (respectively weighted literal \verb§[0.7] -f§ with \hyperref[cmdline:strongnegbelief]{\texttt{--strongnegbelief}}).
	
	\addcontentsline{toc}{subsection}{\texttt{--addnegf}}  \label{cmdline:addnegf}
	
	%\subsection*{\normalsize{\texttt{--searchcc}}} (with \verb|--weights2cc|)tries to find appropriate choice constructs by search (experimental; doesn't work in most cases) 
	
	%\subsection*{\normalsize{\texttt{--fullspan}}} replaces the ordinary spanning program with a program which generates all answer sets over \textit{all} literals found in background knowledge and queries. As a prerequisite, all variable domain declarations and
	%domain predicates required for query formulas need to be declared in the background knowledge (which is good practice also
	%in case \verb|--fullspan| is missing). Unannotated formulas in query files are ignored with \verb|--fullspan|.\\    
	
	\subsection*{\normalsize{\texttt{--spangenconf useFOL2ASP}}}
	allows to configure the spanning formula generation policy. If useFOL2ASP is \verb§true§ (default), spanning formulas are created from FOL forms of non-ground ASP formulas, where possible (provided a FOL$\rightarrow$ASP converter is specified). Otherwise, \prasp creates them without the intermediate FOL form (experimental).
	
	\addcontentsline{toc}{subsection}{\texttt{--spangenConf}}  \label{cmdline:spangenConf} 
	
	\subsection*{\normalsize{\texttt{--spanqueries}}} (\textit{experimental!}) adds spanning formulas of the form $q\ |\ not\ q$ for each query formula $q$. This implicitly
	creates a prior probability for each query formula (0.5 in case the formula is independent).
	
	\addcontentsline{toc}{subsection}{\texttt{--spanqueries}}  \label{cmdline:spanqueries}  
	
	\subsection*{\normalsize{\texttt{--dnf}}} (\textit{deprecated}) uses the Disjunctive Normal Form (DNF) of query and weighted formulas instead of their answer sets to check their truth in possible worlds. In older versions of \prasp, use of  \hyperref[cmdline:dnf]{\texttt{--dnf}} could significantly speed up inference if weighted and query formulas are very small, but with current \prasp versions, \hyperref[cmdline:dnf]{\texttt{--dnf}} is usually much slower than the default approach and other optimization means. Even more important: \hyperref[cmdline:dnf]{\texttt{--dnf}} doesn't observe stable model semantics and isn't fully compatible with ASP syntax and semantics (e.g., no nested terms allowed).\\
	\hyperref[cmdline:dnf]{\texttt{--dnf}} can be used with basic FOL syntax constructs, just observe that the internal translation of FOL formulas adds computation time, of course. This switch might disappear in future versions of \prasp.
	
	\addcontentsline{toc}{subsection}{\texttt{--dnf}}  \label{cmdline:dnf}
	
	%\subsection*{\normalsize{\texttt{--simpletc}}} uses a simplified test for checking if a formula is true in a possible world (answer set subset test; experimental)  
	
	\subsection*{\normalsize{\texttt{--assumegroundersolver c}}}
	forces PrASP to make an assumption (without checking) about the presence and version of the grounder/solver being used. If c = 0, it is assumed that Gringo/Clingo/Lparse is not used. If c = 1, a Gringo 3/Lparse/Clingo 3 grounder/solver is assumed, and with c = 2, Gringo/Clingo $\geq$4 is assumed. Default = -1 (no assumption, automated detection is used).
	
	\addcontentsline{toc}{subsection}{\texttt{--assumegroundersolver}}  \label{cmdline:assumegroundersolver}
	
	\subsection*{\normalsize{\texttt{--grounder filename \qq arguments\qq\  \qq plain text arguments\qq}}} specifies the external  ASP grounder (if omitted, a default external grounder binary will be used, which is currently \verb|gringo| (Linux), \verb|gringo.exe| (Windows) or \verb|gringo_macosx| (Mac)).\\
	\verb§"plain text arguments"§ are those arguments which are used when the grounder is asked for plain text grounding output (e.g., \verb§"-t"§ if the grounder is Gringo or Lparse).\\
	
	\noindent Example: \hyperref[cmdline:grounder]{\texttt{--grounder}}\verb§ "gringo4_macosx" "" "--text"§\\
	
	The grounder must match the combined grounder/solver being used. E.g., if you specify Gringo 3 as grounder, you must specify Clingo 3 as combined grounder/solver (using argument \hyperref[cmdline:groundersolver]{\texttt{--groundersolver}}).\\
	
	Observe that \prasp does not (and cannot) check what the specified program is actually doing, so this switch should be used with care.
	
	\addcontentsline{toc}{subsection}{\texttt{--grounder}}  \label{cmdline:grounder}
	
	\subsection*{\normalsize{\texttt{--groundersolver filename \qq arguments\qq\  \qq randomization arguments\qq}}} specifies the external combined ASP grounder+solver. If omitted, a default external grounder+solver binary will be invoked, which is currently \verb|clingo| (Linux), \verb|clingo.exe| (Windows) or \verb|clingo_macosx| (Mac). \\
	\verb§"arguments"§ is the list (enclosed in \verb§"§...\verb§"§) of arguments passed to the program, \verb§"randomization arguments"§ (enclosed in \verb§"§...\verb§"§) are
	those arguments which are passed to the program if the program should randomize the returned set of models (solver-internal randomization, see \hyperref[cmdline:sirndconf]{\texttt{--sirndconf}}).
	In \verb§"arguments"§, any occurrence of \verb§%NUM§ is replaced with the number of requested models, or 0 if all models are requested. Any occurrence of \verb§%NZNUM§ is replaced with the number of requested models, or 2147483647 if all models are requested.
	In \verb§"randomization arguments"§, any occurrence of \verb§%SEED§ is replaced with a random number.\\
	
	\noindent Default arguments (subject to change in later versions):\\
	
	With Clingo 4: {\small \verb§-n%NUM --verbose=1 --project --save-p=70 --config=jumpy§}\\
		With Clingo 3: {\small \verb§--shift -n%NUM --verbose=1 --project --save-p=70§} \\
			
			\noindent Default randomization arguments (subject to change in later versions):\\
			
			With Clingo 4: {\footnotesize \verb§--project --sign-def=3 --rand-freq=0.9 --restart-on-model --seed %SEED -n%NUM --verbose=1§}\\
				With Clingo 3: {\footnotesize \verb§--shift --project --rand-freq=0.9 --restart-on-model --seed %SEED -n%NUM --verbose=1§}\\
					
					\noindent Example:
					
					{\footnotesize \hyperref[cmdline:groundersolver]{\texttt{--groundersolver}}\verb§ "clingo4_macosx"§\\
						\verb§"-n%NUM --verbose=1 --project --save-p=70 --config=jumpy"§\\
						\verb§"--project --sign-def=3 --rand-freq=0.9 --restart-on-model --seed %SEED -n%NUM --verbose=1"§}\\
						
						\noindent The specified grounder must of course be compatible with the combined grounder/solver being used. E.g., if you specify Gringo 3 as grounder (using argument \hyperref[cmdline:grounder]{\texttt{--grounder}}), you must specify Clingo 3 as combined grounder/solver.\\
						
							Observe that \prasp does not (and cannot) check what the specified program is actually doing, so this switch should be used with care.
						
						\addcontentsline{toc}{subsection}{\texttt{--groundersolver}}  \label{cmdline:groundersolver}
						
	\subsection*{\normalsize{\texttt{--groundingconf groundPreContext allowInternalGrounding sortDomains revertDomains omitBgkInQueryContext genSpanInContext "parameters"}}} 
	
	influences how formulas are being grounded, in particular those annotated with \verb§[[...]]§ or \verb§[[[...]]]§.\\
	
	Please see Sect \ref{grounding} and Sect. \ref{weights} for background information.\\
	
	None of its arguments are optional, you need to specify all seven arguments here.\\		
	Default values correspond to \verb§--groundingconf 1 true true -1 false false ""§\\
	
	If \verb§groundPreContext§ is 1 (default) then \prasp calls the external grounder in order to obtain the set of ground domain predicate instances (which is then used to internally determine the domains of ASP variables) from the \textit{grounding precontext} (basically all unannotated formulas). If it is 0, only ground facts which are explicitly provided are considered as domain predicate instances and variable instantiation during internal grounding. This severely restricts the grounding process but avoids an external grounder call. Even with \verb§groundContext§=0, \prasp is still able to use simple intervals such as \verb§p(1..3,5..9)§ and sequences such as \verb§p(a;b;c)§ to obtain ground domain predicates, however, to obtain the grounding precontext (Sect. \ref{weights}) from more complex such formulas (e.g., with variables as interval boundaries), it needs an external grounder (this restriction applies to the step of grounding the grounding precontext in order to obtain ground domain predicate facts).\\
		
	If \verb§allowInternalGrounding§ is \verb§true§ (default), \prasp grounds basic \verb§[[..]]§- or \verb§[[[..]]]§-annotated ASP formulas itself without the help of an external grounder. This is not necessarily faster (there is no expensive call of an external grounder tool, but on the other hand the grounding method is likely slower than that of the external grounder), but allows \prasp to fix the order of generated formula ground instances (which is otherwise undetermined).\\
	Note that internal grounding still makes use of the external grounder (if present), namely in order to expand domain predicates in the unannotated part of the program.\\
	
	If \verb§sortDomains§ is \verb§true§ (default), \prasp sorts the resulting ground instances lexicographically and token-wise (where the tokens are the words and numbers in a formula).\\
	
	 With \verb§revertDomains§ = 1, the order of the sequences of instances of domain predicates is reversed. With \verb§revertDomains§ = 0, the original sequence is kept, and with \verb§revertDomains§ = -1 (default), reversion is determined automatically. Note that there is no way to ensure that  \verb§revertDomains§ determines the order of generated formula ground instances if an external grounder is being used.\\	 
	 
	 If \verb§omitBgkInQueryContext§ = \verb§true§ (default: \verb§false§), the background knowledge's grounding precontext is not added to the queries' grounding precontext. This is particularly useful if the unannotated part of the background knowledge is large and the query file contains all information needed to ground \verb§[[..]]§- or \verb§[[[..]]]§-annotated formulas in the query file. \\
	 
	 If \verb§genSpanInContext§ is \verb§true§ (experimental; default: \verb§false§), for each weighted formula in background knowledge, its spanning formula is added to the grounding precontext, whereas normally the grounding precontext only comprises unannotated formulas given directly in the file. \verb§genSpanInContext§ can help  grounding \verb§[[...]]§ or \verb§[[[...]]]§ annotated formulas whose grounding depends on other weighted formulas.\\ 
	 	
 	\verb§parameters§ (enclosed in \verb§"..."§) is a string with extra parameters (separated by whitespace) passed to the grounder. Default is \verb§""§. With Gringo 4, you could provide \verb§--keep-facts§ here in order to protect atoms in normal rules from being simplified away. However, even with this, the external grounder might perform certain simplifications which cannot be deactivated (e.g., of aggregates).\\
 	
 	Another means to influence grounding is meta-statement\\
 	\texttt{\#dontExternalize}\texttt{...\#endDontExternalize}, see Sect. \ref{meta}.
	
	%*\texttt{--groundingconf} should be used with care, as with inappropriate arguments, the grounder can easily simplify away ground formulas whose truth values are uncertain at the point of grounding, since the grounder performs simplification only against the given grounding precontext (unannotated formulas). *%
	If you are not sure about the result produced with the various options, you can use \hyperref[cmdline:showexpansion]{\texttt{--showexpansion}} to check if the resulting sequence of ground instance is what you had in mind.
	
	\addcontentsline{toc}{subsection}{\texttt{--groundingconf}}  \label{cmdline:groundingconf}
	
	\subsection*{\normalsize{\texttt{--folconv convertername}}} specifies the optional FOL$\rightarrow$ASP converter. This can be either the built-in converter FOL2ASP (specified using \verb§--folconv internal§) (this converter is also used if \verb§--folconv§ is missing) or some external converter, such as F2LP \cite{f2lp} (in that case, specify \verb§--folconv f2lp§ (Linux), \verb§--folconv f2lp.exe§ (Windows) or \verb§--folconv f2lp_macosx§ (Mac).\\
	Use \hyperref[cmdline:folconv]{\texttt{--folconv none}} to deactivate FOL$\rightarrow$ASP conversion (speeds up \prasp).\\
	Note that if \hyperref[cmdline:folconv]{\texttt{--folconv none}} is not specified and a converter is found, \prasp 
	might call the converter even if the background knowledge and queries are in ASP syntax.\\
	
		Observe that \prasp does not (and cannot) check what the specified program is actually doing, so this switch should be used with care.\\
		
		The built-in converter FOL2ASP uses almost the same algorithm \cite{f2lp} as F2LP, however, there are a number of technical differences:
		
		\begin{itemize}

		\item FOL2ASP is still somewhat less mature than F2LP
		\item FOL2ASP provides better compatibility with Gringo $\geq$4 than F2LP 1.3: Gringo3/Lparse \verb§#domain§ statements are still supported but there is also
		support for the new syntax \verb§![X,Y,Z,dx(X),dy(Y),dz(Z)]:§ (analogously for \verb§?[...]:§) where predicates \verb§dx§, \verb§dy§, \verb§dz§ specify the
		 domains of the variables. Equivalent to \verb§![X,Y,Z]: ... <- ..., dx(X), dy(Y), dz(Z)§, however, we automatically use these domains
		 for newly introduced helper variables too. Only one binding (atom) per variable allowed.
		  Also, use of this syntax might lead to another variable renaming (introduction of a new variable), to make sure that domain predicate
		 bindings are unique for their respective variables (e.g., consider \verb§![X]: p(X) & ?[X]: q(X) -> z§).
		\item   \verb§#domain§ meta-statements (deprecated, Gringo3/Lparse only) are collected and the respective atoms are added to subsequent quantifiers within square brackets.
		\item FOL2ASP doesn't generate any temporary files
		\item FOL2ASP ``learns'' using memoization of past results 
		\item Scope of quantifiers extends to the right as far as possible, i.e., is above that of rules and conditionals (\verb§<-§, \verb§->§, \verb§<->§).
		E.g., \verb§![X]: a(X) -> b(X)§ (FOL2ASP) corresponds to \verb§![X]: (a(X) -> b(X))§ (F2LP).
		To improve readability, or if in doubt about precedence, it is recommended to use redundant brackets.
		\item "\verb§not§" always denotes default negation, "\verb§-§" always stands for classic negation (allowed in front of atoms only, as usual in ASP)
		\item Instead of \verb§{not a}0§, FOL2ASP generates \verb§not not a§ if used with Gringo 4.
		\item Formula size only limited by memory
		\item Currently infix term operators such as \verb§<§, \verb§>§, \verb§=§ are not allowed in FOL formulas (will possibly change in a future version)
		\item No \verb§#extensional§ (but see F2LP manual for a simple substitute)
		\item FOL2ASP doesn't generate \verb§#hide§'s for newly introduced predicates (as \verb§#hide§ isn't supported anymore with Gringo4)
				
		\end{itemize}
				
		Note that with the converter being active, any script (e.g., Lua or Python) needs to be placed at the beginning of the input string. Only one script is allowed. The script is copied to the output too.\\
		Aggregates are allowed but treated as black-box atoms.		
		
	\addcontentsline{toc}{subsection}{\texttt{--folconv}}  \label{cmdline:folconv}
	
	\subsection*{\normalsize{\texttt{--SMTsolver filename arguments}}} specifies the executable of the optional external SMT solver which is called as a fallback-approach in case PrASP's built-in linear system solver failed to find a solution. Default SMT solver executable is \verb|cvc4| (Linux), \verb|cvc4.exe| (Windows) or \verb|cvc4_macosx|.\\
	
		Observe that \prasp does not (and cannot) check what the specified program is actually doing, so this switch should be used with care.\\
		
		Example: \verb§--SMTsolver cvc4 "--lang smt2 -m" § (default under Linux)
	
	\addcontentsline{toc}{subsection}{\texttt{--SMTsolver}}  \label{cmdline:SMTsolver}

	\subsection*{\normalsize{\texttt{--nonorm}}} specifies that learning results should not be normalized. Makes learning somewhat faster, but results are possibly probabilistically inconsistent if this switch is activated.
	
	\addcontentsline{toc}{subsection}{\texttt{--nonorm}}  \label{cmdline:nonorm}  
	
	\subsection*{\normalsize{\texttt{--maxconjexamples}}} computes the combined weight of learning examples as $e_1 \wedge e_2 \wedge ...$.
	
	\addcontentsline{toc}{subsection}{\texttt{--maxconjexamples}}  \label{cmdline:maxconjexamples}
	
	\subsection*{\normalsize{\texttt{--keepduplicateexamples}}} keeps any duplicate learning examples instead of filtering them out.
	
	\addcontentsline{toc}{subsection}{\texttt{--keepduplicateexamples}}  \label{cmdline:keepduplicateexamples} 
	
	\subsection*{\normalsize{\texttt{--showexpansion}}} 
	
	prints the expansion of the background knowledge and query file obtained by grounding and other transformations. This is particularly useful for debugging of programs which contain double- and triple-square formula weights.
	
	\addcontentsline{toc}{subsection}{\texttt{--showexpansion}}  \label{cmdline:showexpansion}
	
	\subsection*{\normalsize{\texttt{--showspan}}} prints the \textit{spanning program}.
	
	\addcontentsline{toc}{subsection}{\texttt{--showspan}}  \label{cmdline:showspan}
	
	\subsection*{\normalsize{\texttt{--debug}}} shows detailed information about the internal processing of inference and learning tasks. All important intermediate results of the inference pipeline are shown. Also, any temporary files are kept for inspection when \prasp ends.
	
	\addcontentsline{toc}{subsection}{\texttt{--debug}}  \label{cmdline:debug}
}

\newpage
\section{Remarks \& Troubleshooting}
\label{hints}

\begin{itemize}
\item Suggestions for the improvement of \prasp are welcome. Also, we are always interested in possible applications of \prasp - don't hesitate to contact us if you are interested in using \prasp in a research, educational or industry context.

\item When using PrASP, please keep in mind that the current version is a prototype. If you discover bugs or other issues (such as excessive running times), please report them to the author (for contact details see Sect. \ref{license}). Together with a description of the bug, if possible please also provide input files with which the bug can be reproduced. Also very helpful for us would be the console output from running your \prasp task with added switch \hyperref[cmdline:debug]{\texttt{--debug}}\\
With MacOSX and most Linux variants, you can redirect all output into a log file using \\
\verb§./prasp.sh§ $...$ \hyperref[cmdline:debug]{\texttt{--debug}} \verb§> praspLog.txt 2>&1§

\item Global variable domain declarations using \verb|#domain|, while being discouraged in plain ASP programming, are often useful in \prasp programs, in particular with Gringo/Clingo3 and in case FOL quantifiers are being used. The reason is that \prasp needs to create certain ASP formulas from weighted formulas, and this conversion might introduce unsafe variables. With Gringo/Clingo4, the only way to do this is do bind variables to domain predicates (predicates which can be fully evaluated during grounding) in the bodies of rules (ASP as well as FOL rules!), see Section \ref{weights}. With Gringo/Clingo 3, unsafe variables can additionally be prevented by explicitly stating the domains of variables using global declarations of form \verb|#domain predicate(Variable).| \\

Grounding happens at various stages of the preprocessing and inference pipeline. As for the grounding of formulas with \verb§[[...]]/[[[...]]]§ weights, please see Section \ref{weights}. As for the prevention of unsafe variables due to generation of spanning formulas, in some cases the use of \verb|#domain| can be avoided by allowing for FOL$\rightarrow$ASP conversion even if the \prasp program uses ASP syntax only while sticking to pure ASP syntax apart from the weights). See command line option \hyperref[cmdline:spangenconf]{\texttt{--spangenconf}} for how to enable this. This way, \prasp applies different spanning program creation patterns which are less prone of unsafe variable introduction. However, this and the consequential omission of \hyperref[cmdline:folconv]{\texttt{--folconv none}} slows \prasp down a bit (since a FOL2ASP converter needs to be called) and the generated spanning formulas become larger. \\

\item Gringo/Clingo 4 should basically work with \prasp, but support for ASPCore-2 syntax is still preliminary in \prasp \version.  The informal list of Gringo 3$\leftrightarrow$4 differences under\\ \verb§http://sourceforge.net/projects/potassco/files/gringo/4.4.0/§ might be helpful.\\
Observe that not only the syntax but also the semantics of aggregates such as \verb§#count§ and \verb§#sum§ has changed from Gringo/Clingo version 3 to version 4.

\item \underline{Reserved keywords}: Do not use any predicate, variable or constant name starting with any of the following reserved name prefixes: ``\verb§hp__atom§'', ``\verb§npred_atom§'', ``\verb§_new_pred§'', ``\verb§n_pred_§'', ``\verb§_NV_§'', ``\verb§___str§'', ``\verb§condPr§'', ``\verb§lhPp§'', ``\verb§Hp__var§''. 

\item Do not use FOL-style connectives within ASP-style rules or vice versa (e.g., instead of \verb§a :- b & c§, write \verb§a :- b, c § or  \verb§a <- b & c§).

\item \verb§a,b,c§ is not a well-formed formula. Write \verb§3{a,b,c}3§ or \verb§a & b & c§ (FOL).

\item Logical connective precedence imposed by the external converter F2LP and the internal converter FOL2ASP differ from each other. See \hyperref[cmdline:folconv]{\texttt{--folconv}}§.

%\item You cannot use E-notation (scientific number notation) in weights.

\item A probability of zero as a query result might indicate that some predicates occurring in the query formula aren't defined.

\item If weights with a question mark (\verb§[?]§ or \verb§[?|...]§) appear as query or learning results, \prasp couldn't solve the task or decided that the internally computed result isn't accurate enough.  Question marks in results can have a variety of causes, for example accumulated round-off errors, an insufficient number of sampled models, an unsuitable chosen sampling or inference method, or diverging parameter learning. To get rid of these problems, it is recommended to check input files for errors, to omit any command-line switches which may have a negative effect on accuracy (for example, \hyperref[cmdline:limitindepcombs]{\texttt{--limitindepcombs}}), and to consider selecting a different sampling or inference approach.

\item A message of the form ``warning: x is never defined (issued from external program clingo)'' can indicate a mistake in the background knowledge, but it can also be caused by flip-sampling (in which case the message can be ignored).

\item Message ``WARNING: Specified probabilities appear to be inconsistent ...'' indicates that the given background knowledge does not adhere to the probability axioms. In some such cases, \prasp nevertheless manages to compute a result, though it might be incorrect or inaccurate. Refer to Sect. \ref{inconsistency} for more information about inconsistency handling.\\
 On the other hand, not all occurrences of this warning indicate an actual problem, because this warning can - in rare cases - also result from ``over-declaration''
of weights: this means that redundant weights which are inferable from other weights are given explicitly in the background knowledge and the two respective probabilities (the given and the inferred one) differ slightly, e.g., due to internal round-off errors.
There is no way for \prasp to distinguish this case from an actual inconsistency.\\
In any
case, such warnings should be taken seriously and the background knowledge should be checked for mistakes. \\

Note that inconsistencies are \underline{not} detected by default (as this would make inference rather slow), so you need to enable consistency checking using \hyperref[cmdline:checkconsistency]{\texttt{--checkconsistency}}.

\item Automatic discovery of mutually independent atoms is not guaranteed to discover all independent atoms (as it relies on a rather coarse dependency graph output from Gringo 3). It might be preferable to declare independent atoms explicitly using meta-statement (see Section \ref{meta}) or conditional probabilities, and to deactivate automated discovery (using switch \hyperref[cmdline:noautoindeps]{\texttt{--noautoindeps}}), or to work with unspecified independence where possible. 

\item Simultaneously learning the weights of two or more hypotheses is a lot more difficult than learning just one hypothesis. The \verb§.hypoth§ file should therefore contain as few as possible formulas, ideally just a single formula. If multiple hypotheses do not depend on each other, it is better to learn them separately.

\item The following Gringo/Clingo- and Lparse/Smodels-features are currently not supported (this might change in future versions): multi-line formulas, assignments (operator \verb§:=§), bitwise-operators (because \prasp cannot distinguish them from certain FOL-symbols), optimization statements (e.g., \verb§#maximize§, \verb§#minimize§) (because they are no logical formulas).\\
\verb|#include| statements are supported but they are handled by \prasp directly, not by Gringo.\\
Within Gringo4-style aggregates, currently only literals are supported (e.g., \verb§1{a;b(x,y);not c}3§), no tuples as elements). Also, a Gringo4-style aggregate must be of the form \verb§s1<=aggrname{...}<=s2§ or \verb§s1{...}s2§. However, these restrictions only apply in cases where \prasp needs to parse the aggregate (e.g., with \hyperref[cmdline:dnf]{\texttt{--dnf}}, \hyperref[cmdline:mod0]{\texttt{--mod0}} and \hyperref[cmdline:mod1]{\texttt{--mod1}}). \textit{More tbw.}

%\item Quoted "\textit{string literals}" are not supported by the current input languages (they are also missing in plain Answer Set Prolog). Future support of string literals depends on whether they are supported by suitable ASP grounders. For now, if you just need alphanumeric strings without whitespace or special characters, writing, e.g., \verb§predicate(abc)§ instead of \verb§predicate("abc")§ should work.
 
\end{itemize}

\section{FAQ}
\label{faq}
\begin{description}

\item {\textbf{I'm getting confusing results for query ``\texttt{not not f}'' - why?}}

 the meaning of ``\verb§not not f§'' depends on the reasoner being used: if FOL$\rightarrow$ASP conversion is active, \prasp interprets this formula as a first-order formula which is equivalent to \verb§f§. However, if FOL$\rightarrow$ASP conversion is inactive (command-line switch \hyperref[cmdline:folconv]{\texttt{--folconv none}}), the result depends on the ASP grounder/solver being used. With Gringo/Clingo 3, ``\verb§not not f§'' is not a well-formed formula and thus you get an error message if you attempt to use it. With Gringo/Clingo 4, ``\verb§not not f§'' is interpreted as an ASP formula. If it is the only formula in a program, this program is unsatisfiable.  

\item{\textbf{The resulting weight of a query formula is \texttt{[0.2,0.1]}. That doesn't seem to make sense as an interval.}}

 \verb§[0.2,0.1]§ doesn't denote an interval but a list of point probabilities, e.g., from using \hyperref[cmdline:ndistrs]{\texttt{--ndistrs 2}} on the command-line. PrASP's interval notion uses a semicolon instead of a comma. A list of point probabilities also indirectly indicates a range of probabilities (in the case above, the range is [0.1...0.2]), but this range is not necessarily exhaustive. If \hyperref[cmdline:intervalresults]{\texttt{--intervalresults}} is used together with the default inference approach, \prasp uses linear programming to obtain interval results which are supposed to be exhaustive. 

\item {\textbf{I'm getting ``Unsafe variable'' errors with the following probabilistic FOL program. I'm using Gringo/Clingo 4.}}

\begin{verbatim}
   flup(1..4).
   
   nump(3).
   nump(4).
   
   [0.3] ![M,N]: (nump(N) & flup(M) -> flop(M,N)).
\end{verbatim}

 Although variables \verb§M§ and \verb§N§ are properly bound by domain predicates, the spanning formula generated for the FOL rule cannot be grounded because certain variables automatically introduced by F2LP aren't bound. This issue occurs because the FOL$\rightarrow$ASP converter F2LP is (at least up to version 1.3) designed for Gringo 3, where existing \verb§#domain§ statements need to be used by F2LP to prevent this problem. You can alternatively use the internal converter (see \hyperref[cmdline:folconv]{\texttt{--folconv}}) which harmonizes better with Gringo/Clingo 4. With the internal FOL2ASP converter, you can write the rule as follows, in order to indicate that \verb§nump§ and \verb§flup§ are domain predicates required to resolve the rule:
 
 \begin{verbatim}
 [0.3] ![M,N,nump(N),flup(M)]: (nump(N) & flup(M) -> flop(M,N)).
 \end{verbatim}
 With this, the "\verb§nump(N) & flup(M) ->§" part of the rule is even redundant.\\
  
 There are several other workarounds too: 1) switch to Gringo/Clingo 3 and add \verb§#domain§ statements for variables \verb§M§ and \verb§N§, \textit{or} 2) change the rule into ASP syntax, \textit{or} 3) change the weight to double-square notion (however, this changes the semantics of the rule, of course). Also, it is unclear if with future versions of F2LP this issue will still exist. \\
Note that even with Gringo/Clingo 4, the problem also doesn't occur if the weight is removed from the rule. 

% cannot happen anymore since 0.9.0:
%\item  {\textbf{I'm getting a ``Grounding failed'' error with the following program:}}
%
%\begin{verbatim}
%#domain bird(X).
%
%[[0.5]] fly(X) :- bird(X).
%
%bird(tweety).
%bird(twooty).
%\end{verbatim}
%
%Answer: The error occurs because when \prasp tries to ground weighted formula\\
% \verb§[[0.5]] fly(X) :- bird(X)§ it doesn't know yet the instances of variable \verb§X§. Grounding of individual formulas happens in the context of the preceding formulas. Reorder the formulas as follows to get rid of the error message:
%
%\begin{verbatim}
%bird(tweety).
%bird(twooty).
%
%#domain bird(X).
%
%[[0.5]] fly(X) :- bird(X).
%\end{verbatim}

\item{\textbf{I'm getting a syntax error with the following annotated formula:}}

\begin{verbatim}
[0.4] x :- 1 <= #sum{1:a:a;1:b:b;2:c:c} <= 3.
\end{verbatim}

The cause of this error is that in order to create a spanning formula (which covers both the aggregate and its negation), the sum-aggregate needs to be brought into a body position at one point, but the provided aggregate is a head-aggregate. To solve the issue, try to rewrite so that the aggregate is a body-aggregate (see Gringo/Clingo user's guide).

\item{\textbf{Using the internal FOL$\rightarrow$ASP converter, why do I receive an error with the following formula:}}

\texttt{[0.5] ![X, d(X)]: (X != 2 -> p(X)).}

The cause of the error is that the current beta-version of the built-in converter doesn't understand the infix \verb§!=§ operator yet. More precisely, it doesn't bind \verb§X§ to \verb§d§ in the presence of \verb§!=§. Rewrite as, e.g., follows.

\begin{verbatim}	
		[0.5] ![X, d(X)]: (not is2(X) -> p(X)).
		is2(2).
\end{verbatim}

\item{\textbf{In Section \ref{corealgos}, it is explained that \prasp does not necessarily ground annotated formulas individually. However, when examining the \prasp pipeline results with \texttt{--debug}, it seems non-ground weighted rules annotated with single-square brackets are converted into ground spanning formulas.}}

 Whether non-ground formulas annotated with weights in single-square brackets are grounded individually or not depends i.a. on whether a FOL$\longrightarrow$ASP converter is used or not. With \hyperref[cmdline:folconv]{\texttt{--folconv none}}, the spanning formula (actually a group of formulas) for such a weighted formula is non-ground, whereas with active converter, a single ground formula is generated (more precisely, a \textit{conjunction} of ground formulas). This behavior might change in future versions of \prasp.\\
Note that single-square weights are rarely useful with non-ground formulas; consider double- or triple-square weights, annotated disjunctions or conditional probabilities as alternatives with different semantics. 

\item {\textbf{I'm getting an error message with \texttt{--maxwalksat} which says that} \textbf{\texttt{:\hbox{-}\ \{not\ coin\_out(1,heads)\}0}} \textbf{is not a ``simple formula''}}

 This formula might have been automatically generated during generation of the spanning program or using \verb§--addnegf§. See \hyperref[cmdline:spangenconfv]{\texttt{--spangenconf}} for how to make PrASP use a different spanning formula policy.

\item {\textbf{I got result X for my query but I expected Y}}

 There are several possible reasons for this, but the two most common ones (apart from typos or other mistakes in the background knowledge or query) are 1) there are multiple valid solutions for your query including both X and Y and 2) you are using an approximate inference approach and X is not close enough to true result Y. To check for 1), you could try with \hyperref[cmdline:intervalresults]{\texttt{--intervalresults}} or \hyperref[cmdline:ndistrs]{\texttt{--ndistrs}} to see if there is a range of valid solutions. 2) is more difficult to tackle, but switch \hyperref[cmdline:check]{\texttt{--check}} might give you an estimation of how closely the selected inference approach reproduces weights in the background knowledge. You could then decide, e.g., to increase the number of sampled models, to change parameters of the inference approach you are using, to switch to a more suitable sampling approach, or to choose a different inference approach.\\
If you still get inexplicable results, please contact the author of \PrASP.

\item{\textbf{What is the purpose of meta-statements and unannotated formulas in query-files?}}

 These formulas and meta-statements are used as an additional context for the query formulas, e.g., to expand double-square weighted query formulas. Also, answer sets of query formulas are computed in this context in addition to the context in form of the spanning program computed from the background knowledge. \\

 If the context is already fully provided by means of the background knowledge, it is not required to repeat it in the query file (e.g., no need to repeat \verb§#domain§ declarations or domain predicate definitions in the query file). On the other hand, if the context required to ground query formulas is fully provided directly in the query file, there is no need to add information from the background knowledge to this context. As for the particular case of grounding formulas annotated with \verb§[[...]]§ or \verb§[[[...]]]§ also see \hyperref[cmdline:groundingconf]{\texttt{--groundingconf}} for how to instruct \prasp to use only the context in the query file itself to ground such query formulas (might provide some performance gain if the background knowledge is large).\\
 
 Note that unannotated formulas in the query file implicitly add to each query. E.g., if you would place \verb§false :- true§ in the query file, all query results would be 0.
  
\item{\textbf{What does it mean if \prasp quits with message ``primal infeasible''?}}

 This error message is issued by the linear programming solver employed by \prasp when called with switch \hyperref[cmdline:intervalresults]{\texttt{--intervalresults}}. It means that the solver couldn't find a solution. This might be because the system is unsolvable (inconsistent) or because the solver's default parameters are too tight (see  \hyperref[cmdline:linoptimconf]{\texttt{--linoptimconf}}).  

\item{\textbf{With \texttt{[0.5] a(X)} and the domain of X being $\mathtt{ \{1,2,3 \}}$, I'd expected that I'd get models $ \mathtt{\{a(1),a(2),a(3) \}}$, $ \mathtt{\{a(1),a(2) \}}$, $ \mathtt{\{a(1),a(3) \}}$, $ \mathtt{\{a(2),a(3) \}}$, $ \mathtt{\{a(1) \}}$, $ \mathtt{\{a(2) \}}$, $ \mathtt{\{a(3) \}}$, $\mathtt{ \{ \}}$.\\ But instead I get only $\mathtt{ \{a(1),a(2),a(3) \}}$ and $ \mathtt{\{ \}}$. Why?  }}

The spanning formula of \texttt{[0.5] a(X)} is $\mathtt{(\forall X: a(X)) | (not\ \forall X: a(X))}$. The right operand of the disjunction ($\mathtt{not\ \forall X: a(X))}$) stands for a constraints which informally says ``The model must not contain all three instances of \verb§a(X)§ (i.e., \verb§a(1)§ and \verb§a(2)§ and \verb§a(3)§) at the same time''.
What it does \textit{not} say (under ASP semantics) is that there must be any models which contain a true subset of $ \mathtt{\{a(1),a(2),a(3) \}}$ (in ASP terminology: $\mathtt{not\ \forall X: a(X)}$ is a consistency constraint which does not ``generate'' any atoms).\\

So the spanning formula only demands ``Either a model contains all three instances of \verb§a(X)§ (that is, \verb§a(1)§, \verb§a(2)§ and \verb§a(2)§), \textit{or} a model does not contain all three instances of \verb§a§''.\\

If there are further formulas in your background knowledge which generate individual instances of \verb§a(X)§, you'd obtain a different result (e.g., with \verb§[.] a(X)§ you get the result you had originally expected).

\item{\textbf{I'm getting unexpected results with background knowledge which contains \texttt{[0.3] h(X) :- t(Y).}}}

 Annotating a non-ground rule with a single-square bracket annotation applies the weight to the conjunction of all ground instances of the rule (see Sect. \ref{nonground}). If this is not what you want to express, consider annotated disjunctions, double-/square-bracket annotations or conditional probabilities instead.  

\item{\textbf{What does ``WARNING: ... info: atom does not occur in any rule head: \texttt{x}'' mean?}}

Atom \verb§x§ is undefined. This warning should be taken seriously, since \prasp makes by default a closed-world assumption: if an atom is not defined, it gets probability zero and doesn't appear in models, which is likely not intended. However, if the name of the atom is \verb§false§, the warning can be safely ignored (constraint \verb§:- false§ is automatically generated).\\
\noindent Example: If \verb§[0.7] bar :- foo§ would be the only formula in the background knowledge, the spanning program would only have a single possible world (the empty model) and thus \verb§bar :- foo§ couldn't have a probability of 0.7 (i.e., the background knowledge would be inconsistent). In this case, the problem would be solved by adding \verb§[.] foo§, i.e., \verb§foo | not foo§ (which defines \verb§foo§). 

\end{description}

%\newpage \section{API}
%
%\noindent {\color{Red} (tbw.) }
%
%\section{Building from Source Code}
%
%\prasp \version\ is written in Scala. To build \prasp from source code, the build tool SBT is required.\\
%
%\noindent {\color{Red} (tbw.) }

\newpage

\section{Contact Details, Terms of Use \& Disclaimer}
\label{license}

The developer of \PrASP and author of this report is Matthias Nickles.\\

\noindent \underline{\textbf{Contact details}}:\\

\noindent Dr. Matthias Nickles \\
College of Engineering \& Informatics \\
National University of Ireland, Galway \\
University Road 1, Galway City, Republic of Ireland\\
\noindent Email: \verb§matthias<dot>nickles<at>deri<dot>org§\\
%\noindent or \verb§matthias<dot>nickles<at>gmx<dot>net§\\
Tel. +353 (0)91 494441\\
%Web: \verb§https://www.linkedin.com/in/matthiasnickles§\\

\noindent \underline{\textbf{Terms of Use}}: [\textit{\textbf{see file TERMS\_OF\_USE\_AND\_DISCLAIMER.txt}}]\\

\noindent \underline{\textbf{Disclaimer}}: THE SOFTWARE \prasp IS PROVIDED “AS IS”, WITHOUT WARRANTY OF ANY KIND, EXPRESS OR IMPLIED, INCLUDING BUT NOT LIMITED TO THE WARRANTIES OF MERCHANTABILITY, FITNESS FOR A PARTICULAR PURPOSE AND NONINFRINGEMENT. IN NO EVENT SHALL THE AUTHORS OR COPYRIGHT HOLDERS BE LIABLE FOR ANY CLAIM, DAMAGES OR OTHER LIABILITY, WHETHER IN AN ACTION OF CONTRACT, TORT OR OTHERWISE, ARISING FROM, OUT OF OR IN CONNECTION WITH THE SOFTWARE \prasp OR THE USE OR OTHER DEALINGS IN THE SOFTWARE \prasp. 
%
%%\noindent PrASP Copyright (c) 2013-2015 Matthias Nickles\\
%
% PrASP (henceforth called "the software") is provided for non-commercial research and teaching purposes only. Any other use of the software or of derivative work (in particular any for-profit use) is strictly prohibited without the prior written consent of the copyright holders of the software.   
% THERE IS NO WARRANTY FOR THE SOFTWARE, TO THE EXTENT PERMITTED BY APPLICABLE LAW. THE COPYRIGHT HOLDERS AND OTHER PARTIES PROVIDE THE SOFTWARE ``AS IS'' WITHOUT WARRANTY OF ANY KIND, EITHER EXPRESSED OR IMPLIED, INCLUDING, BUT NOT LIMITED TO, THE IMPLIED WARRANTIES OF MERCHANTABILITY AND FITNESS FOR A PARTICULAR PURPOSE.  \\ 
% Any computer program which uses the software (or parts thereof) in any way, shape or form, and any derivative work, must retain the copyright notice above, these Terms of Use \& Disclaimer and the author contact details above.\\
%
%\noindent \textbf{Third-party software and licenses}\\

%\noindent \prasp includes the following third-party software: [\textit{see file README.txt}]

\newpage

%\addcontentsline{toc}{chapter}{Bibliography}

\newpage
%\printindex
%\addcontentsline{toc}{section}{Index}


\begin{thebibliography}{10}
\label{bibliography}

\bibitem{aspasp} Michael Gelfond, Vladimir Lifschitz: The stable model semantics for logic programming. In ICLP/SLP, Vol. 88, pp. 1070-1080, 1988.

\bibitem{asp} Vladimir Lifschitz: What Is Answer Set Programming? Proceedings AAAI'08, 2008. %\url{http://www.cs.utexas.edu/~ai-lab/pubs/wiasp.pdf}

\bibitem{f2lp} Joohyung Lee, Ravi Palla: F2LP - Computing Answer Sets of First Order Formulas. Proceedings of LPNMR-09, 2009.

\bibitem{cvc4} Clark Barrett, Christopher L. Conway, Morgan Deters, Liana Hadarean, Dejan Jovanovi\'{c}, Tim King, Andrew Reynolds, Cesare Tinelli: CVC4. Lecture Notes in Computer Science (LNCS), vol. 6806. Springer, 2011.

\bibitem{prism0} Taisuke Sato, Yoshitaka Kameya: PRISM: A Language for Symbolic-Statistical Modeling. Proceedings of the 15th International Joint Conference on Artificial Intelligence (IJCAI'97), 1997.

\bibitem{icl} David Poole: Abducing Through Negation as Failure: Stable Models Within the Independent Choice Logic. In Journal of Logic Programming. 44(1-3):5-35, 2000.

\bibitem{xor} Carla P. Gomes, Ashish Sabharwal, Bart Selman: Near-Uniform Sampling of Combinatorial Spaces Using XOR Constraints. In Procs. of Advances in Neural Information Processing Systems 19 (NIPS'06), 2006.

\bibitem{sato} Taisuke Sato: A Statistical Learning Method for Logic Programs with Distribution Semantics. In Proceedings
of the 12th International Conference on Logic Programming (ICLP), 1995.

\bibitem{prism} Taisuke Sato, Yoshitaka Kameya: Parameter Learning of Logic Programs for Symbolic-Statistical Modeling. Journal of Artificial Intelligence Research (JAIR), vol. 15, 2001.

\bibitem{plog} Chitta Baral, Michael Gelfond, Nelson Rushton: Probabilistic Reasoning with Answer Sets. Theory and Practice of Logic Programming, 2008.

\bibitem{nilsson} Nils J. Nilsson: Probabilistic Logic. Artificial Intelligence, 28:71–88, 1986.

\bibitem{problog} Luc De Raedt, Angelika Kimmig, Hannu Toivonen: ProbLog: A Probabilistic Prolog and its Application in Link Discovery. IJCAI 2007: 2462-2467, 2007.

\bibitem{progol} Stephen Muggleton: Inverse Entailment and Progol. In New Generation Comput. 13 (3 \& 4), 245-286, 1995.

\bibitem{weichsel} Kurt Weichselberger: The Theory of Interval Probability as a Unifying Concept for Uncertainty. In International Journal of Approximate Reasoning 24 (2–3): 149–170, 2000.

\bibitem{cozman2} Fabio Cozman: Credal Networks. In Artificial Intelligence Journal, vol. 120, pp. 199-233, 2000.

\bibitem{spirit} Wilhelm R{\"{o}}dder, Carl{-}Heinz Meyer: Coherent Knowledge Processing at Maximum Entropy by {SPIRIT}. Proceedings of the Twelfth Conference on Uncertainty in Artificial Intelligence (UAI'96), 1996.

\bibitem{emc} Wei Wei, Bart Selman: A New Approach to Model Counting. In Theory and Applications of Satisfiability Testing (SAT 2005). Springer LNCS, 2005.

\bibitem{mln} Matthew Richardson, Pedro Domingos: Markov Logic Networks. Machine Learning, vol. 62(1-2), 107–136, 2006.

\bibitem{mln2} Parag Singla, Pedro Domingos: Memory-Efficient Inference in Relational Domains. In Proceesings AAAI'06, AAAI Press, 2006.

\bibitem{mlnasp} Joohyung Lee, Yunsong Meng, Yi Wang: Markov Logic Style Weighted Rules under the Stable Model Semantics. In ICLP'15 (Technical Communications), Vol. 1433, 2015.

\bibitem{streamrule} Alessandra Mileo, Ahmed Abdelrahman, Sean Policarpio, Manfred Hauswirth: StreamRule: A Nonmonotonic Stream Reasoning System for the Semantic Web. In Procs. of Web Reasoning and Rule Systems - 7th International Conference (RR 2013). Springer LNCS Vol. 7994, 2013.

\bibitem{cozman1} Glauber De Bona, Fábio Gagliardi Cozman, Marcelo Finger:
Towards classifying propositional probabilistic logics. Journal of Applied Logic 12(3): 349-368, 2014.

\bibitem{thimm1} Matthias Thimm, Marc Finthammer, Sebastian Loh, Gabriele Kern-Isberner, Christoph Beierle: A System for Relational Probabilistic Reasoning on Maximum Entropy. Procs. 23rd International FLAIRS Conference, FLAIRS'10. AAAI Press, 2010.

\bibitem{Barzilai} Jonathan Barzilai and Jonathan M. Borwein: Two point step size gradient methods. IMA J. Numer. An., 8, 141–148, 1988.

\bibitem{ASPILP} Mark Law, Alessandra Russo, Krysia Broda: Inductive Learning of Answer Set Programs. In Proceedings of the 14th European Conference of Logics in Artificial Intelligence (JELIA'14). Springer, 2014.

\bibitem{kautz} Henry Kautz, Bart Selman, Yueyen Jiang: A General Stochastic Approach to Solving Problems with Hard and Soft Constraints. In The Satisfiability Problem: Theory and Applications, pages 573–586. American Mathematical
Society, New York, 1997.

\bibitem{annotatedDisjunctions} Joost Vennekens, Sofie Verbaeten, Maurice Bruynooghe: Logic Programs with Annotated Disjunctions. Proc. Intenational Conference on Logic Programming, 2004.

\bibitem{prasp15} Matthias Nickles: A Tool for Probabilistic Reasoning Based on Logic Programming and First-Order Theories Under Stable Model Semantics. Proceedings of the 15th European Conference On Logics In Artificial Intelligence (JELIA'16). Springer LNAI, 2016. 

\bibitem{prasp11} Matthias Nickles, Alessandra Mileo: Web Stream Reasoning Using Probabilistic Answer Set Programming. In Procs. of Web Reasoning and Rule Systems - 8th International Conference (RR 2014). Springer LNCS, 2014.

\bibitem{prasp12} Matthias Nickles, Alessandra Mileo: Probabilistic Inductive Logic Programming based on Answer Set Programming. Proceedings of the 15th International Workshop on Non-Monotonic Reasoning (NMR'14), 2014. 

\bibitem{prasp13} Matthias Nickles, Alessandra Mileo: A System for Probabilistic Inductive Answer Set Programming
Proceedings of the 9th International Conference on Scalable Uncertainty Management (SUM'15). Springer LNAI, 2015. 

\bibitem{prasp14} Matthias Nickles, Alessandra Mileo: A Hybrid Approach to Inference in Probabilistic Non-Monotonic Logic Programming. Proceedings of the 2015 Probabilistic Logic Programming Workshop (PLP'15), 2015.

\bibitem{coala} Martin Gebser, Torsten Grote, Torsten Schaub: Coala: A Compiler from Action Languages to ASP. Proceedings of the Twelfth European Conference on Logics in
Artificial Intelligence (JELIA'10), Springer LNAI, 2010.

\bibitem{james} Edwin T. Jaynes: Information Theory and Statistical Mechanics. Physical Review. Series II 106 (4): 620–630, 1957.

\bibitem{ad2} Fabrizio Riguzzi, Terrance Swift: Well-Definedness and Efficient Inference for Probabilistic Logic Programming Under the Distribution Semantics. In TPLP 13 (2), 279-302, 2013.

\bibitem{potassco} Martin Gebser, Benjamin Kaufmann, Roland Kaminski, Max Ostrowski, Torsten Schaub, Marius Schneider: Potassco: The {P}otsdam Answer Set Solving Collection. AI Communications, vol. 24, no 2, pp. 105-124, 2011.

\bibitem{potasscoUserGuide} \verb§http://sourceforge.net/projects/potassco/files/guide/§

\bibitem{MontyHall} Ronald J. Gould: Mathematics in Games, Sports, and Gambling: - The Games People Play. CRC Press, 2009

\bibitem{MontyHallWiki} \verb§https://en.wikipedia.org/wiki/Monty_Hall_problem§ (as of 08/11/2015)

\bibitem{annotDisjProblog} Bernd Gutmann: On Continuous Distributions and Parameter Estimation in Probabilistic Logic Programs. PhD thesis, Katholieke Universiteit Leuven, 2011.

\bibitem{lparse} Ilkka Niemel\"{a}, Patrik Simons, Tommi Syrj\"{a}nen: Smodels: A System for Answer Set Programming. Proceedings of the 8th International Workshop on Non-Monotonic Reasoning, 2000.

\bibitem{lee2} Joohyung Lee, Yi Wang: A Probabilistic Extension of the Stable Model Semantics. 2015 AAAI Spring Symposium on Logical Formalizations of Commonsense Reasoning, 2015.

\bibitem{Morais} Eduardo Menezes de Morais: Approximated Probabilistic Answer Set Programming. FLoC Workshop on Probabilistic Logic Programming, 2014.

\bibitem{NNLS} Charles L. Lawson, Richard J. Hanson: Solving Least Squares Problems. Society for Industrial and Applied Mathematics (SIAM), 1995.

\bibitem{FNNLS} Rasmus Bro, Sijmen De Jong: A Fast Non-Negativity-Constrained Least Squares Algorithm. In Journal of Chemometrics 11 (5): 393, 1997.

\bibitem{cqels} D. Le Phuoc, J. Xavier Parreira, M. Hausenblas, M. Hauswirth: Continuous Query Optimization and Evaluation over Unified Linked Stream Data and Linked Open Data. Technical Report TR-2010-09-27, DERI, 2010.

\bibitem{denecker} Marc Denecker, Maurice Bruynooghe, Victor Marek: Logic Programming Revisited: Logic Programs as
Inductive Definitions. In ACM Transactions on Computational Logic (TOCL) - Special issue devoted to Robert A. Kowalski, Volume 2 Issue 4, 2001

\bibitem{wiki1} {\small \verb|http://en.wikipedia.org/wiki/Marginal_probability#Real-world_example|} (as of 04/20/2015)

\bibitem{Lua} {\small \verb|http://www.lua.org/|} (as of 05/16/2016)

\end{thebibliography}
\end{document}